\documentclass[10pt,journal,compsoc]{IEEEtran}

\ifCLASSOPTIONcompsoc
  \usepackage[nocompress]{cite}
\else
  \usepackage{cite}
\fi

\ifCLASSINFOpdf
\else
\fi

\usepackage{graphicx}
\usepackage{subfigure}
\usepackage{comment}
\usepackage{amsmath,amssymb} 
\usepackage{color}
\usepackage[noend]{algpseudocode}
\usepackage{algorithmicx,algorithm}
\usepackage{multirow}
\usepackage{url}
\usepackage{booktabs}
\usepackage{float}
\usepackage[colorlinks,
            linkcolor=blue,
            anchorcolor=blue,
            citecolor=blue]{hyperref}

\begin{document}
%
\title{Globally-Optimal Contrast Maximisation\\for Event Cameras}
%
%
%
%

\author{Xin~Peng,~\IEEEmembership{Student~Member,~IEEE,}
        Ling~Gao,~\IEEEmembership{Student~Member,~IEEE,}\\
        Yifu~Wang,~\IEEEmembership{Student~Member,~IEEE,}
        and~Laurent~Kneip,~\IEEEmembership{Member,~IEEE}
\IEEEcompsocitemizethanks{
\IEEEcompsocthanksitem X. Peng is with the Mobile Perception Lab of School of Information Science and Technology, ShanghaiTech University, the Shanghai Institute of Microsystem and Information Technology, Chinese Academy of Sciences, and University of Chinese Academy of Sciences.
\IEEEcompsocthanksitem L. Gao and L. Kneip are with the Mobile Perception Lab of School of Information Science and Technology, ShanghaiTech University. L. Kneip is also with the Shanghai Engineering Research Center of Intelligent Vision and Imaging.
\IEEEcompsocthanksitem Y. Wang is with the Australian National University. 
\IEEEcompsocthanksitem E-mail: see http://mpl.sist.shanghaitech.edu.cn 
}
}

\IEEEtitleabstractindextext{%
\begin{abstract}
Event cameras are bio-inspired sensors that perform well in challenging illumination conditions and have high temporal resolution. However, their concept is fundamentally different from traditional frame-based cameras. The pixels of an event camera operate independently and asynchronously. They measure changes of the logarithmic brightness and return them in the highly discretised form of time-stamped events indicating a relative change of a certain quantity since the last event. New models and algorithms are needed to process this kind of measurements. The present work looks at several motion estimation problems with event cameras. The flow of the events is modelled by a general homographic warping in a space-time volume, and the objective is formulated as a maximisation of contrast within the image of warped events. Our core contribution consists of deriving globally optimal solutions to these generally non-convex problems, which removes the dependency on a good initial guess plaguing existing methods. Our methods rely on branch-and-bound optimisation and employ novel and efficient, recursive upper and lower bounds derived for six different contrast estimation functions. The practical validity of our approach is demonstrated by a successful application to three different event camera motion estimation problems.
\end{abstract}

}

\maketitle

\IEEEdisplaynontitleabstractindextext

%
\IEEEpeerreviewmaketitle

\IEEEraisesectionheading{\section{Introduction}\label{sec:introduction}}

\begin{figure*}[t]
\centering
\includegraphics[width=\textwidth]{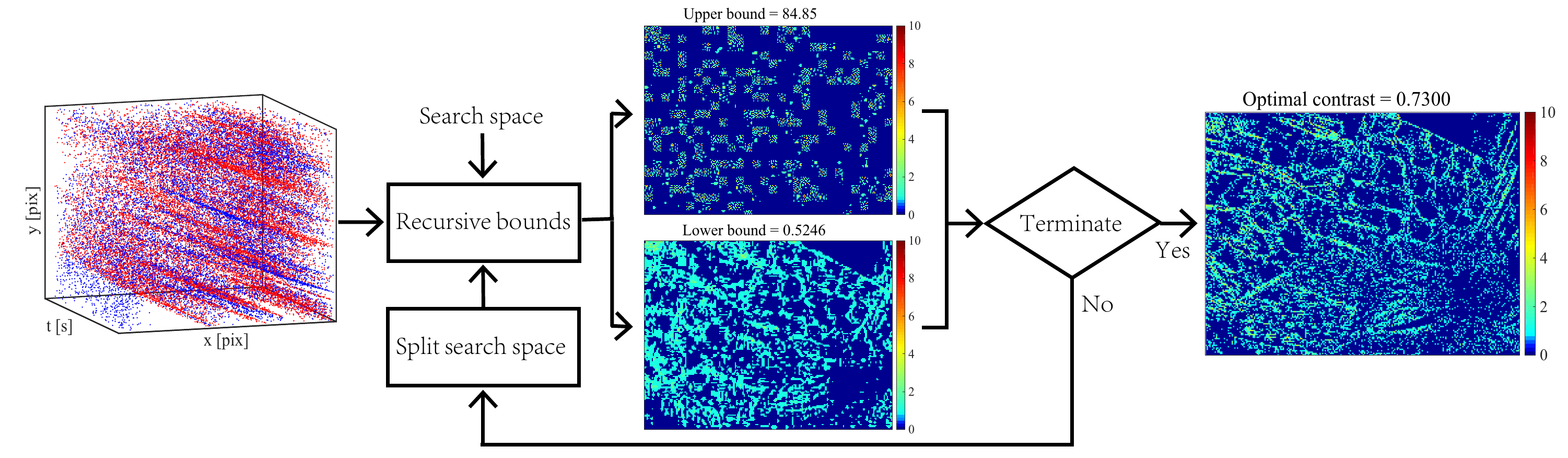} 
\label{fig:framework}
\caption{Globally-Optimal Contrast Maximisation Framework (GOCMF): Given a spatiotemporal event stream $\mathcal{E}$ and a parameter search space $\boldsymbol{\Theta}$, our method applies the branch-and-bound optimisation paradigm with a recursive evaluation on upper and lower bounds' values. The search space is split until upper and lower bounds' values converge, upon which the algorithm returns the globally optimal motion parameters of the considered contrast maximisation problem.}
\end{figure*}

\IEEEPARstart{V}{isual} perception plays an increasingly important role in a number of fields such as robotics, smart vehicles, and augmented/virtual reality (AR/VR). These are broad and complex areas of application that require the solution of a variety of problems including but not limited to image matching \cite{ma2020image,aires2008optical}, camera motion estimation \cite{kneip2011robust,scaramuzza2011visual,kneip2014opengv}, localisation \cite{ben2014review}, 3D reconstruction \cite{scharstein2002taxonomy, seitz2006comparison, butime20063d, ham2019computer}, and object segmentation \cite{papazoglou2013fast, che2019object}. Over the past several decades, the community has achieved great progress in traditional camera based solutions to these problems \cite{fuentes2015visual,cadena2016past}. Their robust application to real-world problems remains nonetheless difficult, which is---at least partially---due to the fact that traditional camera measurements are easily affected in situations of high dynamics, low texture or structure distinctiveness, and challenging illumination conditions.

Event cameras---also called Dynamic Vision Sensors (DVS)---represent an interesting alternative to traditional cameras pairing High Dynamic Range (HDR) with high temporal resolution. Different from standard cameras which capture intensity images at a certain frame rate, the pixels of an event camera sense changes of the logarithmic brightness and operate asynchronously and independently of one-another. An \textit{event} is triggered when the absolute difference between the current logarithmic intensity and the one at the time of the most recent event surpasses a given threshold. Due to their special bio-inspired design, event cameras have very low latency ($\thicksim1\mu s$) and very low power consumption. Moreover, event cameras possess a high dynamic range (e.g. 140 dB compared to 60 dB for standard cameras)\cite{gallego2019focus}.

Although event cameras have the potential to outperform standard cameras in challenging scenarios, the particular asynchronous and discretised nature of their outputs makes it difficult to directly migrate traditional computer vision algorithms to an event-based vision problem. Hence novel algorithms are needed. Recently, Gallego et al.~\cite{gallego2018unifying} introduced a unifying contrast maximisation (CM) pipeline with applications to various event-based vision problems, such as motion estimation, 3D reconstruction and optical flow estimation. The core idea of contrast maximisation is given by modelling trajectories in a space-time volume for the high-gradient points that generate events. Based on the assumption that the texture is given by a sparse set of sharp edges, the optimal motion parameters are found when the events exhibit maximum alignment with as few as possible point trajectories. Practically, the quality of the alignment is simply judged by measuring the contrast in the so-called Image of Warped Events (IWE). The objective has been successfully used for solving a variety of problems with event cameras such as optical flow~\cite{zhu2017event,gallego2018unifying,stoffregen2018simultaneous,DBLP:journals/corr/abs-1809-08625,zhu2019unsupervised,zhu2018ev}, moving object segmentation~\cite{stoffregen2018simultaneous,stoffregen2019event,mitrokhinevent}, 3D reconstruction~\cite{rebecq2018emvs,zhu2018realtime,zhu2019unsupervised,DBLP:journals/corr/abs-1809-08625}, and camera motion estimation~\cite{gallego2017accurate,gallego2018unifying}. However, existing methods mostly rely on local optimisation of the generally non-convex contrast maximisation objective (cf. Fig.~\ref{fig:heat map}), and thus fail if no good initial guess is given.

In this work, we present an efficient, globally-optimal contrast maximisation framework (GOCMF) based on the Branch-and-Bound (BnB) optimisation paradigm (epsilon-optimal solution). The work is inspired by our previous work \cite{peng2020globally} and relies on recursively evaluated upper and lower bounds of the optimisation objective. These bounds are paramount for the accuracy and efficiency of the algorithm, and we present their full derivation for a total of six different focus loss functions. We furthermore present an application of GOCMF to three different image or camera motion estimation problems. The work also shares commonalities with Liu et al.'s \cite{liu2020globally} globally optimal event camera rotation estimation algorithm, which however does not calculate the bounds recursively and thus is significantly slower.

Our detailed contributions are as follows:
\begin{itemize}
\item We propose a globally optimal contrast maximisation framework---GOCMF---which solves the contrast maximisation problem via Branch and Bound (BnB).
\item We derive bounds for six different contrast evaluation functions. The bounds are furthermore calculated recursively, which enables efficient processing.
\item We successfully apply this strategy to three common computer vision problems: optical flow estimation, camera rotation estimation, and motion estimation with a downward-facing event camera.
\end{itemize}

The paper is organised as follows: Section~\ref{sec:related_work} reviews related literature on event-based vision and applications of BnB in computer vision. Section~\ref{sec:premilinaries} reviews the general idea of contrast maximisation for event cameras. In Section~\ref{sec:GOCMF}, we then introduce our globally optimal contrast maximisation framework and derive the recursive upper and lower bounds. Sections~\ref{sec:optical_flow}, \ref{sec:downward-facing} and \ref{sec:Rotational} then present the application to optical flow estimation, downward-facing event camera motion estimation, and camera rotational estimation, respectively. To conclude, Sections~\ref{sec:analysis} and \ref{sec:conclusion} discuss a further analysis of the proposed algorithm as well as final remarks.

\section{Related Work}
\label{sec:related_work} 

Event cameras (e.g DVS) became commercially available in 2008, and there have been an increasing number of works on event-based computer vision problems in recent years. We first introduce existing works on event-based vision as well as specific algorithms that employ contrast maximisation as a core objective to be optimised for the solution of several different event-based vision problems. We furthermore review existing literature on the application of BnB for the solution of such problems with normal cameras.

\subsection{Event-based vision}

Most works for event-based motion estimation, focus on homography scenarios \cite{cook2011interacting,kim2008simultaneous,weikersdorfer2013simultaneous,gallego2017accurate} or fusion with other sensors \cite{kim2016real,zihao2017event,mueggler2018continuous}. Zhu~et~al.\cite{zhu2019neuromorphic} propose visual odemetry (VO) with an event camera and known map by feature tracking and PnP methods. Zhou~et~al.~\cite{zhou2020event} come up with the first parallel tracking-and-mapping VO with a stereo event camera. A spiking neural network is adopted by Gehrig~et~al. \cite{gehrig2020event} for regression of angular velocity. 
As for optical flow estimation, Benosman~et~al. \cite{benosman2013event} estimate normal flow by assuming the events are locally planar. Bardow and Davison \cite{bardow2016simultaneous} show a generic optical flow estimation method which also outputs image intensity by minimising a loss function with  smoothness regularisation. Stoffregen and Kleeman \cite{stoffregen2018simultaneous} employ contrast maximisation to show a simultaneous optical flow and segmentation algorithm.  Moreover, learning-based approaches recently became popular to be applied to event-based optical flow estimation \cite{zhu2019unsupervised,lee2020spike,kepple2020jointly}.

Event cameras also have been used for other computer vision tasks: object tracking \cite{wu2018high,rodriguez2020asynchronous,chen2020end},  pattern recognition \cite{li2016classification,belbachir2010high,chadha2019neuromorphic}, 3D reconstruction \cite{rebecq2016emvs,haessig2019spiking,zhou2018semi}, intensity image reconstruction \cite{pan2019bringing,rebecq2019high,lin2020learning} and so on.

\subsection{Event-based vision with contrast maximisation}

Gallego and Scaramuzza~\cite{gallego2017accurate} introduce the contrast of the image of warped events as a valid objective to be maximised in image registration problems. Their method estimates accurate angular velocities even in the presence of high-speed motion. At the same time, Zhu~et~al.~\cite{zhu2017event} come up with an expectation-maximisation based data tracking approach to find the best alignment of warped events. Rebecq~et~al.\cite{rebecq2018emvs ,rebecq2016emvs} propose Event-based Multi-View Stereo (EMVS) to estimate semi-dense 3D structure from an event camera with known poses. In their later work, they extend it to a real-time 6-DOF tracking and mapping system---Event-based Visual Odometry (EVO) \cite{rebecq2016evo}. The contrast maximisation concept is concluded in \cite{gallego2018unifying} with applications to motion, depth, and optical flow estimation. Mitrokhin~et~al. \cite{mitrokhinevent} also ultilize a parametric model to estimate the motion of the camera via motion compensation. Moving objects are detected during an iterative process that identifies events that are inconsistent with the primary displacement model. Zhu~et~al. \cite{zhu2019unsupervised} adopt contrast maximisation as a loss function to train an unsupervised neural network to estimate optical flow, depth and egomotion. Various reward functions that maximise contrast have been presented and analysed in the recent works of Gallego~et~al.~\cite{gallego2019focus} and Stoffregen and Kleeman~\cite{stoffregen2019event1}. However, practically all aforementioned papers apply local contrast maximisation, only.

 Liu~et~al.\cite{liu2020globally} propose a globally optimal contrast maximisation algorithm via BnB for camera rotation estimation. They derive the upper bound of the contrast objective by relaxing the sum of squares maximisation to an integer quadratic program. Nevertheless, time consumption is a severe problem for globally optimal methods. Our work introduces an alternative globally optimal contrast maximisation framework based on a more efficient, recursively evaluated upper bound. We present the detailed derivations for six different objective functions, and furthermore extend our previous work \cite{peng2020globally} to optical flow, planar motion and rotational motion estimation.

\subsection{Branch and bound in computer vision}

Branch-and-bound (BnB) is an optimisation strategy that guarantees global optimality without requiring any priors. There are quite a few solutions to geometric computer vision problems that are grounded on BnB, such as 2D-2D registration \cite{breuel2003implementation,pfeuffer2012discrete}, 3D-3D registration \cite{li20073d,olsson2008branch,bazin2012globally,bulow2009fast,campbell2016gogma,yang2015go,parra2016fast,liu2018efficient,hu2020globally}, 2D-3D registration \cite{brown2015globally,campbell2017globally,campbell2018globally,campbell2019alignment,brown2019family,liu20182d}, and relative pose estimation \cite{hartley2007global,kim2008motion,enqvist2009two,hartley2009global,zheng2011branch,yang2014optimal,ling2020efficient} methods. One of the earliest applications is proposed by Breuel  \cite{breuel2003implementation} who analyses its implementation and derives various bounds for 2D-2D point registration. 
Another pioneering application of BnB is given by Li and Hartley's \cite{li20073d} global 3D point registration method with unknown correspondences, which uses Lipschitz optimisation to search the space of 3D rotations. More modern, practically usable globally optimal 3D-3D registration methods are given by Go-ICP \cite{yang2015go} and GOGMA \cite{campbell2016gogma}, which use points and Gaussian Mixture Models (GMAs) to represent 3D structure. Bustos et al. \cite{parra2016fast} propose novel bounds to achieve faster rotation search based on cardinality maximisation. Liu et al. \cite{liu2018efficient} propose a new rotation invariant feature (RIF) that allows a prior, efficient, globally optimal estimation of the translation. More recently, Hu et al. \cite{hu2020globally} propose a simultaneous estimation of a symmetry plane for 3D-3D registration of partial scans with limited overlap.
Brown et al. \cite{brown2015globally, brown2019family} utilize the BnB paradigm for 2D-3D registration without known correspondences. Their globally optimal approach is applicable to both point and line features. Campbell et al. \cite{campbell2017globally,campbell2018globally} finally present a similar global inlier set cardinality maximisation method for simultaneous pose and correspondence estimation, however introduce novel tighter and approximation-free bounds. Most recently, Liu et al. \cite{liu20182d} and Campbell et al. \cite{campbell2019alignment} present further solutions to the 2D-3D problem by again relying on density distribution mixtures. With respect to relative pose estimation, Yang et al. \cite{yang2014optimal} extend the globally optimal \textit{rotation space search} by Hartley et al. \cite{hartley2007global} to essential matrix estimation in the presence of feature mismatches or outliers. Zheng et al. \cite{zheng2011branch} furthermore use BnB for finding a globally optimal fundamental matrix, however with an objective function that assumes no outliers. Ling et al. \cite{ling2020efficient} most recently introduce globally optimal homography matrix estimation for featureless scenarios. 

Although BnB has seen many past use cases in geometric vision, our application to event-camera based motion estimation relies on the substantially different objective of contrast maximisation, for which we derive novel recursive upper and lower bounds.

\section{Contrast Maximisation}
\label{sec:premilinaries}

Gallego~et~al.~\cite{gallego2018unifying} recently introduced contrast maximisation as a unifying framework allowing the solution of several important problems for dynamic vision sensors, in particular motion estimation problems in which the effect of camera motion may be described by a homography (e.g. motion in front of a plane, pure rotation). Our work relies on contrast maximisation, which we therefore briefly review in the following.

The core idea of contrast maximisation is relatively straightforward: the flow of the events is modelled by a time-parametrised homography. An event camera outputs a sequence of \textit{events} denoting temporal logarithmic brightness changes above a certain threshold. An event $e = \left\{ \mathbf{x},\ t,\ s \right\}$ is described by its pixel position $\mathbf{x} = [x\text{ }y]^\mathsf{T}$, timestamp $t$, and polarity $s$ (the latter indicates whether the brightness is increasing or decreasing, and is ignored in the present work). Given its position and time-stamp, every event may therefore be warped back along a point-trajectory into a reference view with timestamp $t_{\text{ref}}$. Since events are more likely to be generated by high-gradient edges, the correct homographic warping parameters are found when the unwarped events align along an as sharp as possible edge-map in the reference view, i.e. the Image of Warped Events (IWE). Gallego~et~al.~\cite{gallego2018unifying} simply propose to consider the contrast of the IWE as a reward function to identify the correct homographic warping parameters. Note that homographic warping functions include 2D affine and Euclidean transformations, and thus can be used in a variety of vision problems such as optical flow, feature tracking, or fronto-parallel motion estimation.
 
Suppose we are given a set of $N$ events $\mathcal{E} = \{e_{k}\}_{k=1}^{N}$. We define a general warp function ${\mathbf{x}_k^{\prime}}=W(\mathbf{x}_k,t_{k};\boldsymbol{\theta})$ that returns the position ${\mathbf{x}_k^{\prime}}$ of an event $e_k$ in the reference view at time $t_{\text{ref}}$. $\boldsymbol{\theta}$ is a vector of warping parameters. The IWE is generated by accumulating warped events at each discrete pixel location:
\begin{equation}
  I(\mathbf{p}_{ij};\boldsymbol{\theta}) = \sum_{k = 1}^{N}\mathbf{1}(\mathbf{p}_{ij}-{\mathbf{x}_k^{\prime}}) =
  \sum_{k = 1}^{N}\mathbf{1}(\mathbf{p}_{ij}-W(\mathbf{x}_k,t_{k};\boldsymbol{\theta})) \,,
  \label{equ:pixel intensity}
\end{equation}
where $\mathbf{1}(\cdot)$ is an indicator function that counts 1 if the absolute value of $(\mathbf{p}_{ij} - {\mathbf{x}_k^{\prime}})$ is less than a threshold $\epsilon$ in each coordinate, and otherwise 0. $\mathbf{p}_{ij}$ is a pixel in the IWE with coordinates $[i\text{ }j]^\mathsf{T}$, and we refer to it as an \textit{accumulator} location. We set $\epsilon = 0.5$ such that each warped event will increment one accumulator only.

Existing approaches replace the indicator function with a Gaussian kernel to make the IWE a smooth function of the warped events, and thus solve contrast maximisation problems via local optimisation methods (cf. \cite{gallego2017accurate,gallego2018unifying,gallego2019focus}).
In contrast, we propose a method that is able to find the global optimum of the above, discrete objective function.

As introduced in \cite{stoffregen2019event1,gallego2019focus}, reward functions for event un-warping all rely on the idea of maximising the contrast or sharpness of the IWE (they are also denoted as \textit{focus loss functions}). They proceed by integration over the entire set of accumulators, which we denote $\mathcal{P}$. The most relevant six objective functions for us are introduced in Section~\ref{sec:GOCMF} and Table~\ref{tab:upper_bounds}.

\begin{table*}[t]
\centering
\caption{Recursive Upper and Lower Bounds for six focus loss functions}
\label{tab:upper_bounds}
\renewcommand\arraystretch{2}
\begin{tabular}{llll}
\toprule
 & \textbf{Upper Bound} $\overline{L_{N}}$ & \textbf{Lower Bound} $\underline{L_{N}}$ & $L_0$ \\
\midrule
 \textbf{SoS}       &  $ \overline{L_{N-1}} + 1 + 2 Q$                    &   $\underline{L_{N-1}} + 1 + 2 I^{N-1}(\boldsymbol{\eta}_N^{\boldsymbol{\theta}_{0}};\boldsymbol{\theta}_0)$  &    0                     \\ 
 \textbf{Var}       &   $\overline{L_{N-1}} + \frac{1}{N_{p}} - \frac{2 \mu_{I}}{N_{p}} + \frac{2}{N_{p}} Q$   &   $\underline{L_{N-1}} + \frac{1}{N_{p}} - \frac{2 \mu_{I}}{N_{p}} + \frac{2}{N_{p}} I^{N-1}(\boldsymbol{\eta}_N^{\boldsymbol{\theta}_{0}};\boldsymbol{\theta}_0)$                    &   $\mu_{I}^2$                     \\
 \textbf{SoE}       &   $\overline{L_{N-1}} + (e-1) e^{Q}$                   &  $\underline{L_{N-1}} + (e-1) e^{I^{N-1}(\boldsymbol{\eta}_N^{\boldsymbol{\theta}_{0}};\boldsymbol{\theta}_0)}$                    &   $N_{p}$                    \\ 
 \textbf{SoSA}      &   $\overline{L_{N-1}} + (e^{-\delta}-1) e^{-\delta \cdot Q}$                   &   $\underline{L_{N-1}} + (e^{-\delta}-1) e^{-\delta \cdot I^{N-1}(\boldsymbol{\eta}_N^{\boldsymbol{\theta}_{0}};\boldsymbol{\theta}_0)}$                    &    $N_{p}$                    \\ 
 \textbf{SoEaS}     &  $\overline{L_{N-1}} + w_1 (e-1) e^{Q} + w_2 + 2 w_2 Q $            & $\underline{L_{N-1}} + w_1 (e-1) e^{I^{N-1}(\boldsymbol{\eta}_N^{\boldsymbol{\theta}_{0}};\boldsymbol{\theta}_0)} + w_2 + 2 w_2 I^{N-1}(\boldsymbol{\eta}_N^{\boldsymbol{\theta}_{0}};\boldsymbol{\theta}_0) $            &   $w_1 N_{p}$                                \\ 
 \textbf{SoSAaS}    &  $\overline{L_{N-1}} + w_1 (e^{-\delta}-1)e^{-\delta Q} + w_2 + 2 w_2 Q $   &   $\underline{L_{N-1}} + w_1 (e^{-\delta}-1)e^{-\delta I^{N-1}(\boldsymbol{\eta}_N^{\boldsymbol{\theta}_{0}};\boldsymbol{\theta}_0)} + w_2 + 2 w_2 I^{N-1}(\boldsymbol{\eta}_N^{\boldsymbol{\theta}_{0}};\boldsymbol{\theta}_0) $                   &                       $w_1 N_{p}$                                \\ 
 \bottomrule
\end{tabular}
\end{table*}

\section{Globally Maximised Contrast using Branch and Bound}
\label{sec:GOCMF}

\begin{figure}[t]
\centering
\subfigure[N/E = 0]{
\includegraphics[width=0.20\textwidth]{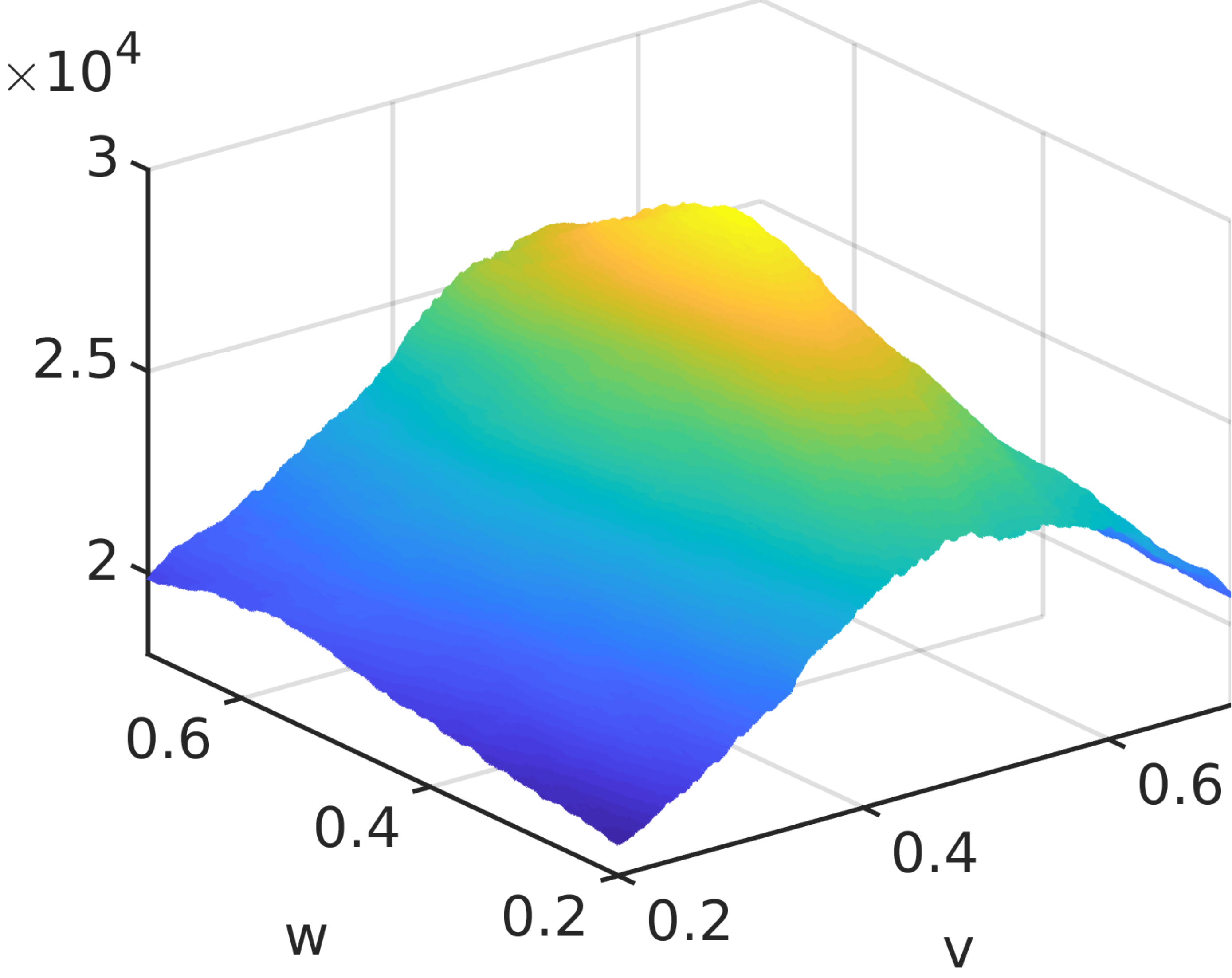} 
\label{fig:NtoE_0}
}
\subfigure[N/E = 0.02]{
\includegraphics[width=0.20\textwidth]{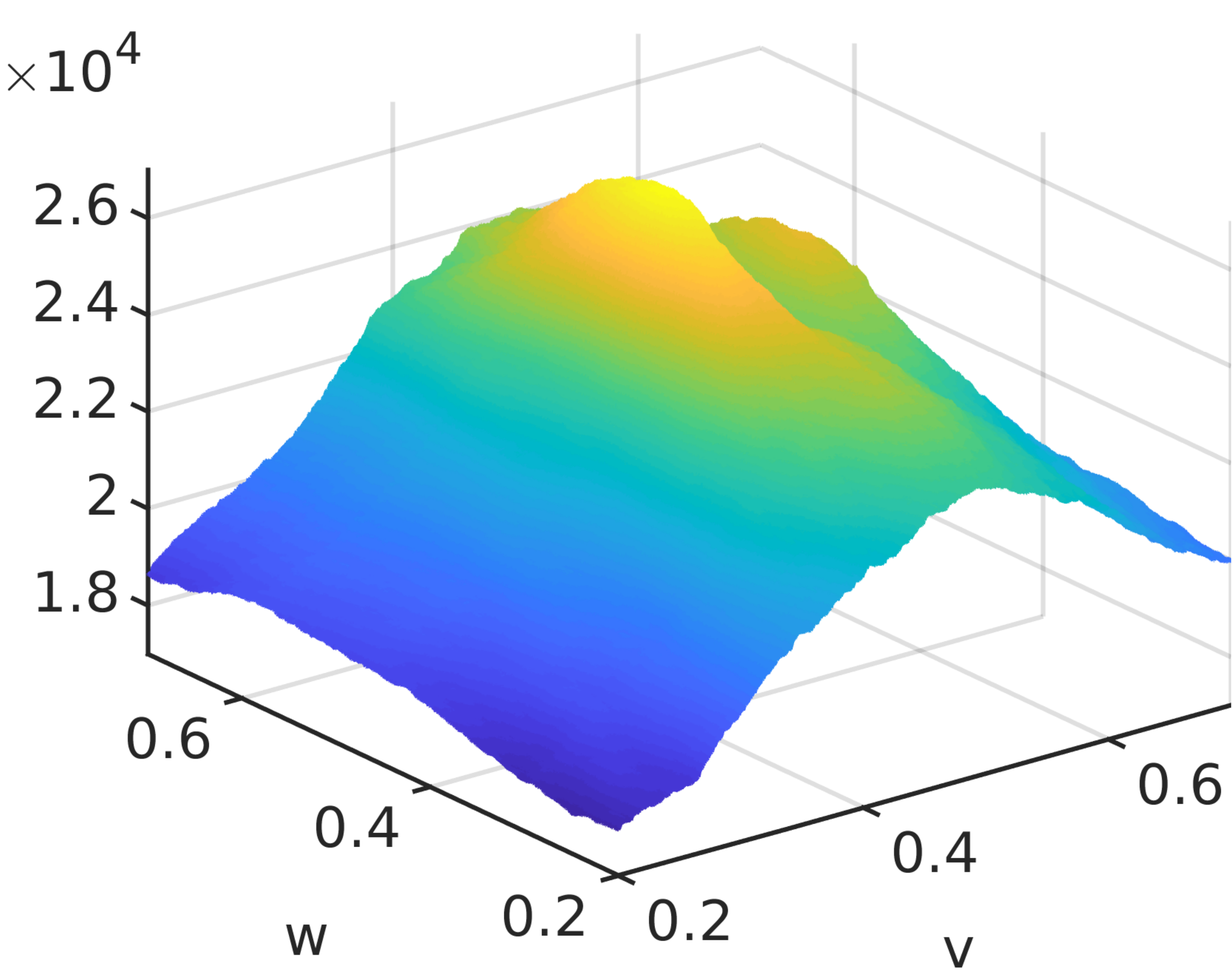}
\label{fig:NtoE_002}
}
\subfigure[N/E = 0.10]{
\includegraphics[width=0.20\textwidth]{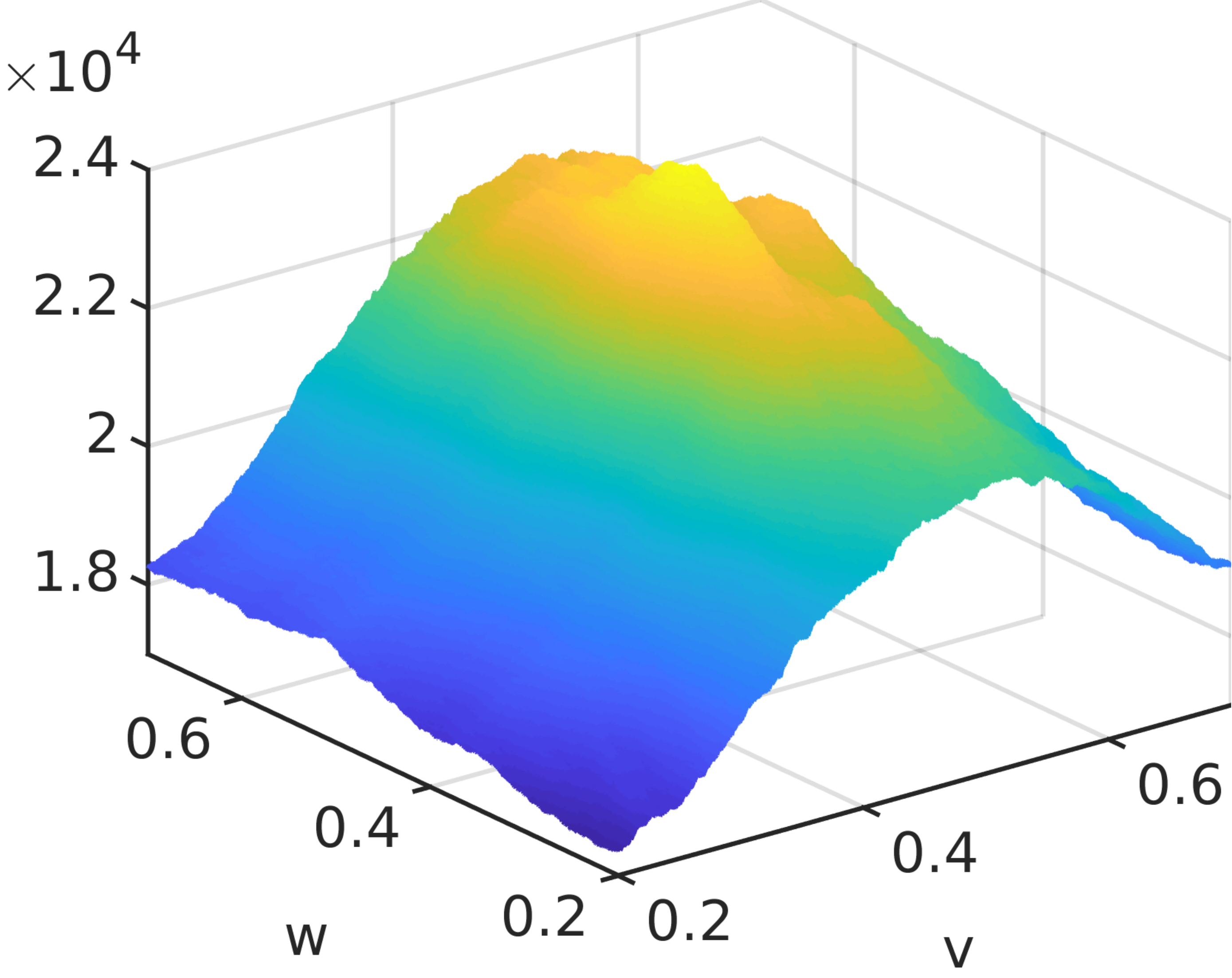}
\label{fig:NtoE_01}
}
\subfigure[N/E = 0.18]{
\includegraphics[width=0.20\textwidth]{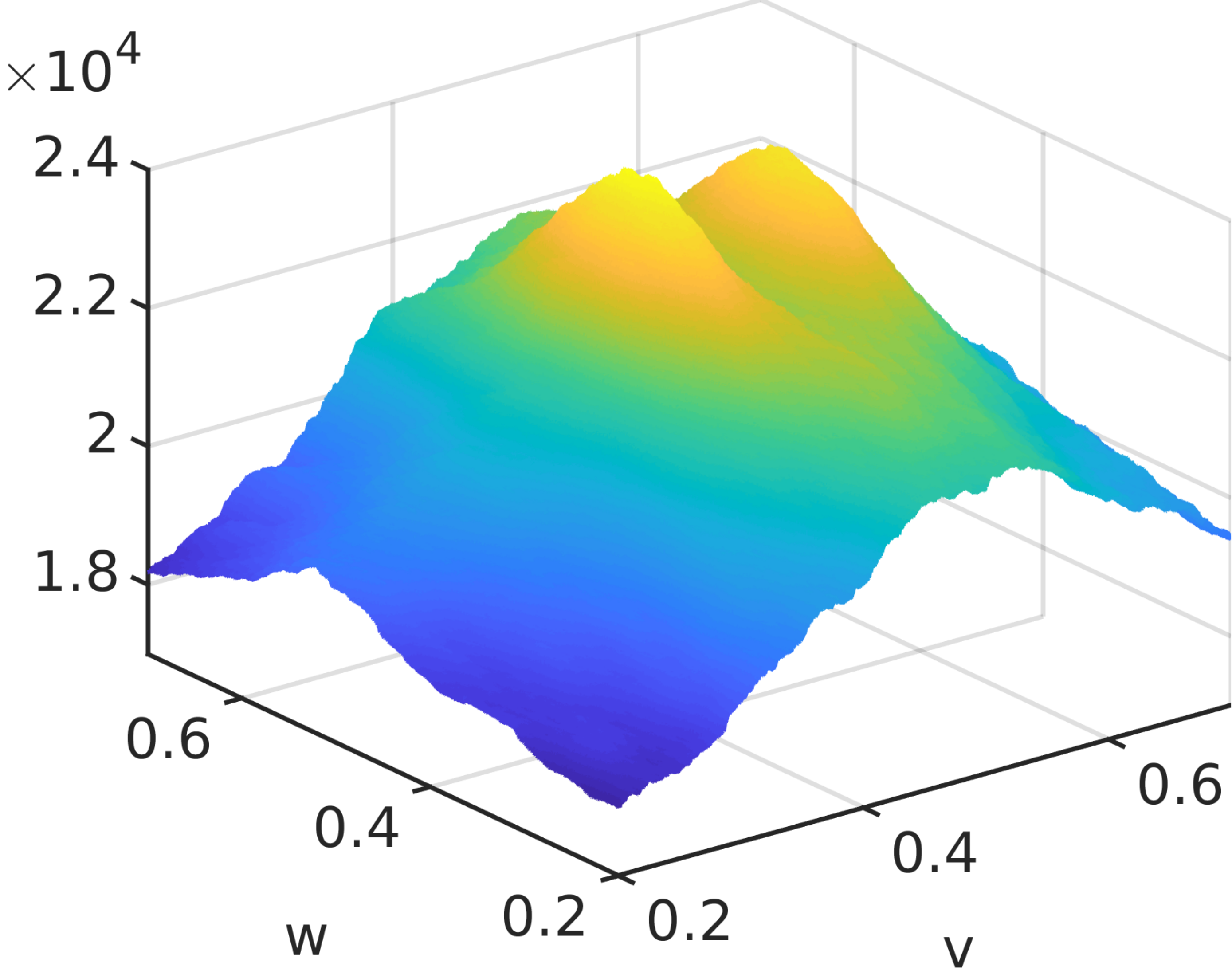} 
\label{fig:NtoE_018}
}
\caption{Visualization of the sum-of-squares contrast function. The camera is moving in front of a plane, and the motion parameters are given by translational and rotational velocity (cf. Section \ref{sec:casestudy}). The sub-figures from (a) to (d) are functions with increasing Noise-to-Events (N/E) ratios. Note that contrast functions are non-convex.}
\label{fig:heat map}
\end{figure}

Fig.~\ref{fig:heat map} illustrates how contrast maximisation for motion estimation is in general a non-convex problem, meaning that local optimisation may be sensitive to the initial parameters and not find the global optimum. We tackle this problem by introducing a globally optimal solution to contrast maximisation using Branch and Bound (BnB) optimisation. BnB is an algorithmic paradigm in which the solution space is subdivided into branches in which we then find upper and lower bounds for the maximal objective value. The globally optimal solution is isolated by an iterative search in which entire branches are discarded if their upper bound for the maximum objective value remains lower than the lower bound in another branch. The most important factor deciding the effectiveness of this approach is given by the tightness of the bounds.

Our core contribution is given by a recursive method to efficiently calculate upper and lower bounds for the maximum value of a contrast maximisation function over a given branch. In short, the main idea is given by expressing a bound over $(N+1)$ events as a function of the bound over $N$ events plus the contribution of one additional event. The strategy can be similarly applied to all six relevant contrast functions, which is why we limit the exposition in Section \ref{sec:uplowbounds} to the derivation of bounds for the objective function ``sum of squares ($L_{\mathrm{SoS}}$)''. Detailed derivations for all further loss functions are provided in Section~\ref{sec:other_bounds}.

\subsection{Objective Function}

In the following, we assume that $L_N=L_{\mathrm{SoS}}$. The maximum objective function value over all $N$ events in a given time interval $[t_{\text{ref}}, t_{\text{ref}}+\Delta T]$ is given by
\begin{equation}
  \max_{\substack{\boldsymbol{\theta}\in\boldsymbol{\Theta}}}L_{N} = \max_{\substack{\boldsymbol{\theta}\in\boldsymbol{\Theta}}} \sum_{\textbf{p}_{ij}\in\mathcal{P}} \left[ \sum_{k = 1}^{N}\boldsymbol{1} \left( \textbf{p}_{ij}-W(\mathbf{x}_k,t_{k};\boldsymbol{\theta}) \right) \right]^2 \,,
\end{equation}
where $\boldsymbol{\Theta}$ is the search space (i.e. branch or sub-branch) over which we want to maximise the objective. 
For alternative classical problems the measurement samples can be evaluated in parallel (e.g. inlier/outlier decisions for pairwise correspondences), whereas the contrast of the IWE is a measure that depends on all events. Note furthermore that events are processed in a given order, which---in our work---is set to the temporal order in which the events are captured. Intuitively speaking, our discrete problem can be described as finding optimal parameters that allocate $N$ events to a grid pattern such that the sum of squares of each cell's value is maximised (cf. Fig.~\ref{fig:iweConstruction}).

\subsection{Upper and Lower Bound}
\label{sec:uplowbounds}
\begin{figure*}[t]
    \centering
    \includegraphics[width=0.95\textwidth]{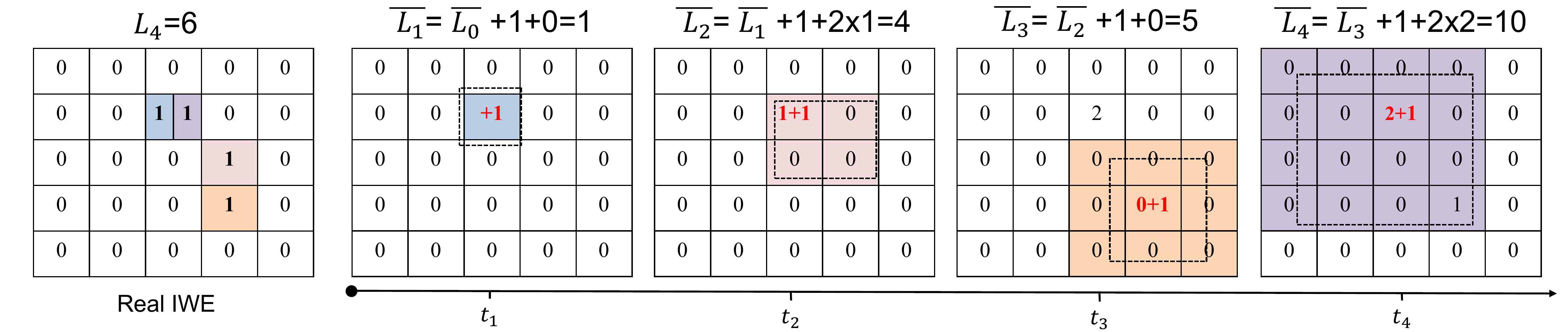}
    \caption{An example of the incremental update of the upper bound IWE $\overline{I}^N$. For this example, the event number $N = 4$, and different events are indicated by different colors. The left matrix shows the constructed IWE with ground truth motion parameters. The incremental update of the upper bound IWE $\overline{I}^N$ is shown in the right matrices. For each new event $e$, we choose and increment the currently maximal accumulator in the bounding box $\mathcal{P}^{\boldsymbol{\Theta}}$ (the rectangle bounding the dashed line formed by all possible locations $W(\mathbf{x},t;\boldsymbol{\theta}\in\boldsymbol{\Theta})$). We simply increment the center of the bounding box if no other accumulator exists. It is easy to see that the upper bound is bigger than the optimal result.}
    \label{fig:iweConstruction}
\end{figure*}
We calculate the bounds recursively by processing the events one-by-one, each time updating the IWE. The events are notably processed in temporal order with increasing timestamps.

For the lower bound, it is readily given by evaluating the contrast function at an arbitrary point on the search space interval $\boldsymbol{\Theta}$, which is commonly picked as the interval center $\boldsymbol{\theta}_{0}$. We present a recursive rule to efficiently evaluate the lower bound.

\noindent \textbf{Theorem 1.} \textit{For search space $\boldsymbol{\Theta}$ centered at $\boldsymbol{\theta}_{0}$, the lower bound of $SoS$-based contrast maximisation may be given by}
\begin{equation}
  \underline{L_{N}} = \underline{L_{N-1}}+1+2 I^{N-1}(\boldsymbol{\eta}_{N}^{\boldsymbol{\theta}_{0}};\boldsymbol{\theta}_{0}) \,,
\label{equ:lower bound}
\end{equation}
\textit{where $I^{N-1}(\mathbf{p}_{ij};\boldsymbol{\theta}_{0})$ is the incrementally constructed IWE, its exponent $(N-1) $ (where $N\geq1$) denotes the number of events that have already been taken into account, and}
\begin{equation}
  \boldsymbol{\eta}_{N}^{\boldsymbol{\theta}_{0}} = \operatorname{round}(W(\mathbf{x}_{N},t_{N};\boldsymbol{\theta}_{0}))
\end{equation}
\textit{returns the accumulator closest to the warped position of the $N$-th event.}

\noindent \textbf{Proof 1.} 
    According to the definition of the sum of squares focus loss function, 
\begin{equation}
\begin{split}
  \underline{L_{N}} &=  \sum_{\textbf{p}_{ij}\in\mathcal{P}} \left[ \sum_{k = 1}^{N}\boldsymbol{1} \left( \textbf{p}_{ij}-W(\mathbf{x}_k,t_{k};\boldsymbol{\theta}_{0}) \right) \right]^2 \\
  &= \sum_{\mathbf{p}_{ij} \in \mathcal{P}} 
           \left[ I^{N-1}(\mathbf{p}_{ij};\boldsymbol{\theta}_0)
                  + \boldsymbol{1} \left( \mathbf{p}_{ij}-W(\mathbf{x}_{N},t_{N};\boldsymbol{\theta}_{0}) \right) \right]^2 \\
          &= a+b+c \, \text{, where}
\end{split}
\end{equation}
\begin{equation*}
\begin{split}
  & I^{N-1}(\mathbf{p}_{ij};\boldsymbol{\theta}_{0}) = \sum_{k=1}^{N-1} \boldsymbol{1} \left( \textbf{p}_{ij}-W(\mathbf{x}_k,t_{k};\boldsymbol{\theta}_0) \right) \\
    & a =  \sum_{\mathbf{p}_{ij} \in \mathcal{P}} I^{N-1}(\mathbf{p}_{ij};\boldsymbol{\theta}_0)^2 \,, \\ 
    & b =  2\sum_{\mathbf{p}_{ij}\in\mathcal{P}}
        \left[ \boldsymbol{1}(\mathbf{p}_{ij}-W(\mathbf{x}_{N},t_{N};\boldsymbol{\theta}_0)) I^{N-1}(\mathbf{p}_{ij};\boldsymbol{\theta}_0) \right] \,, \\
    & c =  \sum_{\mathbf{p}_{ij}\in\mathcal{P}}
        \left[ \boldsymbol{1}(\mathbf{p}_{ij}-W(\mathbf{x}_{N},t_{N};\boldsymbol{\theta}_0)) \right]^2
        \,.
\end{split}
\end{equation*}
It is clear that $a = \underline{L_{N-1}}$. In $c$, owing to the definition of our indicator function, only the $\mathbf{p}_{ij}$ which is closest to $W(\mathbf{x}_{N},t_{N};\boldsymbol{\theta}_0)$ makes a contribution, thus we have $c = 1$. For $b$, the term $\boldsymbol{1}(\mathbf{p}_{ij}-W(\mathbf{x}_{N},t_{N};\boldsymbol{\theta}_0))$ is simply zero unless we are considering an accumulator $\mathbf{p}_{ij} = \boldsymbol{\eta}_{N}^{\boldsymbol{\theta}_0}$, which gives $b = 2I^{N-1}(\boldsymbol{\eta}_{N}^{\boldsymbol{\theta}_0};\boldsymbol{\theta}_{0})$. Thus we obtain (\ref{equ:lower bound}). Note that the IWE is iteratively updated by incrementing the accumulator which locates closest to $ \boldsymbol{\eta}_{N}^{\boldsymbol{\theta}_0}$.
\begin{figure}[t]
    \centering
    \includegraphics[width = 0.36\textwidth, height= 0.2\textwidth]{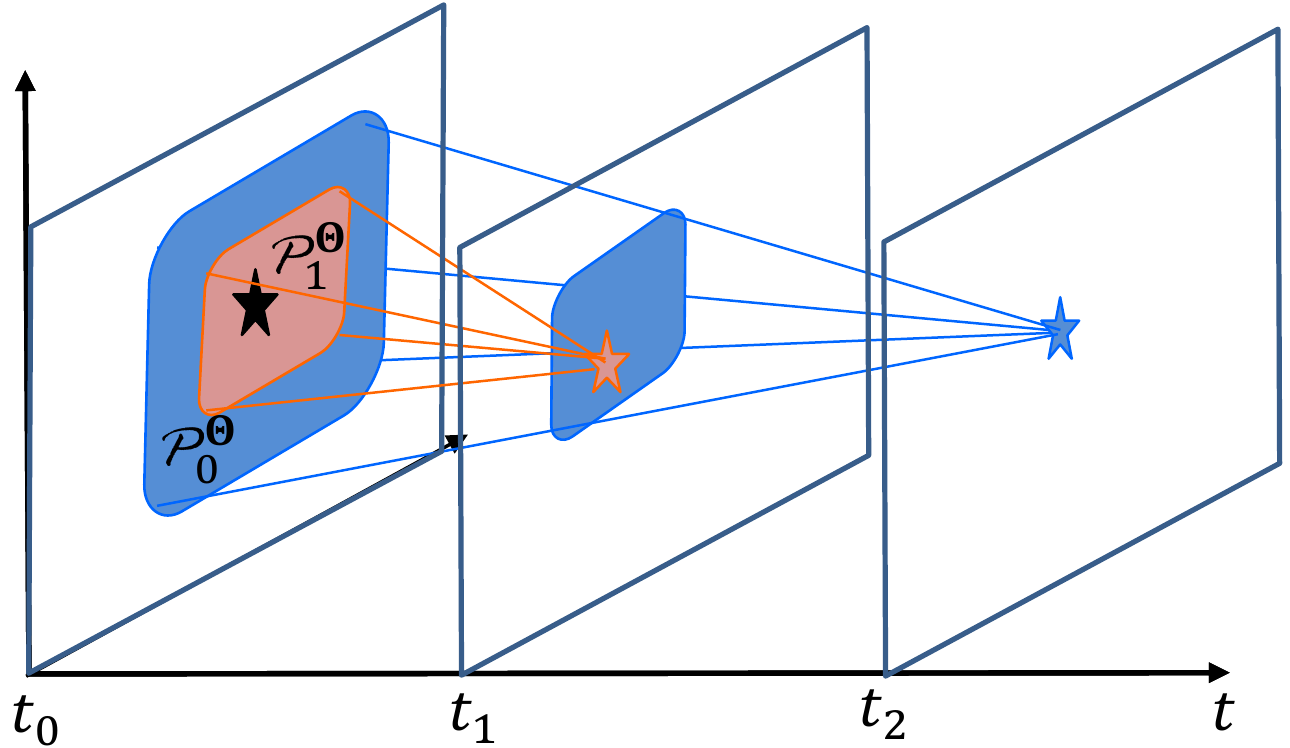}
    \caption{Bounding boxes of two events generated by a same point. Given two events $e_1$ (orange) and $e_2$ (blue) with timestamps $t_1$ and $t_2$ generated by a same 3D point, as illustrated in the figure. Uncertainty typically increases with the timestamp, hence the bounding box $\mathcal{P}^{\boldsymbol{\Theta}}_{1} \subseteq \mathcal{P}^{\boldsymbol{\Theta}}_{2}$. }
    \label{fig:nested_boundingbox}
\end{figure}
\noindent \rule[5pt]{0.49\textwidth}{0.05em}

We now proceed to our main contribution, a recursive upper bound for the contrast maximisation problem. Let us define $\mathcal{P}^{\boldsymbol{\Theta}}_{k}$ as the bounding box around all possible locations $W(\mathbf{x}_k,t_k;\boldsymbol{\theta} \in \boldsymbol{\Theta})$ of the un-warped event.
Lemma~1 is introduced as follows.

\noindent \textbf{Lemma 1.} 
\textit{
Given a search space $\boldsymbol{\theta} \in \boldsymbol{\Theta}$, for a small enough time interval, if $W(\mathbf{x}_i,t_i;\boldsymbol{\theta})$ = $W(\mathbf{x}_j,t_j;\boldsymbol{\theta})$, $t_i \leq t_j$ and $0 < i < j \leq N$, we have $\mathcal{P}^{\boldsymbol{\Theta}}_{i} \subseteq \mathcal{P}^{\boldsymbol{\Theta}}_{j}$. An intuitive explanation is given in Fig.~\ref{fig:nested_boundingbox}. Note that ``a small enough interval'' here simply denotes an interval for which the constant velocity assumption is sufficiently valid.
\label{lemma:nest_boundingbox}
}

Lemma~1 enables deriving our recursive upper bound.

\noindent \textbf{Theorem 2.} 
\textit{
The upper bound of the objective function $L_N$ for SoS-based contrast maximisation satisfies
\begin{eqnarray}
     L_{N} &=& L_{N-1}+1+2I^{N-1}(\boldsymbol{\eta}_{N}^{\hat{\boldsymbol{\theta}}};\hat{\boldsymbol{\theta}}) \label{eq:opt_recursive_formulation}\\
    & \leq& \overline{L_{N-1}}+1+2Q^{N-1} = \overline{L_{N}},  \  \label{eq:upper_bound} \\  
   \text{where } Q^{N-1}  & =& \max_{\substack{\mathbf{p}_{ij}\in\mathcal{P}_{N}^{\boldsymbol{\Theta}}}} \overline{I}^{N-1}(\mathbf{p}_{ij}) \geq I^{N-1}(\boldsymbol{\eta}_{N}^{\hat{\boldsymbol{\theta}}};\hat{\boldsymbol{\theta}}) \nonumber \,
\end{eqnarray}
$\mathcal{P}_{N}^{\boldsymbol{\Theta}}$ is a bounding box for the $N$-th event. $\hat{\boldsymbol{\theta}}$ is the optimal parameter set that maximises $L_{N}$ over  the interval $\boldsymbol{\Theta}$. $\overline{I}^{N-1}(\mathbf{p}_{ij})$ is the value of pixel $\mathbf{p}_{ij}$ in the upper bound IWE, a recursively constructed image in which we always increment the maximum accumulator within the bounding box $\mathcal{P}_{N}^{\boldsymbol{\Theta}}$ (i.e. the one that we used to define the value of $Q^{N-1}$. The incremental construction of $\overline{I}^{N-1}(\mathbf{p}_{ij})$ is illustrated in Fig.~\ref{fig:iweConstruction}.
\label{the:upper bound}
}

\noindent \textbf{Proof 2.} 
(\ref{eq:opt_recursive_formulation}) is derived in the same manner as Theorem 1. The proof of inequation (\ref{eq:upper_bound}) then proceeds by mathematical induction. \\

For $N$ = 0, it is obvious that $L_{0} = \overline{L_{0}}= 0$.  Similarly, for $N$ = 1, $L_{1} = 1 \leq \overline{L_{0}} + 1 + 0$, and $Q^0 = I^{0}(\boldsymbol{\eta}_{1}^{\hat{\boldsymbol{\theta}}};\hat{\boldsymbol{\theta}}) = 0$ (which satisfies Theorem~2).  We now assume that $\overline{L_n}$ as well as the corresponding upper bound IWE $\overline{I}^{n}$ are given for all $0<n\leq N$. We furthermore assume that they satisfy Theorem~2. Our aim is to prove that \eqref{eq:upper_bound} holds for the $(N+1)$-th event. It is clear that $\overline{L_N} \geq L_N$, and we only need to prove that $Q^N \geq I^{N}(\boldsymbol{\eta}_{N+1}^{\hat{\boldsymbol{\theta}}};\hat{\boldsymbol{\theta}})$, for which we will make use of Lemma 1. There are two cases to be distinguished:
\begin{itemize}
    \item The first case is if there exists an event $\epsilon_k$ with $0 < k < N+1$ and for which $\boldsymbol{\eta}_{k}^{\hat{\boldsymbol{\theta}}} = \boldsymbol{\eta}_{N+1}^{\hat{\boldsymbol{\theta}}}$. In other words, the $k$-th and the $(N+1)$-th event are warped to a same accumulator if choosing the locally optimal parameters. Note that if there are multiple previous events for which this condition holds, the $k$-th event is chosen to be the most recent one. Given our assumptions, $\overline{L_{k-1}}$ as well as the $(k-1)$-th constructed upper bound IWE satisfy Theorem~2, which means that $Q^{k-1} \geq I^{k-1}(\boldsymbol{\eta}_{k}^{\hat{\boldsymbol{\theta}}};\hat{\boldsymbol{\theta}})$. Let $\mathbf{p}_k \in \mathcal{P}^{\boldsymbol{\Theta}}_k$ now be the pixel location with maximum intensity in $\overline{I}^{k-1}(\mathbf{p}_k)$. Then, the $k$-th updated IWE satisfies  $\overline{I}^k(\mathbf{p}_k) = Q^{k-1}+1 \geq I^{k-1}(\boldsymbol{\eta}_{k}^{\hat{\boldsymbol{\theta}}};\hat{\boldsymbol{\theta}}) +1 $. According to Lemma~1, we have $\mathcal{P}^{\boldsymbol{\Theta}}_k \subseteq \mathcal{P}^{\boldsymbol{\Theta}}_{N+1}$, therefore $\mathbf{p}_k \subseteq \mathcal{P}^{\boldsymbol{\Theta}}_{N+1}$, and $Q^N \geq \overline{I}^k(\mathbf{p}_k) \geq  I^{k-1}(\boldsymbol{\eta}_{k}^{\hat{\boldsymbol{\theta}}};\hat{\boldsymbol{\theta}})+1 $. With optimal warp parameters $\hat{\boldsymbol{\theta}}$, events with indices from $k+1$ to $N$ will not locate at $\boldsymbol{\eta}_{N+1}^{\hat{\boldsymbol{\theta}}} $
    , and therefore $I^{k-1}(\boldsymbol{\eta}_{k}^{\hat{\boldsymbol{\theta}}};\hat{\boldsymbol{\theta}})+1 = I^{N}(\boldsymbol{\eta}_{N+1}^{\hat{\boldsymbol{\theta}}};\hat{\boldsymbol{\theta}}) \leq Q^N $.
    \item If there is no such a event, it is obvious that $Q^N \geq  I^{N}(\boldsymbol{\eta}_{N+1}^{\hat{\boldsymbol{\theta}}};\hat{\boldsymbol{\theta}})$.
\end{itemize}
With the basic cases and the induction step proven, we conclude our proof that Theorem~2 holds for all natural numbers $N$.\\ 
\noindent \rule[5pt]{0.49\textwidth}{0.05em}

\subsection{Bounds for Other Objective Functions}
\label{sec:other_bounds}
We further apply the proposed strategy to derive upper and lower bounds for the other five aforementioned contrast functions including variance ($L_{\mathrm{Var}}$), sum of exponentials ($L_{\mathrm{SoE}}$), sum of suppressed accumulations ($L_{\mathrm{SoSA}}$), sum of exponentials and squares ($L_{\mathrm{SoEaS}}$) and sum of suppressed accumulations and squares ($L_{\mathrm{SoSAaS}}$). For convenience, we employ $L_N$ as a unified representation of each function.

\noindent \textbf{Variance (Var)}

The variance of an IWE is used in\cite{gallego2018unifying,gallego2017accurate} as an objective function. We define $\mu_{I} = N/N_p$ as the mean value of $I^{N}(\mathbf{p}_{ij};\boldsymbol{\theta})$ over all pixels, where $N$ is the number of events and $N_{p}$ is the total number of accumulators in an image plane. Hence $\mu_{I}$ may be approximated to be constant, which renders $L_{\mathrm{SoS}}$ and $L_{\mathrm{Var}}$ essentially equivalent (also implied in \cite{gallego2017accurate}). 
The function reads as follows:
\begin{equation}
    \begin{split}
        L_{N} 
        & =  \frac{1}{N_{p}}\sum_{\mathbf{p}_{ij}\in\mathcal{P}}(I^{N}(\mathbf{p}_{ij};\boldsymbol{\theta}) - \mu_{I})^2 \\
        & = \frac{1}{N_{p}} \sum_{\mathbf{p}_{ij}\in\mathcal{P}} \left[ \sum_{k=1}^{N} \boldsymbol{1}(\textbf{p}_{ij} -  W(\mathbf{x}_k,t_{k};\boldsymbol{\theta}) ) - \mu_I \right]^2  \\
        & = L_{N-1} + a + b + c,
    \end{split}
\end{equation}
where 
\begin{eqnarray}
         a &=& \frac{2}{N_{p}} \sum_{\mathbf{p}_{ij}\in\mathcal{P}} \left\{ I^{N-1}(\mathbf{p}_{ij};\boldsymbol{\theta}) \boldsymbol{1}(\textbf{p}_{ij} - W(\mathbf{x}_N,t_{N};\boldsymbol{\theta}) )  \right\} \nonumber \\
         b &=& - \frac{2  \mu_I}{N_{p}} \sum_{\mathbf{p}_{ij}\in\mathcal{P}}  \boldsymbol{1}(\textbf{p}_{ij} - W(\mathbf{x}_N,t_{N};\boldsymbol{\theta}) ) = - \frac{2  \mu_I}{N_{p}}  \nonumber \\
        c &=& \frac{1}{N_{p}} \sum_{\mathbf{p}_{ij}\in\mathcal{P}} \left[ \boldsymbol{1}(\textbf{p}_{ij} - W(\mathbf{x}_N,t_{N};\boldsymbol{\theta}) ) \right]^2 = \frac{1}{N_p}. 
\label{equ:ele_of_variance}
\end{eqnarray}

The term $\boldsymbol{1}(\mathbf{p}_{ij}-W(\mathbf{x}_{N},t_{N};\boldsymbol{\theta}))$ is simply zero unless we are considering an accumulator $\mathbf{p}_{ij} = \boldsymbol{\eta}_{N}^{\boldsymbol{\theta}}$, which makes $b$ and $c$ in equation (\ref{equ:ele_of_variance}) constant. 
Thus, given a search space $\boldsymbol{\Theta}$ centered at $\boldsymbol{\theta}_{0}$, the lower bound is
\begin{equation}
        \underline{L_{N}} = \underline{L_{N-1}} + \frac{1}{N_{p}} - \frac{2 \mu_{I}}{N_{p}} + \frac{2}{N_{p}}I^{N-1} (\boldsymbol{\eta}_N^{\boldsymbol{\theta}_0};\boldsymbol{\theta}_0). 
\end{equation}
Furthermore, according to Theorem 2, the upper bound becomes
\begin{equation}
        \overline{L_{N}} = \overline{L_{N-1}} + \frac{1}{N_{p}} - \frac{2 \mu_{I}}{N_{p}} + \frac{2}{N_{p}} Q^{N-1}.
\end{equation}
Note that the initial upper and lower bounds $\overline{L_0} = \underline{L_0}=L_0=\mu_{I}^2$.

\noindent \textbf{Sum of Exponentials (SoE)}

The sum of exponentials objective reads
\begin{equation}
\begin{split}
        L_{N} 
        & = \sum_{\mathbf{p}_{ij}\in\mathcal{P}}e^{I^N(\mathbf{p}_{ij};\boldsymbol{\theta})} \\
        & = \sum_{\mathbf{p}_{ij}\in\mathcal{P}}e^{\sum_{k=1}^{N} \boldsymbol{1}(\textbf{p}_{ij} -  W(\mathbf{x}_k,t_{k};\boldsymbol{\theta}) )}.
\end{split}
\end{equation}
It sums up the exponential intensity at each pixel in the IWE. Note that the sum of exponentials is generally much bigger than $L_{\mathrm{Var}}$ and $L_{\mathrm{SoS}}$. We leverage a similar recursive strategy to calculate the loss 
\begin{eqnarray}
    L_{N} 
        & =& \sum_{\mathbf{p}_{ij}\in\mathcal{P}}e^{\sum_{k=1}^{N-1} \boldsymbol{1}(\textbf{p}_{ij} -  W(\mathbf{x}_k,t_{k};\boldsymbol{\theta}) ) + \boldsymbol{1}(\mathbf{p}_{ij} - W(\mathbf{x}_N,t_{N};\boldsymbol{\theta}) ) } \nonumber \\
        & =& \sum_{\mathbf{p}_{ij}\in\mathcal{P}}e^{\sum_{k=1}^{N-1} \boldsymbol{1}(\textbf{p}_{ij} -  W(\mathbf{x}_k,t_{k};\boldsymbol{\theta}) )} e^{ \boldsymbol{1}(\textbf{p}_{ij} - W(\mathbf{x}_N,t_{N};\boldsymbol{\theta}) ) }\nonumber \\
        &=& \sum_{\substack{\mathbf{p}_{ij}\in\mathcal{P} \\ \mathbf{p}_{ij} \neq \boldsymbol{\eta}_N^{\boldsymbol{\theta}}}} e^{I^{N-1}(\mathbf{p}_{ij};\boldsymbol{\theta})} + a \nonumber \\
  &=& \sum_{\mathbf{p}_{ij}\in\mathcal{P}} e^{I^{N-1}(\mathbf{p}_{ij};\boldsymbol{\theta})} - e^{I^{N-1}(\boldsymbol{\eta}_N^{\boldsymbol{\theta}};\boldsymbol{\theta})} + a \nonumber \\
  &=& L_{N-1} - e^{I^{N-1}(\boldsymbol{\eta}_N^{\boldsymbol{\theta}};\boldsymbol{\theta})} + a,        
\end{eqnarray}
where, according to the property of the indicator function, 
\begin{eqnarray}
         a &=& e^{I^{N-1}(\boldsymbol{\eta}_N^{\boldsymbol{\theta}};\boldsymbol{\theta})} e^{ \boldsymbol{1}(\boldsymbol{\eta}_N^{\boldsymbol{\theta}} - W(\mathbf{x}_N,t_{N};\boldsymbol{\boldsymbol{\theta}}) ) } \nonumber \\
        & = & e \cdot e^{I^{N-1}(\boldsymbol{\eta}_N^{\boldsymbol{\theta}};\boldsymbol{\theta})}.  \nonumber 
\end{eqnarray}
Thus, using a similar strategy than in Theorem 1 and Theorem 2, the recursive bounds become
\begin{equation}
    \begin{split} 
      & \underline{L_{N}} = \underline{L_{N-1}} + (e-1) e^{I^{N-1}(\boldsymbol{\eta}_N^{\boldsymbol{\theta}_0};\boldsymbol{\theta}_0)} \\
        & \overline{L_{N}} = \overline{L_{N-1}} + (e-1) e^{Q} .
    \end{split}
\end{equation}
Note that the initial upper and lower bounds $\overline{L_0} = \underline{L_0}=L_0=N_p$.

\noindent \textbf{Sum of Suppressed Accumulations (SoSA)}

For $L_{\mathrm{SoSA}}$, $\delta$ is a design parameter called the \textit{shift factor}. Different from other objective functions, locations with few accumulations will contribute more to the $L_{\mathrm{SoSA}}$. The intuition here is that more empty locations again mean more events that are concentrated at fewer accumulators, and thus a higher-contrast IWE. The focuss loss function reads
\begin{equation}
  \begin{split}
        L_{N}
        & = \sum_{\mathbf{p}_{ij}\in\mathcal{P}}e^{-I(\mathbf{p}_{ij};\boldsymbol{\theta})\cdot\delta} \\
        & = \sum_{\mathbf{p}_{ij}\in\mathcal{P}}e^{ -\delta \cdot \sum_{k=1}^{N} \boldsymbol{1}(\textbf{p}_{ij} -  W(\mathbf{x}_k,t_{k};\boldsymbol{\theta}) ) }. \\
    \end{split}
\end{equation}
The derivation is analogous to the derivation for the sum of exponentials, and leads to 
\begin{eqnarray}
L_{N}
        & =& \sum_{\mathbf{p}_{ij}\in\mathcal{P}} e^{-\delta \cdot I^{N-1}(\mathbf{p}_{ij};\boldsymbol{\theta}) -\delta \cdot \boldsymbol{1}(\textbf{p}_{ij} - W(\mathbf{x}_N,t_{N};\boldsymbol{\theta}) ) } \nonumber \\
        & =& L_{N-1}(\boldsymbol{\theta}) - e^{ -\delta \cdot I^{N-1}(\boldsymbol{\eta}_N^{\boldsymbol{\theta}};\boldsymbol{\theta})} + a,
\end{eqnarray}
where
\begin{eqnarray}
        a &=& e^{ -\delta \cdot I^{N-1}(\boldsymbol{\eta}_N^{\boldsymbol{\theta}};\boldsymbol{\theta})} e^{  -\delta \cdot \boldsymbol{1}(\boldsymbol{\eta}_N^{\boldsymbol{\theta}} - W(\mathbf{x}_N,t_{N};\boldsymbol{\theta}) ) }  \nonumber \\
        &=& e^{-\delta} e^{ -\delta \cdot I^{N-1}(\boldsymbol{\eta}_N^{\boldsymbol{\theta}};\boldsymbol{\theta})}.  \nonumber
\end{eqnarray}
Thus, the bounds are
\begin{equation}
    \begin{split} 
    & \underline{L_{N}} = \underline{L_{N-1}} + (e^{-\delta}-1) e^{-\delta \cdot I^{N-1}(\boldsymbol{\eta}_N^{\boldsymbol{\theta}_0};\boldsymbol{\theta}_0)} \\
        & \overline{L_{N}} = \overline{L_{N-1}} + (e^{-\delta}-1) e^{-\delta \cdot Q}. \\
    \end{split}
\end{equation}
Note that the initial upper and lower bounds $\overline{L_0} = \underline{L_0}=L_0=N_p$.

\noindent \textbf{Sum of Exponentials and Squares (SoEaS)}

$L_{\mathrm{SoEaS}}$ is actually a hybrid function, which is a weighted sum of $L_{\mathrm{SoE}}$ and $L_{\mathrm{SoS}}$. We therefore have 
\begin{eqnarray}
  L_{N} &=& w_1 L_{\mathrm{SoE}} + w_2 L_{\mathrm{SoS}},
\end{eqnarray}
where $w_1$ and $w_2$ are linear combination weights.
The equation can still be simplified into recursive form, which leads us to
\begin{eqnarray}
  L_{N} &=& w_1 \left[ L_{N-1}^{\mathrm{SoE}} + (e-1) e^{I^{N-1}(\boldsymbol{\eta}_N^{\boldsymbol{\theta}};\boldsymbol{\theta})} \right] \nonumber \\
        &+& w_2 \left[ L_{N-1}^{\mathrm{SoS}}+1+2 I^{N-1}(\boldsymbol{\eta}_{N}^{\boldsymbol{\theta}};\boldsymbol{\theta}) \right] \nonumber \\
        &=& L_{N-1} +  w_1 (e-1) e^{I^{N-1}(\boldsymbol{\eta}_N^{\boldsymbol{\theta}};\boldsymbol{\theta})} \nonumber \\
        &+& w_2 +2 w_2 I^{N-1}(\boldsymbol{\eta}_{N}^{\boldsymbol{\theta}};\boldsymbol{\theta})).
\end{eqnarray}
It follows immediately that the bounds are given as
\begin{eqnarray}
  \underline{L_{N}} 
        &=& \underline{L_{N-1}} +  w_1 (e-1) e^{I^{N-1}(\boldsymbol{\eta}_N^{\boldsymbol{\theta}_0};\boldsymbol{\theta}_0)} \nonumber \\
        &+& w_2 +2 w_2 I^{N-1}(\boldsymbol{\eta}_{N}^{\boldsymbol{\theta}_0};\boldsymbol{\theta}_0)) \nonumber \\
  \overline{L_{N}} &=& \overline{L_{N-1}} +  w_1 (e-1) e^Q + w_2 +2 w_2 Q .
\end{eqnarray}
Note that the initial upper and lower bounds $\overline{L_0} = \underline{L_0}=L_0=w_1 N_p$. In this paper, we choose $w_1 = 1.0$ and $w_2 = 1.0$.

\noindent \textbf{Sum of Suppressed Accumulations and Squares (SoSAaS)}
$L_{\mathrm{SoSAaS}}$ is another hybrid function, which is a weighted sum of $L_{\mathrm{SoSA}}$ and $L_{\mathrm{SoS}}$, i.e.
\begin{eqnarray}
  L_{N} &=& w_1 L_{\mathrm{SoSA}} + w_2 L_{\mathrm{SoS}}.
\end{eqnarray}
$w_1$ and $w_2$ are again linear combination weights. The recursive form is given by
\begin{eqnarray}
  L_{N} &=& w_1 \left[ L_{N-1}^{\mathrm{SoSA}} + (e^{-\delta}-1) e^{-\delta I^{N-1}(\boldsymbol{\eta}_N^{\boldsymbol{\theta}};\boldsymbol{\theta})} \right] \nonumber \\
        &+& w_2 \left[ L_{N-1}^{\mathrm{SoS}}+1+2 I^{N-1}(\boldsymbol{\eta}_{N}^{\boldsymbol{\theta}};\boldsymbol{\theta}) \right] \nonumber \\
        &=& L_{N-1} +  w_1 (e^{-\delta}-1) e^{-\delta \cdot I^{N-1}(\boldsymbol{\eta}_N^{\boldsymbol{\theta}};\boldsymbol{\theta})} \nonumber \\
        &+& w_2 +2 w_2 I^{N-1}(\boldsymbol{\eta}_{N}^{\boldsymbol{\theta}};\boldsymbol{\theta})).
\end{eqnarray}
It again immediately follows that the bounds are given by
\begin{eqnarray}
  \underline{L_{N}} 
        &=& \underline{L_{N-1}} +  w_1 (e^{-\delta}-1) e^{-\delta \cdot I^{N-1}(\boldsymbol{\eta}_N^{\boldsymbol{\theta}_0};\boldsymbol{\theta}_0)}  \nonumber \\
        &+& w_2 +2 w_2 I^{N-1}(\boldsymbol{\eta}_{N}^{\boldsymbol{\theta}_0};\boldsymbol{\theta}_0)) \nonumber \\
  \overline{L_{N}} &=& \overline{L_{N-1}} +  w_1 (e^{-\delta}-1) e^{-\delta Q} + w_2 +2 w_2 Q .
\end{eqnarray}
Note that the initial upper and lower bounds $\overline{L_0} = \underline{L_0}=L_0=w_1 N_p$. In this paper, we set $w_1 = 1.0$ and $w_2 = 1.0$.

The derived upper and lower bounds for all aforementioned six loss functions are listed and summarized in Table~\ref{tab:upper_bounds}. Note that the initial case varies depending on the considered loss functions. 

\subsection{Algorithm}

Our complete globally-optimal contrast maximisation framework~(GOCMF) is outlined in Algorithm~\ref{alg:GOCMF} and Algorithm~\ref{alg:RB}. We propose a nested strategy for calculating upper bounds, in which the outer layer $\mathbf{RB}$ evaluates the objective function, while the inner layer $\mathbf{BB}$ estimates the bounding box $\mathcal{P}_{N}^{\boldsymbol{\Theta}}$ and depends on the specific motion parametrisation.

\begin{algorithm}[H]
  \caption{\textbf{GOCMF}: globally optimal contrast maximisation framework} 
  {\bf Input:} 
  event set $\mathcal{E}$, initial search space $\boldsymbol{\Theta}$, termination threshold $\tau$\\
  {\bf Output:} 
  optimal warping parameters $\hat{\boldsymbol{\theta}}$
  \begin{algorithmic}[1]
    \State Initialise $\boldsymbol{\theta}_0$ with the center of $\boldsymbol{\Theta}$
    \State $\hat{\boldsymbol{\theta}} \leftarrow \boldsymbol{\theta}_0$
    \State Initialise priority queue $Q$
    \State $\{\overline{L},\underline{L}\} \leftarrow \mathbf{RB}(\mathcal{E}$, $\boldsymbol{\Theta}$), $\hat{L} \leftarrow \underline{L}$
    \State Push $\boldsymbol{\Theta}$ into $Q$ with priority $ \overline{L}$
    \While{$Q$ is not empty} 
      \State Pop $\boldsymbol{\Theta}$ form $Q$
                   \If{$\overline{L} - \underline{L} \leq \tau$,} terminate
                   \EndIf                                           \State $\boldsymbol{\boldsymbol{\theta}}_0 \leftarrow$ Center of $\boldsymbol{\Theta}$
          \If{ $\underline{L} \geq \hat{L}$,} $\hat{\boldsymbol{\theta}} \leftarrow \boldsymbol{\theta}_0,\ \hat{L} \leftarrow \underline{L}$                 \EndIf
        \State Subdivide $\boldsymbol{\Theta}$ into subspaces $\boldsymbol{\Theta}_j$
        \For{all subspaces $\boldsymbol{\Theta}_j$}
          \State $\{\overline{L},\underline{L}\} = \mathbf{RB}(\mathcal{E}, \boldsymbol{\Theta}_j)$
          \If {$\overline{L} \geq \hat{L}$} Insert $\boldsymbol{\Theta}_j$ into $Q$ with priority $\overline{L}$

          \EndIf
        \EndFor
    \EndWhile
    \State \Return $\hat{\boldsymbol{\theta}}$
  \end{algorithmic}
  \label{alg:GOCMF}
 \end{algorithm}
\begin{algorithm}[H]
  \caption{\textbf{RB}: recursive bounds calculation} 
  {\bf Input:} 
  event set $\mathcal{E}$, search space $\boldsymbol{\Theta}$\\
  {\bf Output:} 
  lower bound $\underline{L}$, upper bound $\overline{L}$
  \begin{algorithmic}[1]
    \State Initialise accumulator image matrices $\overline{I}$ and $\underline{I}$ with zeros
    \State Initialise $\underline{L}$, $\overline{L}$ according to Table~\ref{tab:upper_bounds}
    \State $\boldsymbol{\theta}_0 \leftarrow$ center of $\boldsymbol{\Theta}$
    \For{each event $e_k \in \mathcal{E}$}
      \State $\mathcal{P}_k^{\boldsymbol{\Theta}} \leftarrow \mathbf{BB}(W(\cdot),\boldsymbol{\Theta},e_k) $
      \State $Q = \max_{\substack{
\mathbf{p}_{ij} \in \mathcal{P}_{k}^{\boldsymbol{\Theta}}
}} \overline{I}(\mathbf{p}_{ij})$
      \State $\boldsymbol{\eta}_k^{\boldsymbol{\theta}_0} = \operatorname{round}(W(\mathbf{x}_{k},t_{k};\boldsymbol{\theta}_{0}))$
      \State $\boldsymbol{\nu}_k = \operatorname{argmax}_{\substack{
\mathbf{p}_{ij} \in \mathcal{P}_{k}^{\boldsymbol{\Theta}}
}} \overline{I}(\mathbf{p}_{ij})$
      \State Update $\underline{L}$, $\overline{L}$ (cf. Table~\ref{tab:upper_bounds})
      \State $\overline{I}(\boldsymbol{\nu}_k) = \overline{I}(\boldsymbol{\nu}_k)+ 1$
      \State $\underline{I}(\boldsymbol{\eta}_k^{\boldsymbol{\theta}_0}) = \underline{I}(\boldsymbol{\eta}_k^{\boldsymbol{\theta}_0})+ 1$ 
    \EndFor
    \State \Return $\underline{L}$, $\overline{L}$
  \end{algorithmic}
\label{alg:RB}
\end{algorithm}

\section{Applications}
\label{sec:casestudy}

As introduced by Gallego et al. \cite{gallego2018unifying}, contrast maximisation can be applied to several event-based vision problems. We first apply GOCMF to a simple optical flow estimation problem (Section~\ref{sec:optical_flow}). We then apply it to fronto-parallel motion estimation (Section~\ref{sec:downward-facing}) in front of noisy or feature-poor, fast-moving textures, and demonstrate the potential of outperforming regular camera alternatives. We finally apply our framework to the problem of camera rotation estimation, where we compare our algorithm against the recently proposed, alternative globally-optimal framework by Liu et al. \cite{liu2020globally} (Section~\ref{sec:Rotational}).

\subsection{Optical Flow Estimation}
\label{sec:optical_flow}

Optical flow plays a vital role in object tracking, image registration, visual odometry and other navigation tasks \cite{aires2008optical}. Horn-Schunck~\cite{horn1981determining} and Lucas-Kanade~\cite{lucas1981iterative} are classical algorithms for optical flow estimation with  standard cameras. Contrast maximisation methods for event cameras are good alternatives to estimate optical flow in challenging scenarios in which regular cameras would suffer from motion blur.  
We start by applying our globally optimal framework (Algorithm \ref{alg:GOCMF}) to event-based optical flow estimation. More specifically, the goal is to estimate a two-dimensional velocity vector for a given point in the image plane by considering the contrast in a small bounding box around that point. This first-order model simply assumes that the trajectory of a pixel in the image plane is a straight, constant-velocity line over short time intervals. Complete flow fields can be computed by repeating the operation for each location in the image plane.

The warping function that takes events back to the reference view is hence given by
\begin{equation}
  W(\mathbf{x}_k,t_{k}; \mathbf{v}) = {\mathbf{x}}'_k = \mathbf{x}_k + \mathbf{v} t_k,
\end{equation}
where $\mathbf{v} = [v^x\text{ }v^y]^\mathsf{T}$ is the velocity (optical flow) at the considered point, $\mathbf{x}_k$ the location where the $k$-th event occurred, and $t_k$ the elapsed time since the time of the reference view.

It is intuitively clear that---given a branch of the search space $\bold{\mathcal{V}} = [v^x_{\text{min}},v^x_{\text{max}}] \times [v^y_{\text{min}},v^y_{\text{max}}]$---the warped event may locate in a rectangular bounding box defined by
\begin{eqnarray}
 \underline{x_k^{\prime}} &=& x_k + v^x_{\text{min}} t_k ,\ \underline{y_k^{\prime}} = y_k + v^y_{\text{min}} t_k \nonumber \\ 
 \overline{x_k^{\prime}} &=& x_k + v^x_{\text{max}} t_k ,\ \overline{y_k^{\prime}} = y_k + v^y_{\text{max}} t_k .
\end{eqnarray}
We can then employ the derived bounding box to GOCMF to explore the globally optimal optical flow. Note that the objective function we used in the experiments is $L_{\mathrm{SoS}}$ which is a common loss used in previous contrast maximisation work.

\subsubsection{Results and Discussion}

We test our globally optimal optical flow estimation algorithm on the two sequences \textit{Circle} and \textit{Line} collected by ourselves with a downward-facing event camera mounted on an Autonomous Ground Vehicle (AGV) (cf. Fig.~\ref{fig:robot and frame}).

Ground truth optical flow is obtained by using the ground truth camera motion parameters, and calculating the first-order differential of the image motion. We find the optical flow at a image point by considering the contrast within a $40 \times 40$ pixel patch centered around that point. We thus assume that the optical flow of all the pixels in a patch is identical. We divide the \textit{Circle} and \textit{Line} sequences into sub-sequences of 0.04s.

\begin{figure}
\centering
\subfigure
{
\includegraphics[width=0.48\textwidth , height=0.15\textwidth]{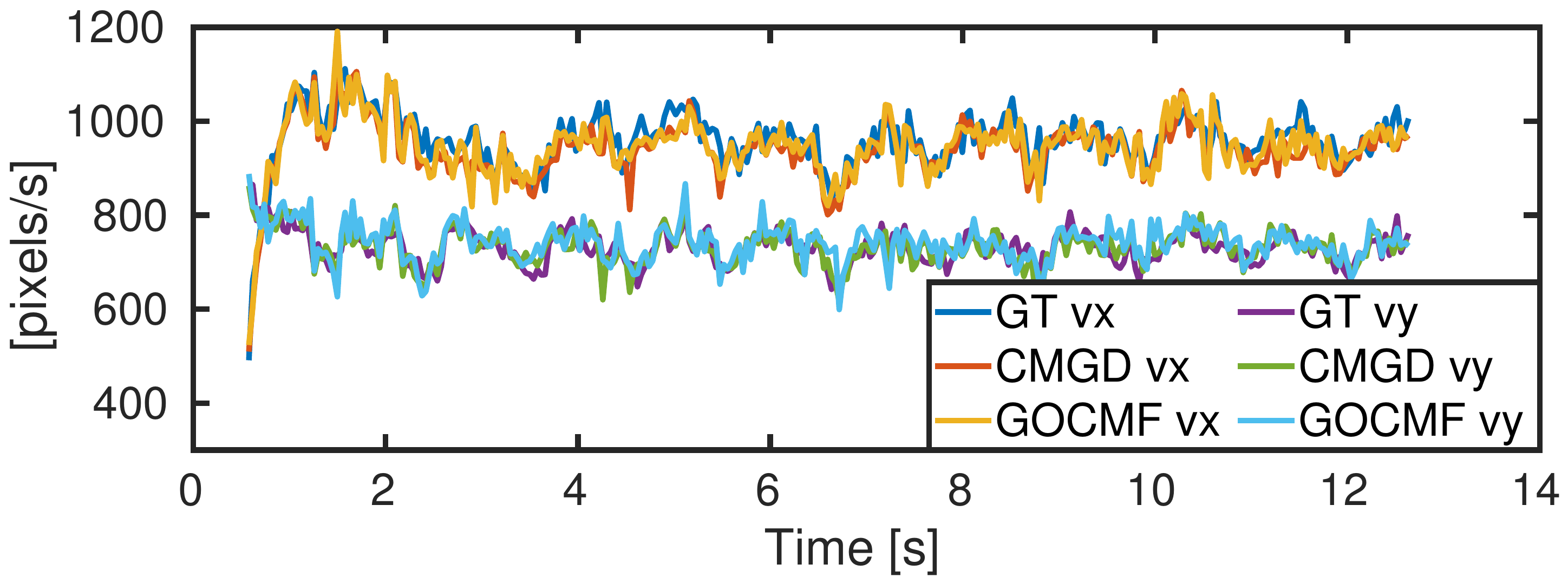}
}
\subfigure
{
\includegraphics[width=0.48\textwidth , height=0.15\textwidth]{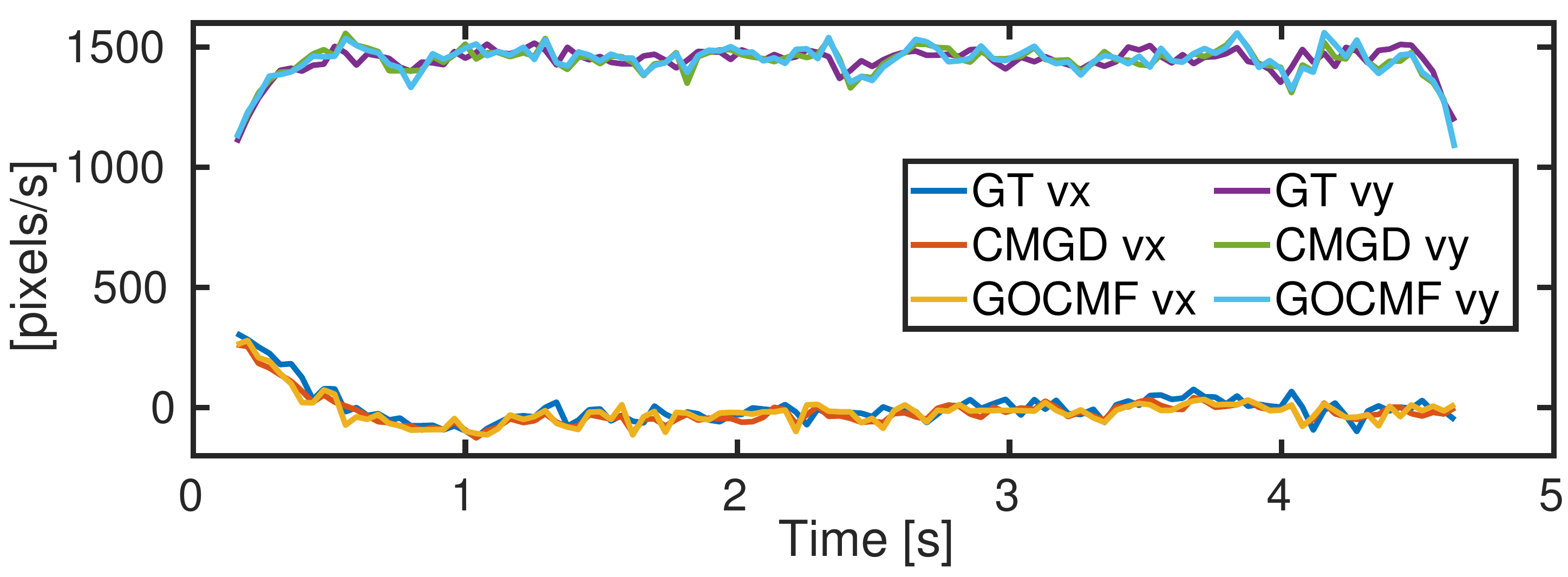}
}
\caption{Estimated pixel velocities compared against ground truth on sequences \textit{Circle} (top) and \textit{Line} (bottom). 
}
\label{fig:optical_flow_planar}
\end{figure}

We compare the proposed GOCMF with CMGD (local optimisation method using Matlab's \texttt{fmincon} function). Fig.~\ref{fig:optical_flow_planar} displays the optical flow estimated by GOCMF and CMGD as well as ground-truth. The estimated results match well with ground-truth. 
Quantitative results are exhibited in Table~\ref{tab:opticalflow_estimated_errors} showing the Average Endpoint Error ($AEE = \frac{1}{N}\sum \|\bold{x}^{\prime} - \bold{x}^{\prime *}\|_2$ with $\bold{x}^{\prime}$ and $\bold{x}^{\prime*}$ being the warped location with ground truth and estimated optical flows, and $N$ being the number of events) and the runtimes. GOCMF is more accurate than the locally optimising CMGD, while CMGD runs faster than our method.

\begin{table}[t]
\centering
\caption{Runtimes and Average Endpoint Errors (AEE) for GOCMF and CMGD}
\label{tab:opticalflow_estimated_errors}
\renewcommand\arraystretch{2}
\begin{tabular}{ccccc}
\toprule
\multirow{2}{*}{\textbf{Method}} & \multicolumn{2}{c}{\textit{Cicle}} & \multicolumn{2}{c}{\textit{Line}} \\ 
\cline{2-5} &  AEE  &  time [s]  & AEE & time [s] \\ \midrule
\textbf{CMGD}    &1.22      &12.12      &1.12   &9.59 \\ 
\textbf{GOCMF}  &1.15   &34.23    &0.98   &28.73      \\ 
\bottomrule
\end{tabular}
\end{table}

\begin{figure}[t]
\centering
\subfigure[Two patches on the reference view]
{
\label{fig:opticalflow_reference_view}
\begin{minipage}{0.26\textwidth}
\centering
\includegraphics[width=\textwidth]{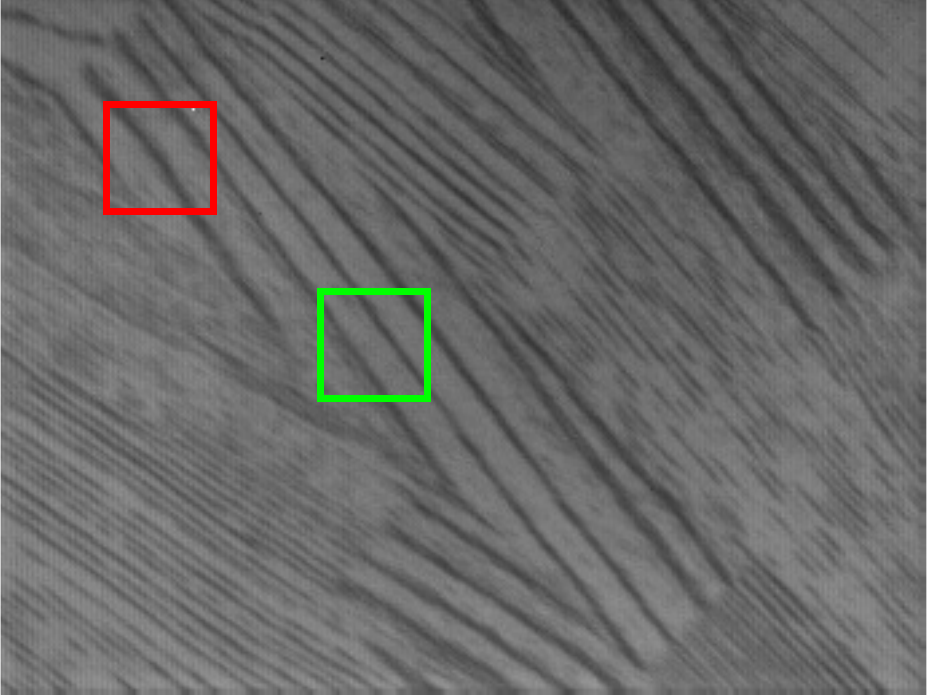}
\vspace{0.1cm}
\end{minipage}
}
\subfigure[Warped events]
{
\label{fig:opticalflow_warped_events}
\begin{minipage}{0.18\textwidth}
\centering
{
\includegraphics[width=0.48\textwidth]{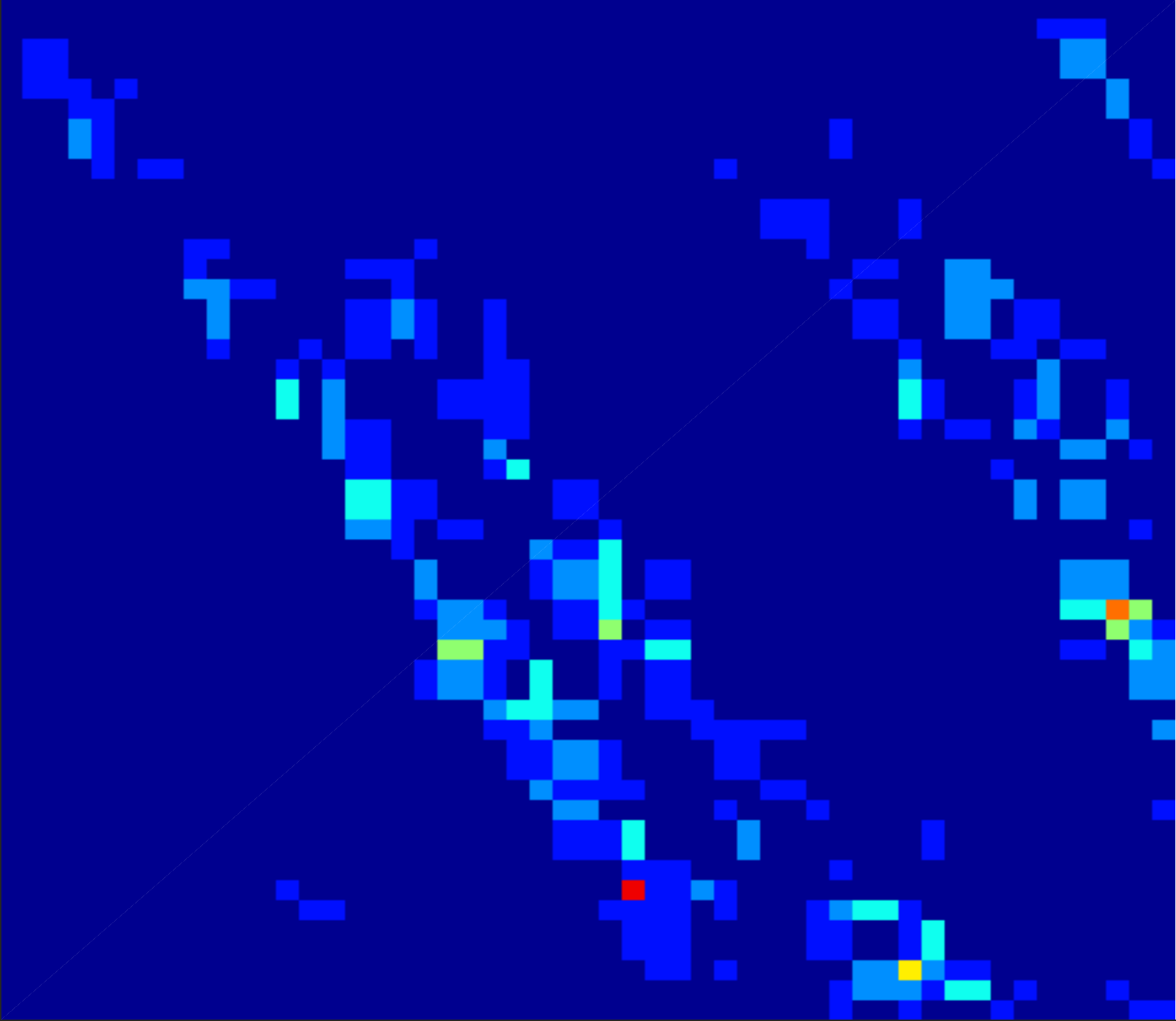}
\includegraphics[width=0.48\textwidth]{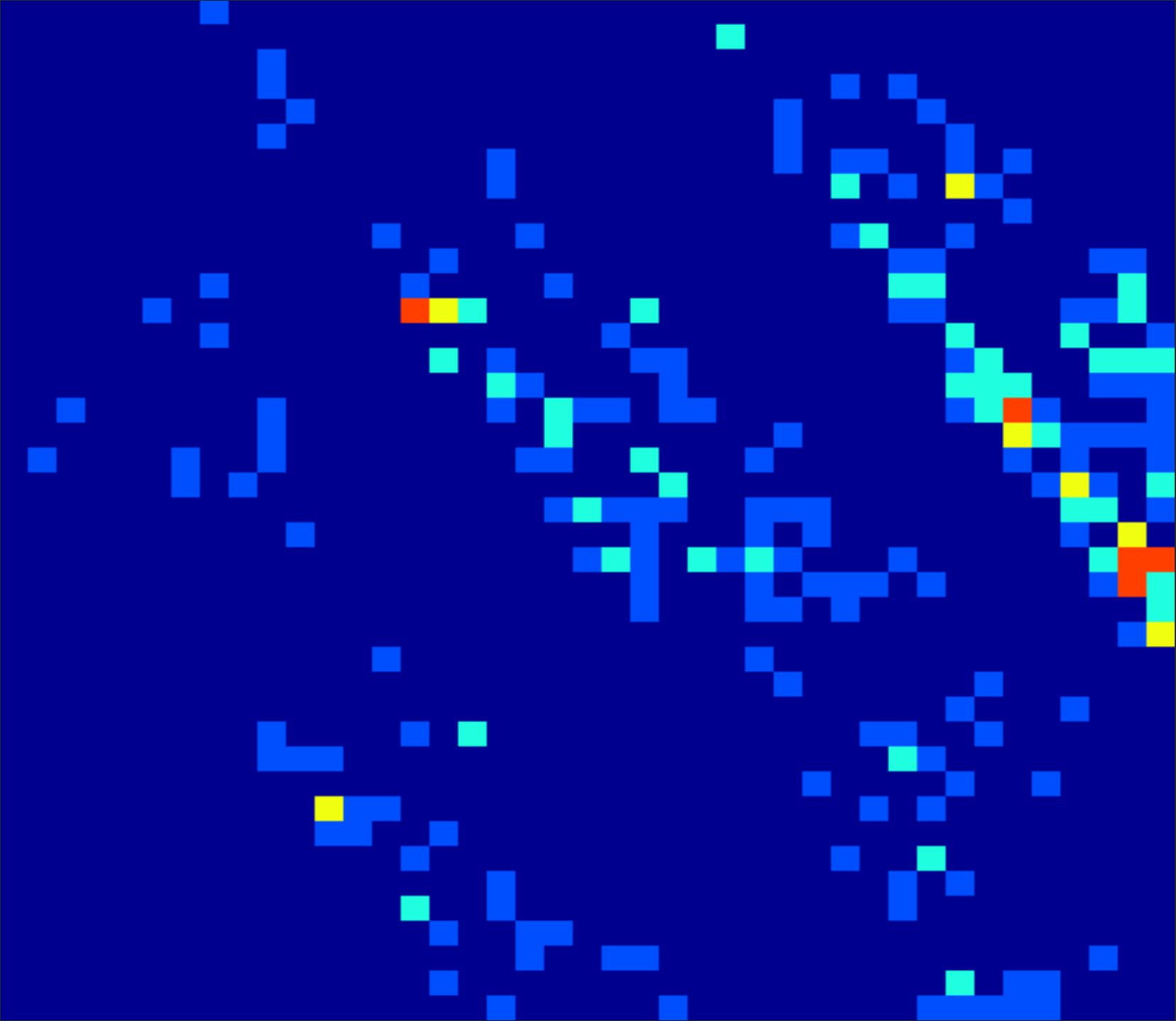} 
\centerline{GOCMF} 
}
{
\includegraphics[width=0.48\textwidth]{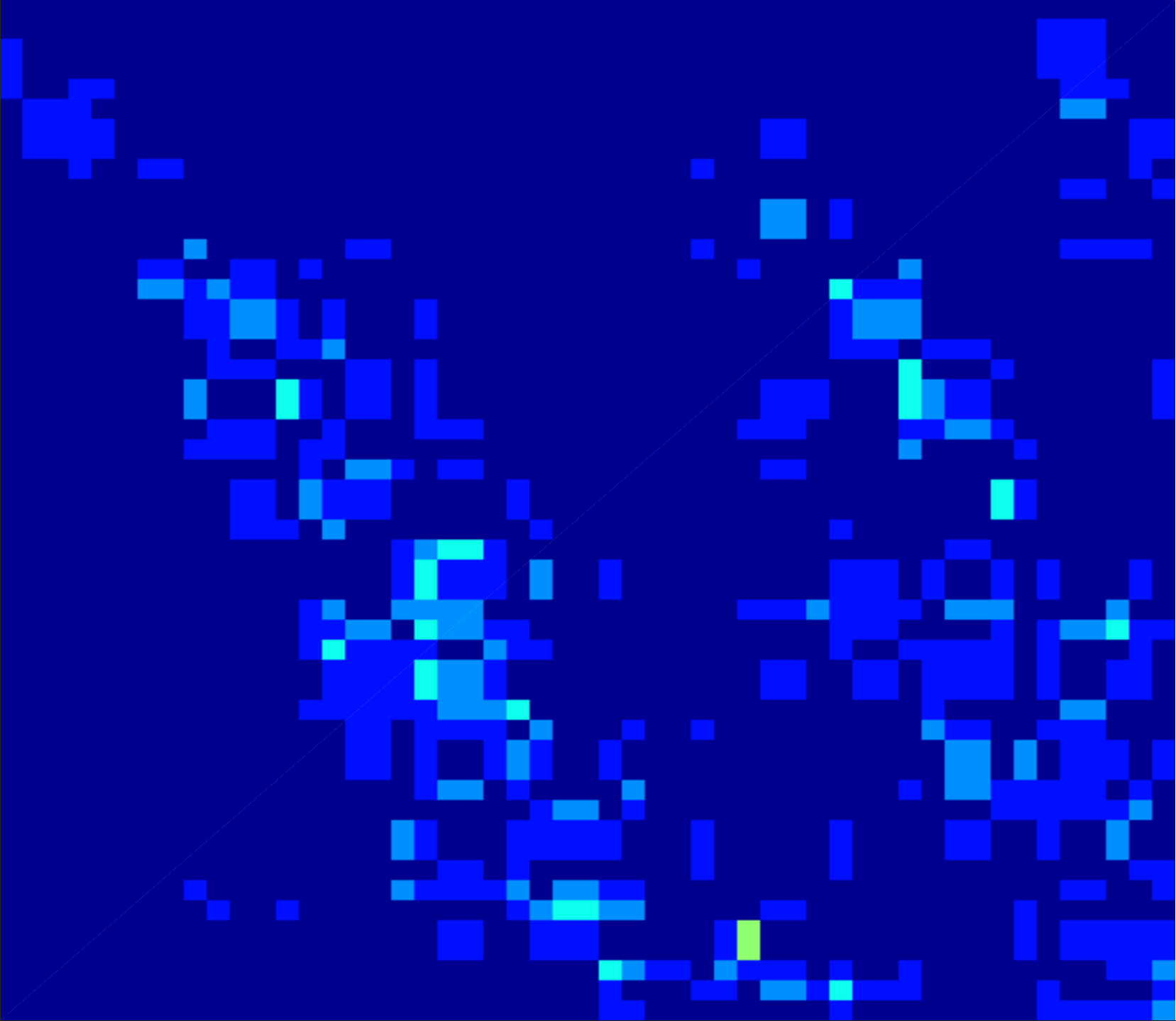}
\includegraphics[width=0.48\textwidth]{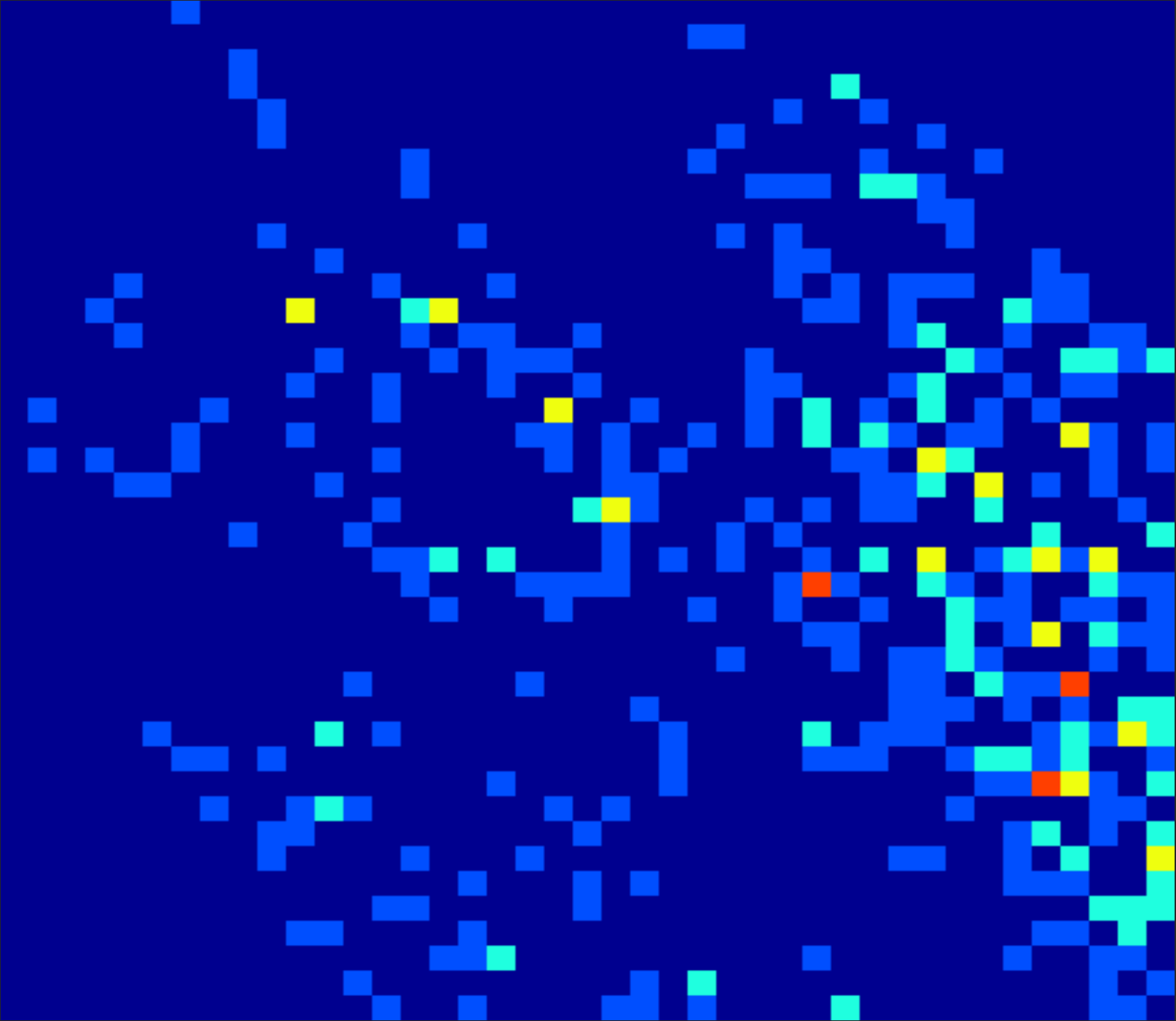}
\centerline{CMGD}
}
\vspace{0.1cm}
\end{minipage}
}
\caption{(a) Frame captured at the reference time and two patches (green and red) for optical flow estimation. (b) Images of warped events with optical flow estimated by GOCMF (top) and CMGD (bottom). Our method finds global optima and leads to significantly sharper IWEs than CMGD.}
\label{fig:optical_flow_rotational}
\end{figure}

Qualitative results are shown in Fig.~\ref{fig:optical_flow_rotational}. It illustrates the frame captured at the reference time and the images of warped events by GOCMF and CMGD. It is clear that IWEs with parameters estimated by GOCMF are much sharper than IWEs generated by CMGD.

\subsection{Visual odometry with a downward-facing event camera}
\label{sec:downward-facing}

\begin{figure}[t]
\centering
\subfigure[AGV]
{
\hspace{0.005\textwidth}
\includegraphics[width=0.21\textwidth, height=0.17\textwidth]{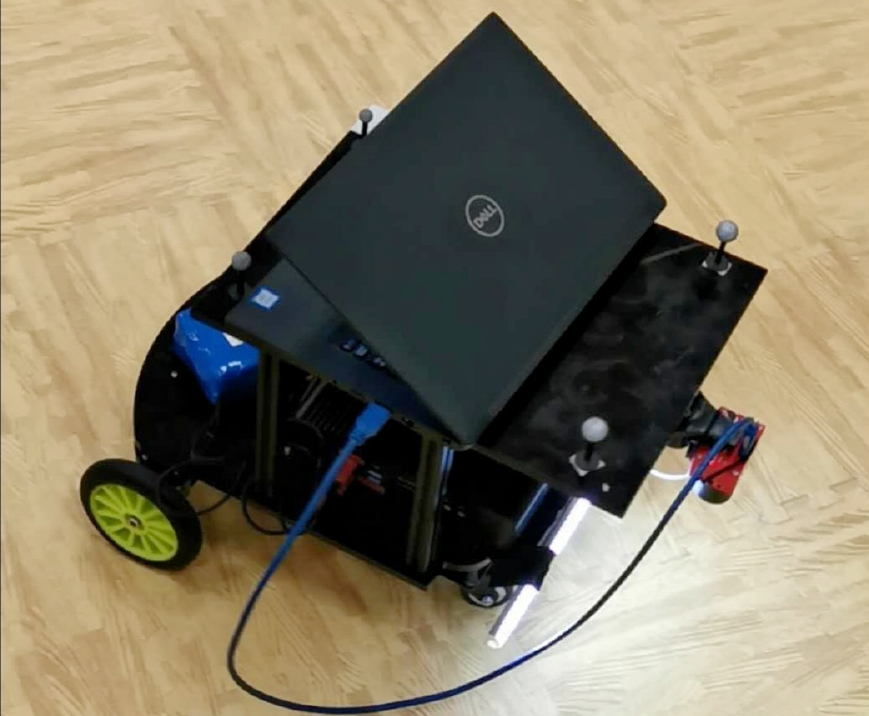}
\label{fig:AGV}
}
\hspace{0.005\textwidth}
\subfigure[wood grain foam]{
  \includegraphics[width=0.23\textwidth , height=0.17\textwidth]{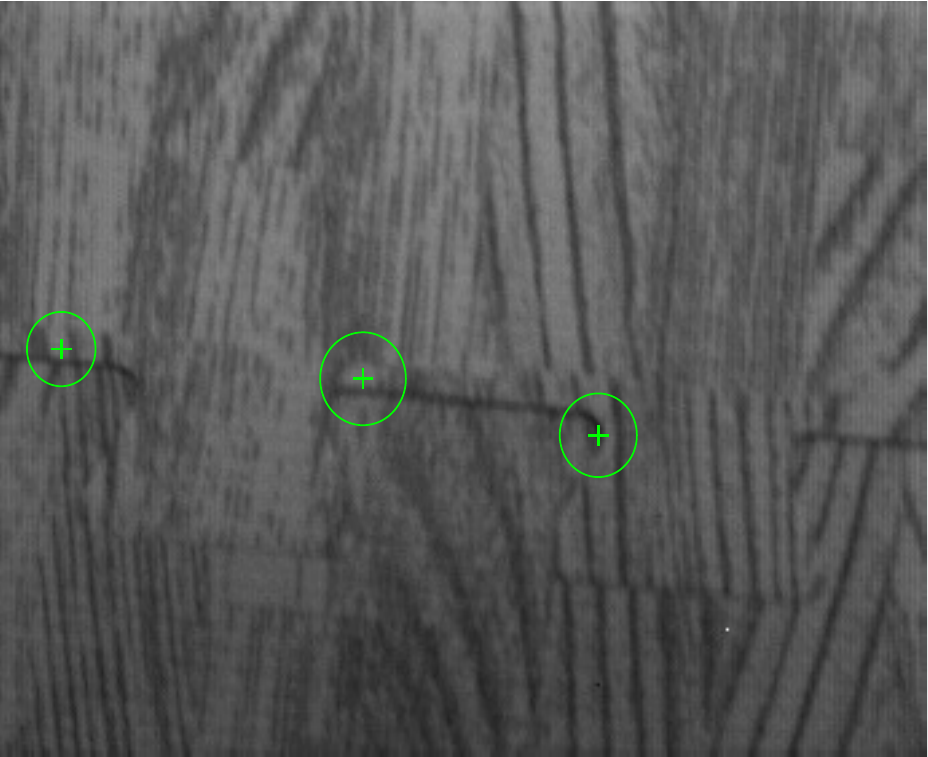} }
\subfigure[$\boldsymbol{\theta} = \mathbf{0}$]{
  \includegraphics[width=0.23\textwidth]{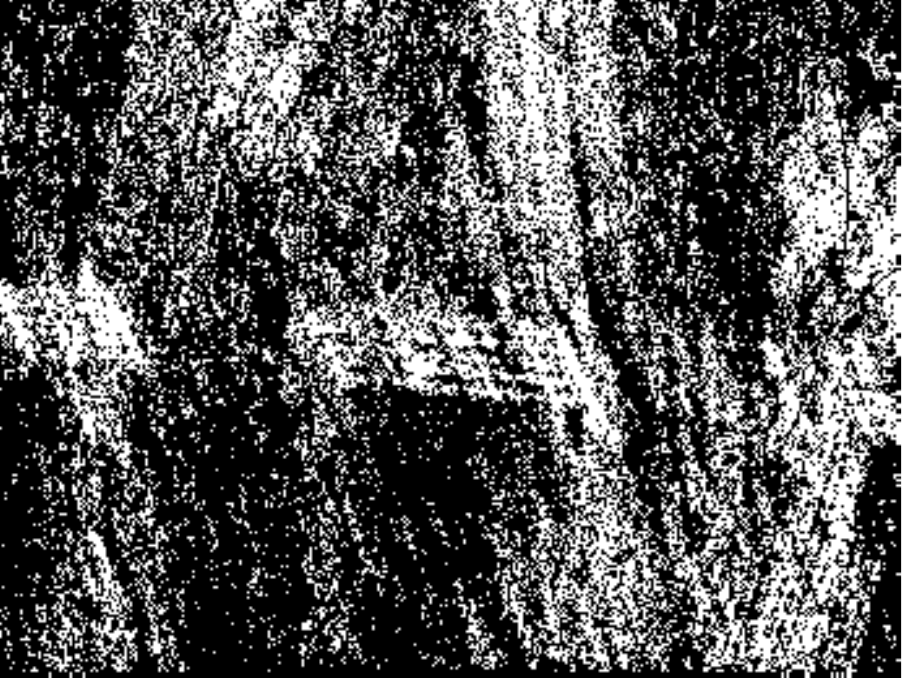}}
\subfigure[$\boldsymbol{\theta} = \hat{\boldsymbol{\theta}}$]{
  \includegraphics[width=0.23\textwidth]{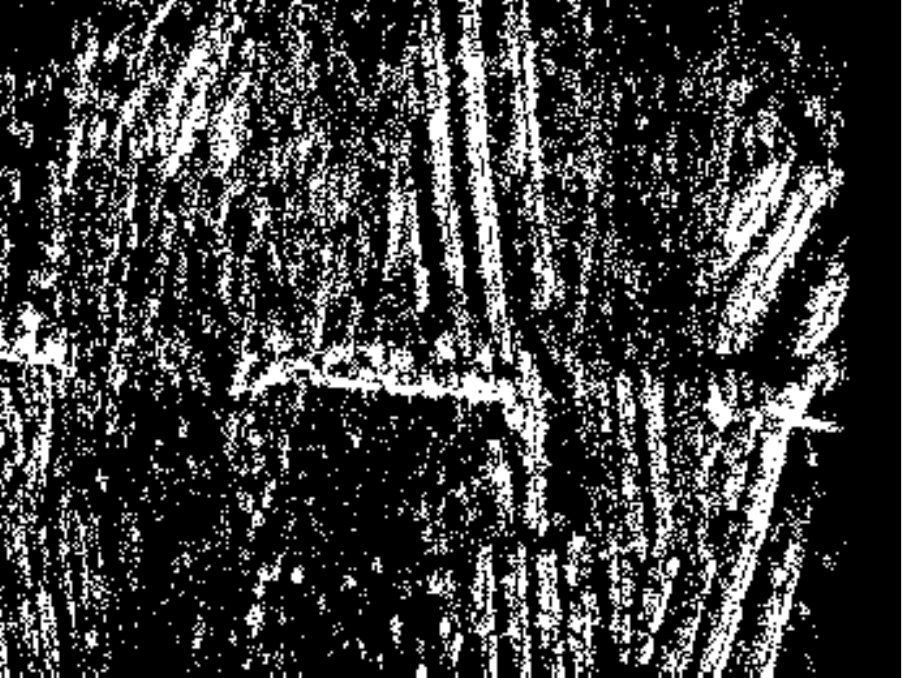}} 
\caption{(a) AGV equipped with a downward facing event camera for vehicle motion estimation . (b) collected image with detectable corners, (c) image of warped events with $\boldsymbol{\theta} = \mathbf{0}$, and (d) image of warped events with optimal parameters.}
\label{fig:robot and frame}
\end{figure}

Motion estimation for planar Autonomous Ground Vehicles~(AGVs) is an important problem in intelligent transportation. An interesting alternative is given by employing a downward instead of a forward facing camera, which turns the image-to-image warping into a homographic mapping with known depth (cf. Fig.~\ref{fig:AGV}). A traditional camera based method would be affected by the following, potentially severe challenges: a) reliable feature matching or even extraction may be difficult for certain noisy ground textures, b) fast motion may easily lead to motion blur, and c) stable appearance may require artificial illumination. We therefore consider an event camera as a highly interesting and much more dynamic alternative visual sensor for this particular scenario.

\subsubsection{Homographic Mapping and Bounding Box Extraction}
\begin{figure}[t]
  \centering
  {
  \includegraphics[width = 0.45\textwidth]{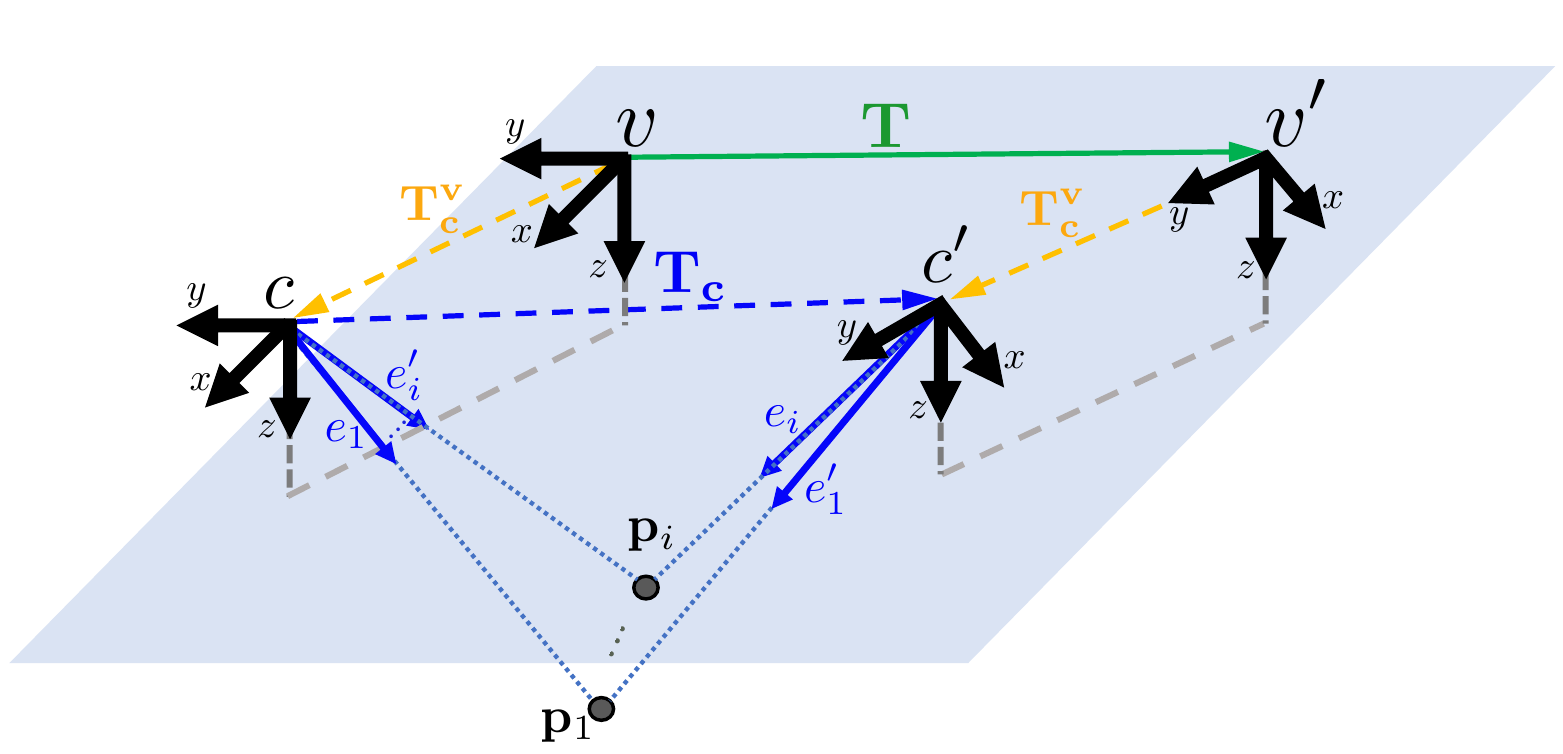}
  \label{Chaining of Transform}
  }
  \caption{Connections between vehicle displacement, extrinsic transformation, and relative camera pose.}
  \label{Ackermann Steering Model + Chaining of Transform}
\end{figure}

We rely on a recently published globally-optimal BnB solver \cite{ling2020efficient} for correspondence-less AGV motion estimation with a normal, downward facing camera. We employ the two-dimensional Ackermann steering model\cite{ling2020efficient,peng2019articulated,Huang_2019_CVPR} describing the commonly non-holonomic motion of an AGV. Employing this 2-DoF model leads to benefits in BnB, the complexity of which strongly depends on the dimensionality of the solution space. The Ackermann model constrains the motion of the vehicle to follow a circular-arc trajectory about an Instantaneous Centre of Rotation~(ICR). The motion between successive frames can be conveniently described at the hand of two parameters: the half-angle of the relative rotation angle~$\phi$, and the baseline between the two views~$\rho$. However, the alignment of the events requires a temporal parametrisation of the relative pose, which is why we employ the angular velocity $\omega = \frac{\theta}{t} = \frac{2\phi}{t}$ as well as the translational velocity $v = \omega r = \omega \rho \frac{1}{2\sin(\phi)}$ in our model. The relative transformation from vehicle frame $v^{\prime}$ back to $v$ is therefore given by
\begin{equation}
  \label{transform_with_respect_to_vehicle}
  \mathbf{R}_v(t) = \left[
                              {\small \begin{matrix}
                                \cos\!\left(\omega t\right) & - \sin\!\left(\omega t\right) & 0 \\
                                \sin\!\left(\omega t\right) &   \cos\!\left(\omega t\right) & 0 \\
                                0 & 0 & 1
                              \end{matrix}}
                            \right], 
  \mathbf{t}_v(t) = \frac{v}{\omega} \left[
                              {\small \begin{matrix}
                                1 - \cos\!\left(\omega t\right) \\
                                    \sin\!\left(\omega t\right) \\
                                                              0
                              \end{matrix}}
                            \right] \,.
\end{equation}

In practice the vehicle frame hardly coincides with the camera frame. The orientation and the height of the origin can be chosen to be identical, and the camera may be laterally mounted in the centre of the vehicle. However, there is likely to be a displacement along the forward direction, which we denote by the signed variable~$l$. In other words, $\mathbf{R}_{v}^{c}=\mathbf{I}_{3\times 3}$ and $\mathbf{t}_{v}^{c}=\left[ \begin{matrix} 0 \text{ } l \text{ } 0 \end{matrix} \right]^\mathsf{T}$. As illustrated in Fig.~\ref{Ackermann Steering Model + Chaining of Transform}, the transformation from camera pose $c^{\prime}$ (at an arbitrary future timestamp) to $c$ (at the initial timestamp $t_{\text{ref}}$) is therefore given by
\begin{equation}
\label{rotation_matrix_and_translation_vector}
\begin{split}
  & \mathbf{R}_c(t) = {\mathbf{R}_{v}^{c}}^\mathsf{T}\mathbf{R}_v(t) \mathbf{R}_{v}^{c} \,, \\
  & \mathbf{t}_c(t) = -{\mathbf{R}_{v}^{c}}^\mathsf{T}\mathbf{t}_{v}^{c} + {\mathbf{R}_{v}^{c}}^\mathsf{T}\mathbf{t}_v(t) + {\mathbf{R}_{v}^{c}}^\mathsf{T}\mathbf{R}_v(t)\mathbf{t}_{v}^{c}\,.
\end{split}
\end{equation}

Using the known plane normal vector $\mathbf{n}=[0\text{ } 0\text{ }-1 ]^\mathsf{T}$ and depth-of-plane~$d$, the image warping function~$W(\mathbf{x}_k,t_k;[\omega\text{ }v]^\mathsf{T})$ that permits the transfer of an event~$e_k = \{ \mathbf{x}_k,t_k,s_k \}$ into the reference view at~$t_{\text{ref}}$ is finally given by the planar homography equation
\begin{equation}
    \label{homography}
    \textbf{H}(t_k-t_{\text{ref}})\left[ \begin{matrix} \mathbf{x}_k \\ 1 \end{matrix}\right]=\textbf{K}(\textbf{R}_c(t_k-t_{\text{ref}})-\frac{\textbf{t}_c(t_k-t_{\text{ref}})\textbf{n}^\mathsf{T}}{d})\textbf{K}^\mathsf{-1}\left[ \begin{matrix} \mathbf{x}_k \\ 1 \end{matrix}\right] \,.
\end{equation}
Note that $\mathbf{K}$ here denotes a regular perspective camera calibration matrix with homogeneous focal length $f$, zero skew, and a principal point at $\left[ \begin{matrix} u_0 \text{ } v_0 \end{matrix} \right]^\mathsf{T}$. Note further that the result needs to be dehomogenised. After expansion, we easily obtain
\begin{eqnarray}
  \mathbf{x}_k^{\prime} & = & W(\mathbf{x}_k,t_k;[\omega\text{ }v]^\mathsf{T}) = {\left[ \begin{matrix} x_k^{\prime} \  y_k^{\prime} \end{matrix} \right]}^\mathsf{T} \text{, where} \\
  x_k^{\prime} & = & 
      - [y_k - v_0 + l \cdot \frac{f}{d}] \sin(\omega (t_k-t_{\text{ref}})) \nonumber \\
      & & + [x_k - u_0 - \frac{f}{d} \cdot \frac{v}{w}] \cos(\omega (t_k-t_{\text{ref}}))
      +  \frac{f}{d} \cdot \frac{v}{w} + u_0 , \nonumber \\
  y_k^{\prime} & = & [x_k - u_0 - \frac{f}{d} \cdot \frac{v}{w}] \sin(\omega (t_k-t_{\text{ref}})) \nonumber \\
      & & + [y_k - v_0 + l \cdot \frac{f}{d}] \cos(\omega (t_k-t_{\text{ref}})) - l \cdot \frac{f}{d} + v_0 . \nonumber
\end{eqnarray}

Finally, the bounding box $\mathcal{P}_{k }^{\boldsymbol{\Theta}}$ is found by bounding the values of $x_k^{\prime}$ and $y_k^{\prime}$ over the intervals $\omega\in\mathcal{W}=\left[\omega_{\text{min}},\omega_{\text{max}} \right]$ and $v\in\mathcal{V}=\left[v_{\text{min}} , v_{\text{max}} \right]$. The bounding is easily achieved if simply considering monotonicity of functions over given sub-branches. For example, if $\omega_{\text{min}} \geq 0$, $v_{\text{min}} \geq 0$, $x_k \geq u_0$, and $y_k \geq v_0 - l \cdot \frac{f}{d}$, we obtain

\small
\begin{eqnarray}
    \underline{x'_k} &=& - [y_k - v_0 + l \cdot \frac{f}{d}] \sin(\omega_{\text{max}} t) \nonumber \\
                    && + [x_k - u_0 - \frac{f}{d} \cdot  \frac{v_{\text{min}}}{\omega_{\text{max}}}] \cos(\omega_{\text{max}} t)
                    + \frac{f}{d} \cdot  \frac{v_{\text{min}}}{\omega_{\text{max}}} + u_0 \,, \nonumber \\
    \overline{x'_k}  &=& - [y_k - v_0 + l \cdot \frac{f}{d}] \sin(\omega_{\text{min}} t) \nonumber  \\
                    && + [x_k - u_0 - \frac{f}{d} \cdot \frac{v_{\text{max}}}{\omega_{\text{min}}}] \cos(\omega_{\text{min}} t)
                    + \frac{f}{d} \cdot \frac{v_{\text{max}}}{w_{\text{min}}} + u_0 \,, \nonumber \\
    \underline{y'_k} &=&   [x_k - u_0 - \frac{f}{d} \cdot \frac{v_{\text{max}}}{\omega_{\text{min}}}] \sin(\omega_{\text{min}} t) \nonumber  \\
                    && + [y_k - v_0 + l \cdot  \frac{f}{d}] \cos(\omega_{\text{max}} t)
                    - l \cdot \frac{f}{d} + v_0 \,, \text{ and} \nonumber \\
    \overline{y'_k} &=&    [x_k - u_0 - \frac{f}{d} \cdot \frac{v_{\text{min}}}{\omega_{\text{max}}}] \sin(\omega_{\text{max}} t) \nonumber  \\
                   && + [y_k - v_0 + l \cdot \frac{f}{d}] \cos(\omega_{\text{min}} t)
                    - l \cdot \frac{f}{d} + v_0 \,.
\end{eqnarray}
\normalsize

We kindly refer the reader to the appendix for all further cases.

\subsubsection{Results and Discussion}
\begin{figure*}[t!]
\centering
\includegraphics[width=0.33\textwidth]{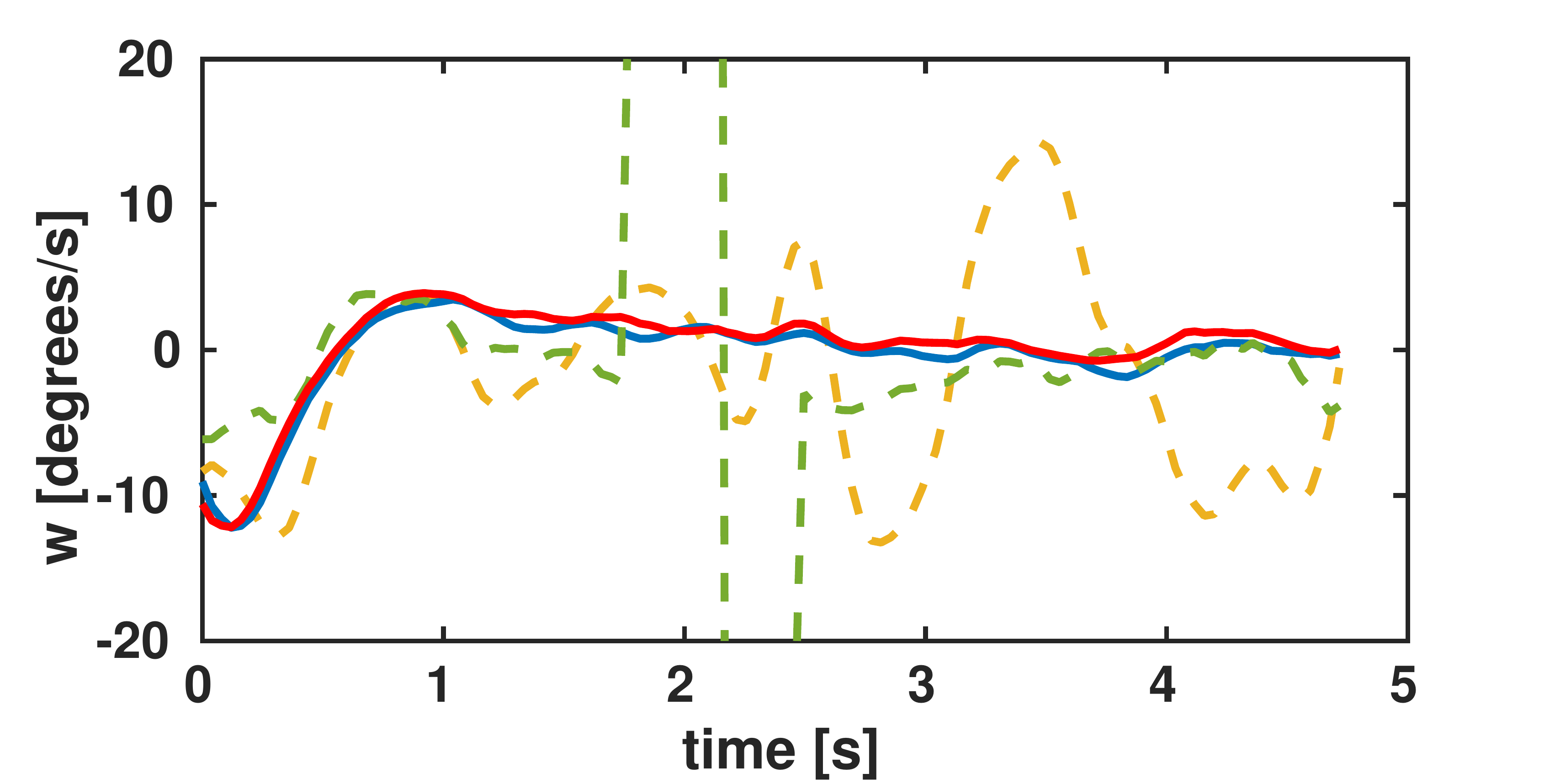}
\includegraphics[width=0.33\textwidth]{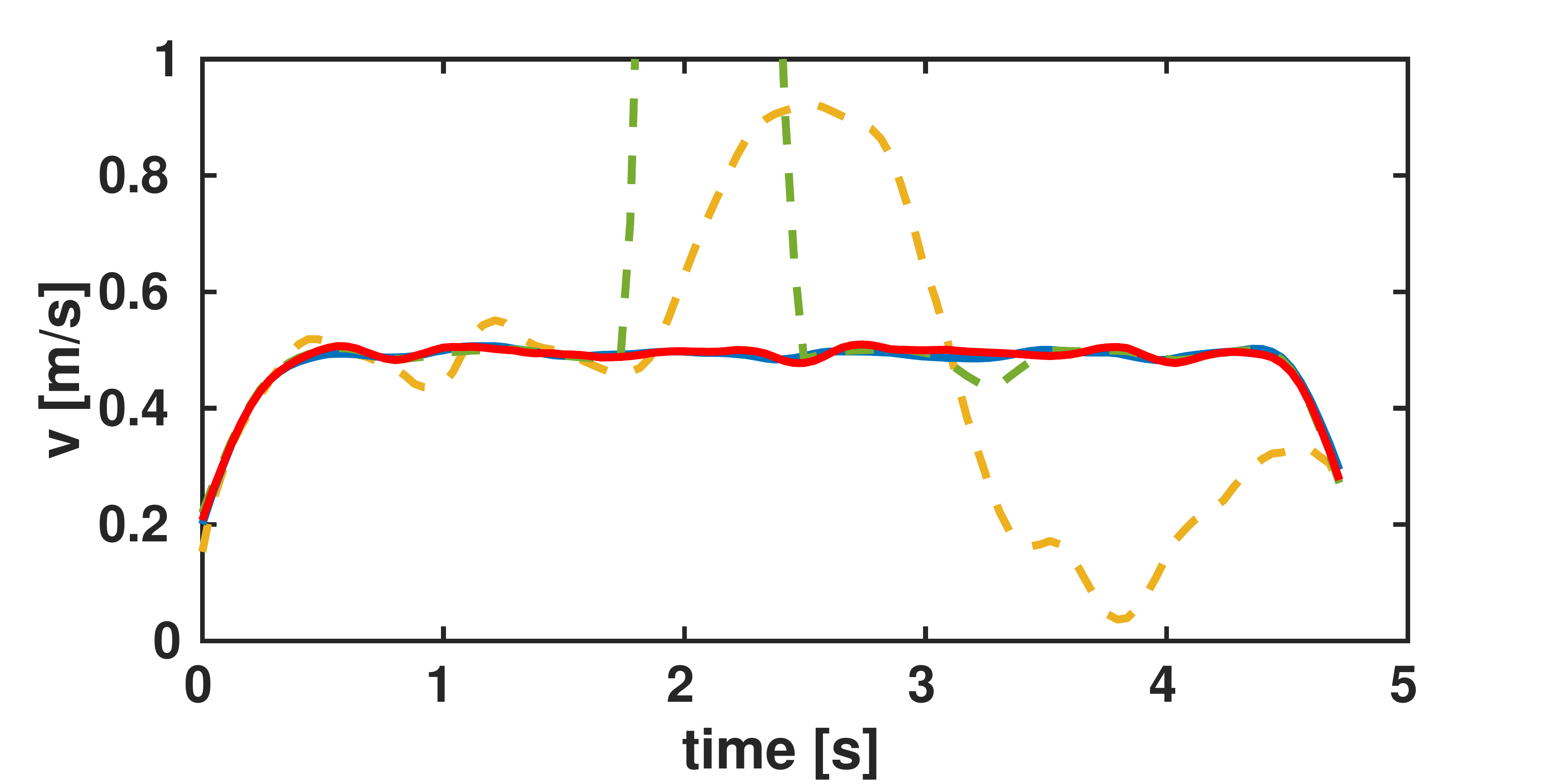}
\includegraphics[width=0.33\textwidth]{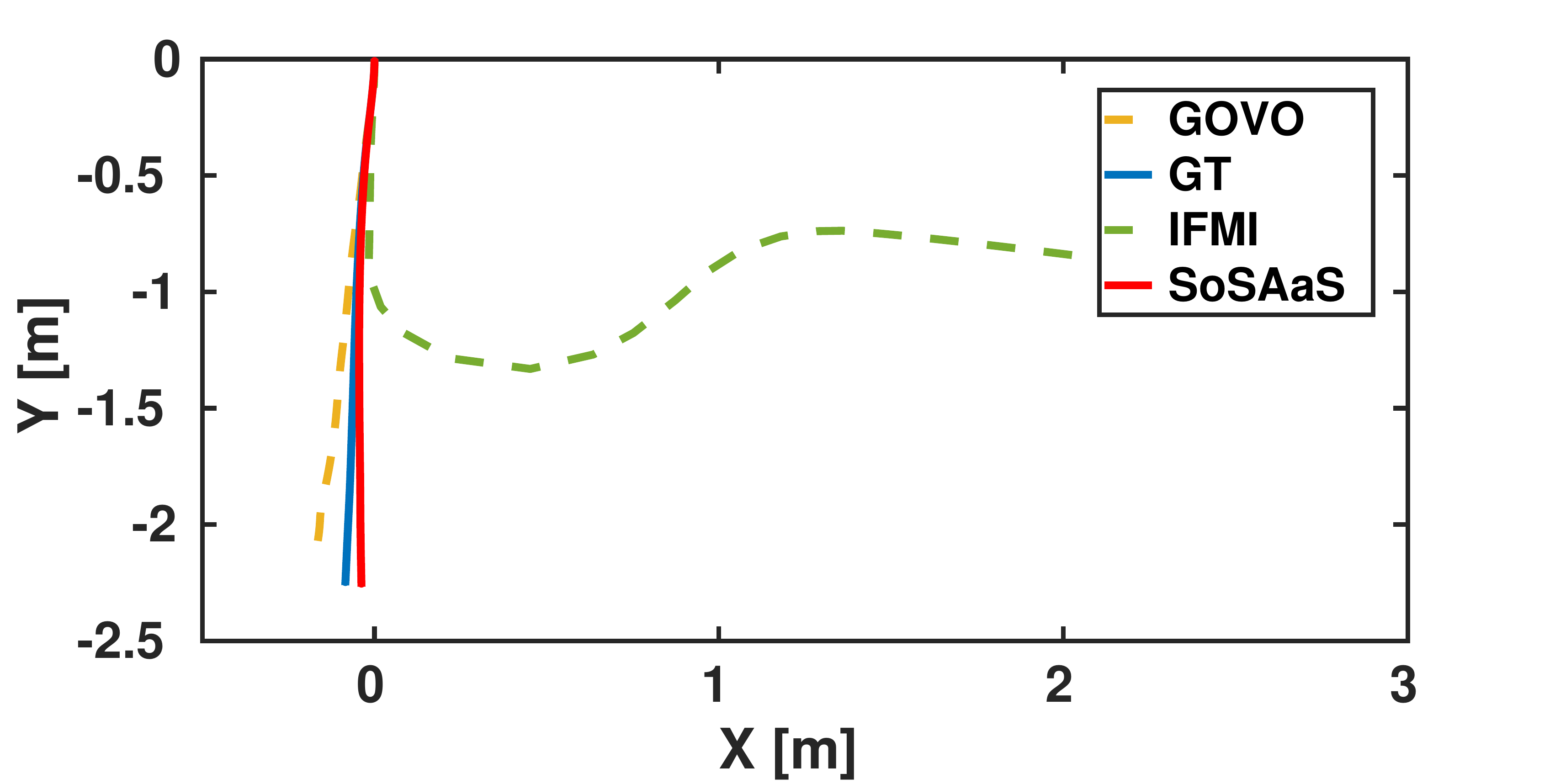}\\
\includegraphics[width=0.33\textwidth]{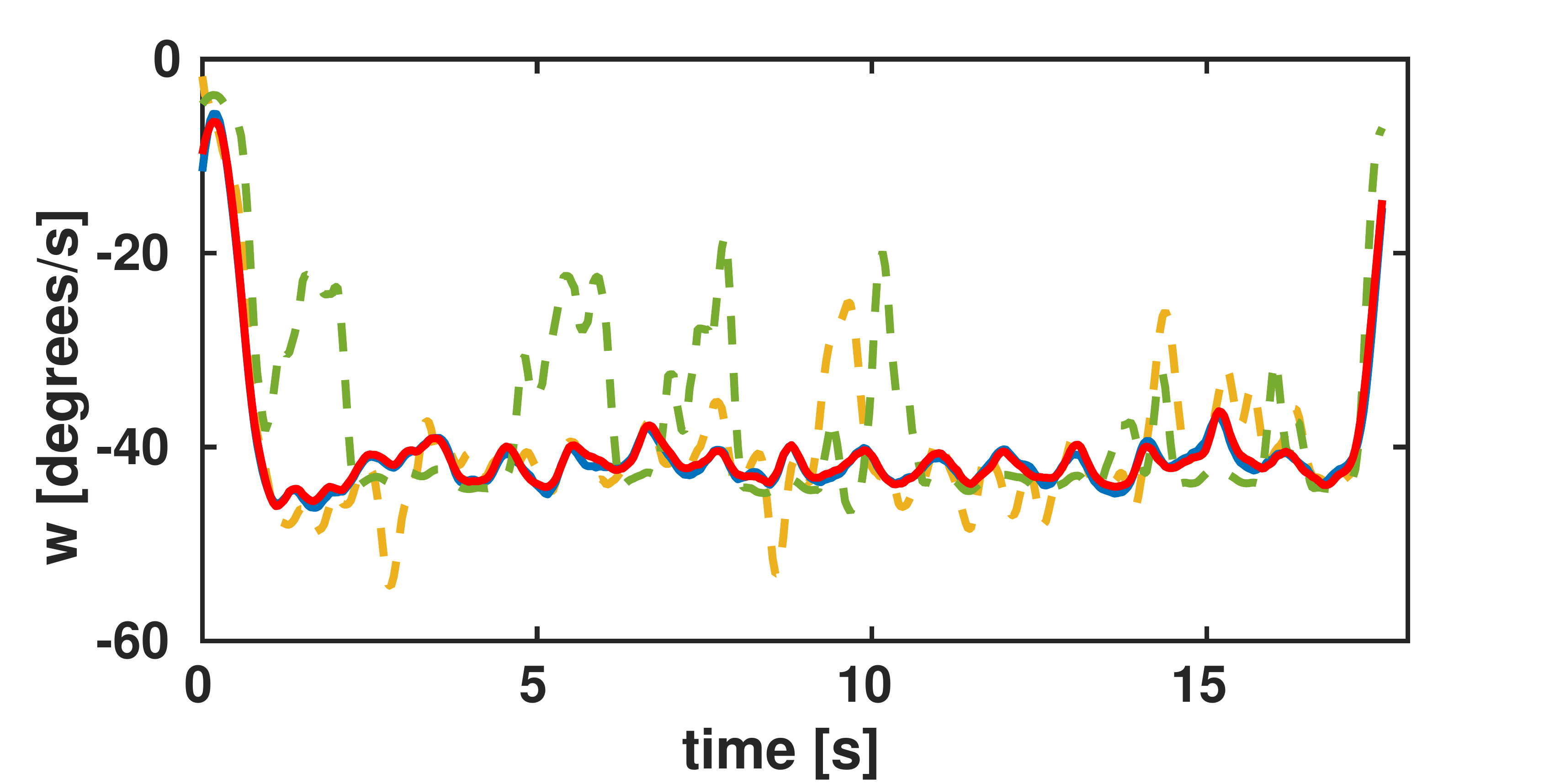}
\includegraphics[width=0.33\textwidth]{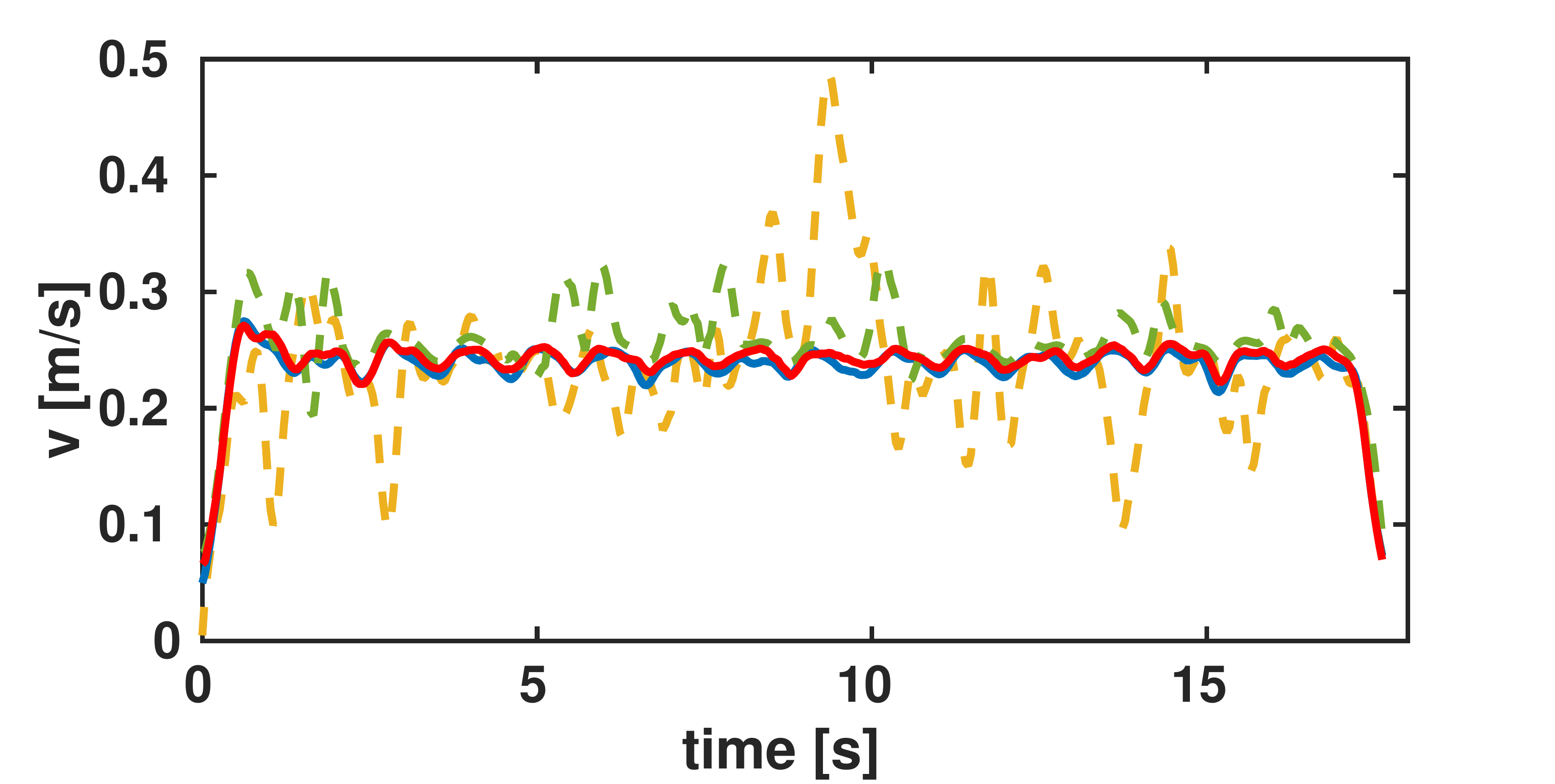}
\includegraphics[width=0.33\textwidth]{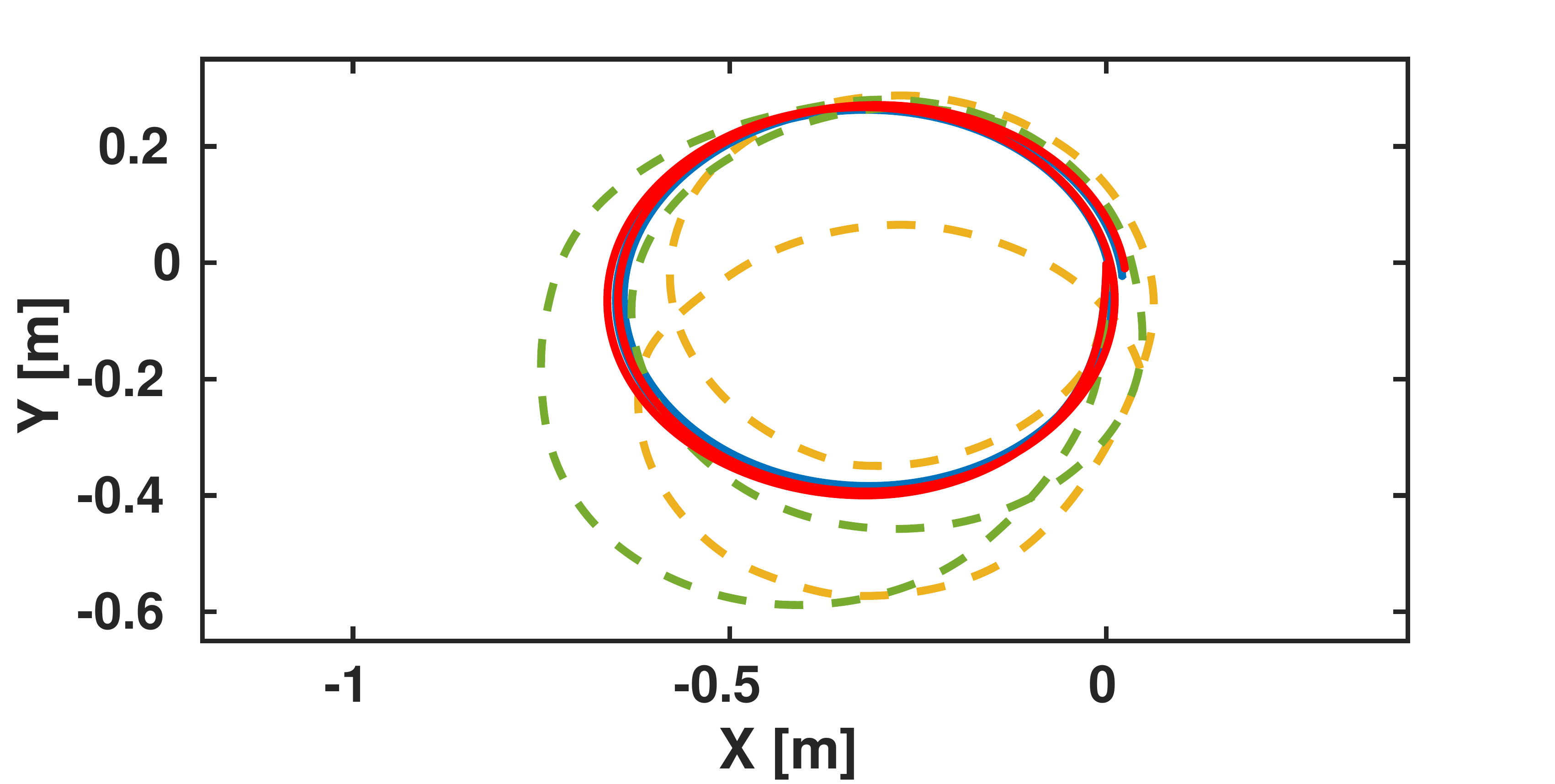}\\
\includegraphics[width=0.33\textwidth]{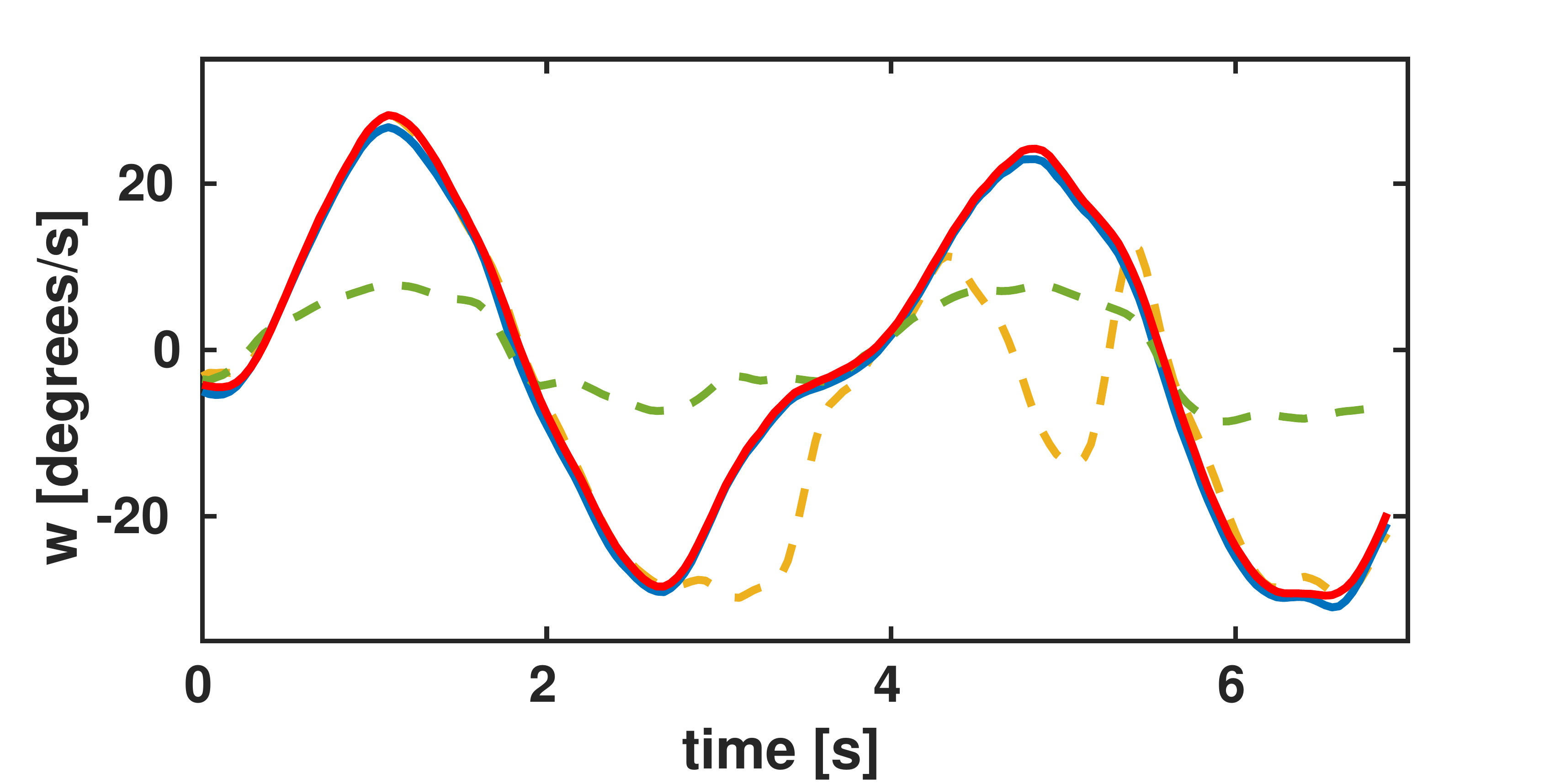}
\includegraphics[width=0.33\textwidth]{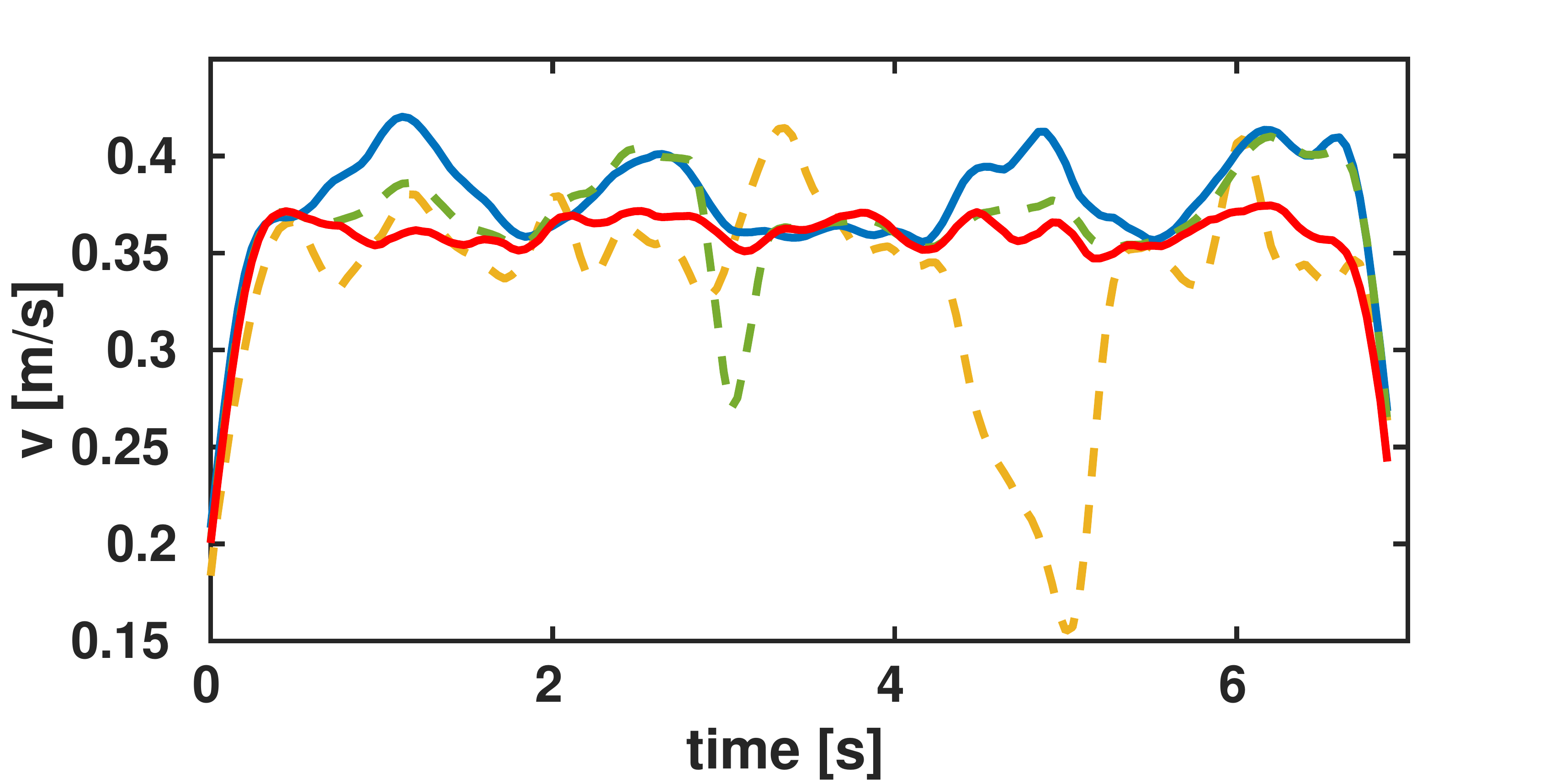}
\includegraphics[width=0.33\textwidth]{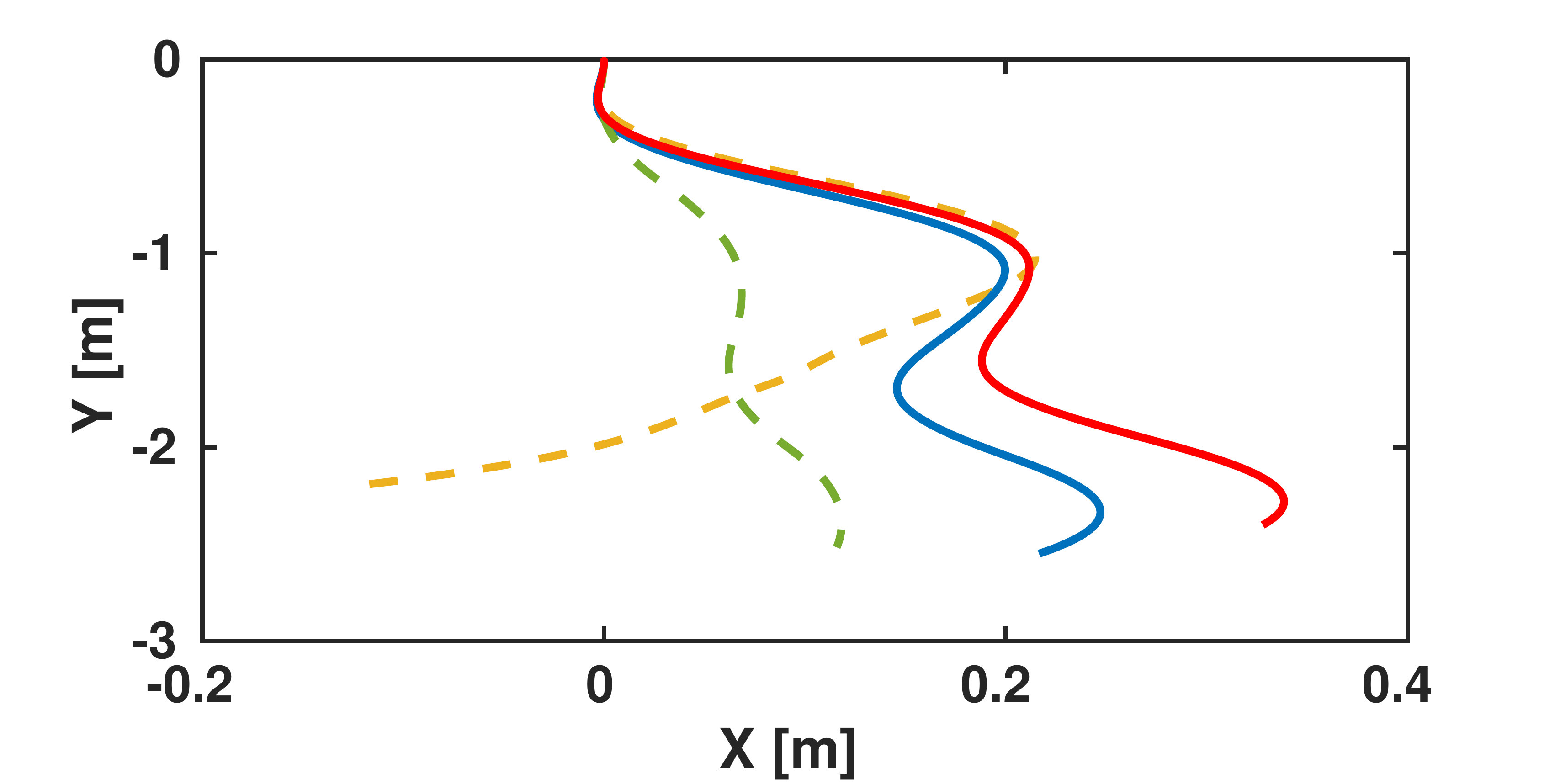} 
\caption{
Results for all methods over different datasets. The first two columns are errors over time for $\omega$ and $v$, and the third column illustrates a bird's eye view onto the integrated trajectories. Both IFMI and GOVO occasionally lose tracking (especially for linear motion), which leaves our proposed globally-optimal event-based method using $L_{\mathrm{SoSAaS}}$ as the clearly outperforming method.}
\label{fig:Real_Data}
\end{figure*}

We apply our method to real data collected by a DAVIS346 event camera, which outputs events streams with a maximum time resolution of $1 \mu s$ as well as regular frames at a frame rate of 30Hz. Images have a resolution of 346$\times$260. The camera is mounted on the front of a XQ-4 Pro robot and faces downwards (see Figure \ref{fig:AGV}). The displacement from the non-steering axis to the camera is $l = -0.45$m, and the height difference between camera and ground is $d = 0.23$m. We recorded several motion sequences on a wood grain foam which has highly self-similar texture and poses a challenge to reliably extract and match features. Ground truth is obtained via an Optitrack optical motion tracking system. Our algorithm is working in undistorted coordinates, which is why normalisation and undistortion are computed in advance. The following aspects are evaluated:

\textbf{Event-based vs frame-based}:
GOVO~\cite{ling2020efficient} and IFMI~\cite{xu2019improved} are frame-based algorithms specifically designed for planar AGV motion estimation under featureless conditions. Fig.~\ref{fig:robot and frame} shows an example frame of the wood grain foam texture, and Fig.~\ref{fig:Real_Data} the results obtained for all methods. As can be observed, GOVO finds as little as three corner features for some of the images, thus making it difficult to accurately recover the vehicle displacement despite the globally-optimal correspondence-less nature of the algorithm. Both IFMI and GOVO occasionally lose tracking (especially for linear motion), which leaves our proposed globally-optimal event-based method using $L_{\mathrm{SoSAaS}}$ as the clearly outperforming method. Table~\ref{tab:frame_based_errors} lists the RMS errors over the angular velocity and linear velocity. Note that the average runtime of GOCMF over \textit{Line}, \textit{Circle} and \textit{Curve} is about 55s, 85s and 71s respectively.
\begin{table}[t]
\renewcommand\arraystretch{1.5}
\caption{RMS errors for Event-based and frame-based methods}
\label{tab:frame_based_errors}
\centering
\setlength{\tabcolsep}{1.4mm}
\begin{tabular}{ccccccc}
\toprule
\textbf{Method} 
& {\begin{tabular}[c]{@{}c@{}}\textit{Line}\\ w {[}$^{\circ}$/s{]}\end{tabular}} 
& {\begin{tabular}[c]{@{}c@{}}\textit{Line}\\ v {[}m/s{]}\end{tabular}} 
& {\begin{tabular}[c]{@{}c@{}}\textit{Circle}\\ w {[}$^{\circ}$/s{]}\end{tabular}} 
& {\begin{tabular}[c]{@{}c@{}}\textit{Circle}\\ v {[}m/s{]}\end{tabular}} 
& {\begin{tabular}[c]{@{}c@{}}\textit{Curve}\\ w {[}$^{\circ}$/s{]}\end{tabular}} 
& {\begin{tabular}[c]{@{}c@{}}\textit{Curve}\\ v {[}m/s{]}\end{tabular}} \\ \midrule
\textbf{GOCMF}  & \textbf{0.517}  & \textbf{0.008}   & \textbf{0.529}   & \textbf{0.004}  & \textbf{0.554} & \textbf{0.018}                                               \\ 
\textbf{IFMI}    & 145.37    & 1.059     & 8.109    & 0.024    & 12.804    & 0.019   \\ 
\textbf{GOVO}   & 6.970    & 0.240     & 4.550   & 0.064   & 9.865  & 0.059        \\ \bottomrule
\end{tabular}
\end{table}

\begin{figure}
\centering
\includegraphics[width=0.5\textwidth]{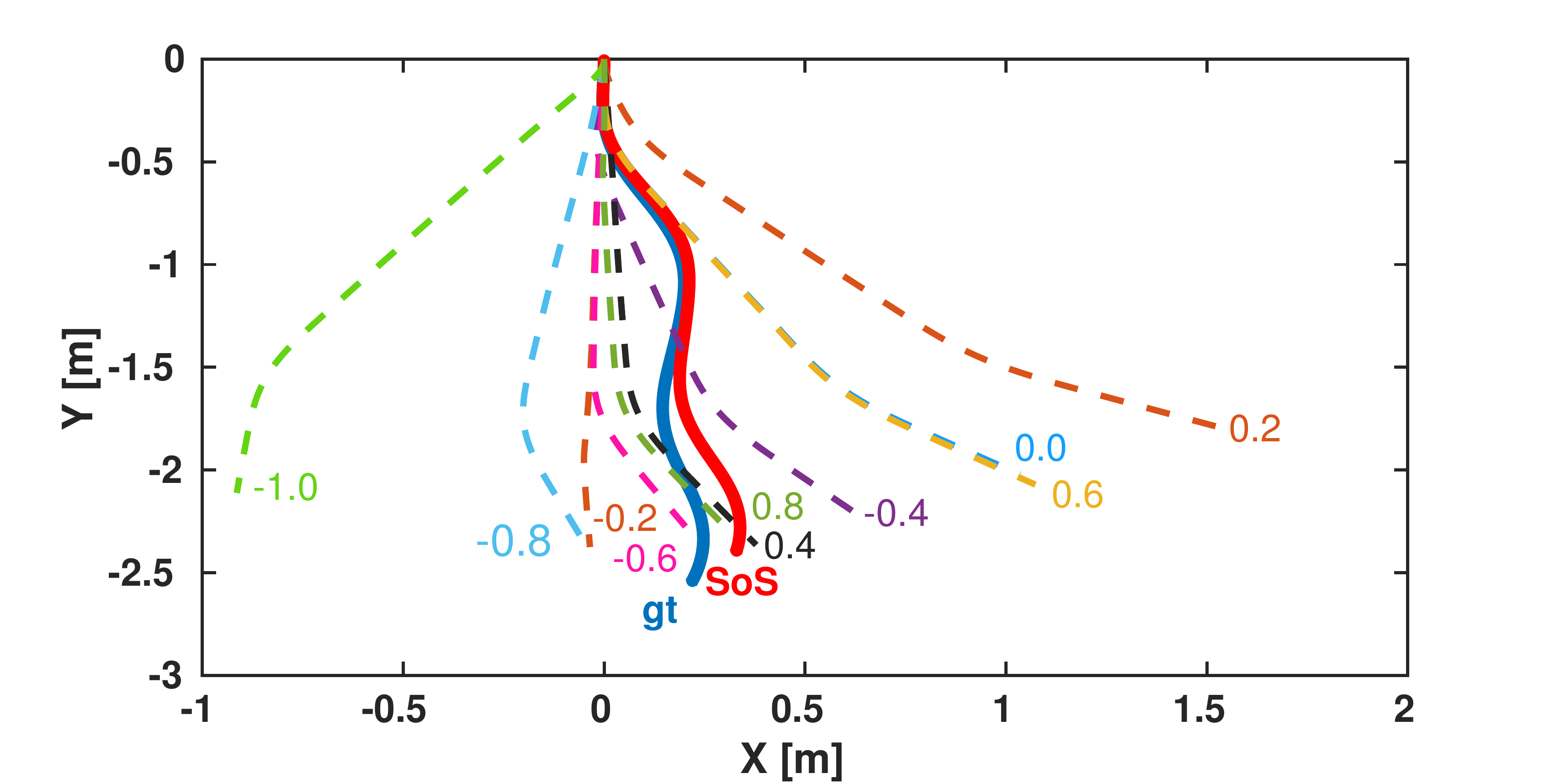}
\caption{Estimated trajectories by our method ($SoS$), gradient ascent with various initializations, and ground truth ($gt$). Obviously, good initializations are important for local optimisation, while GOCMF has no such requirements.}
\label{fig:gd_tra}
\end{figure}

\textbf{BnB vs Gradient Ascent}: We apply both gradient descent as well as BnB to the \textit{Foam} dataset with curved motion. For the first temporal interval and the local search method, we vary the initial angular velocity $\omega$ and linear velocity $v$ between -1 and 0.8 with steps of 0.2 (rad/s or m/s, respectively). For later intervals, we use the previous local optimum. Fig.~\ref{fig:gd_tra} illustrates the estimated trajectories for all initial values, compared against ground truth and a BnB search using $L_{\mathrm{SoS}}$. RMS errors are also indicated in Table~\ref{tab:error_ga}. As clearly shown, even the best initial guess eventually diverges under a local search strategy, thus leading to clearly inferior results compared to our globally optimal search.
\begin{table}
\begin{minipage}{0.26\textwidth}
\centering
\renewcommand\arraystretch{1.2}
\caption{RMS errors for \protect\\ gradient ascent and SoS}
\label{tab:error_ga}
\begin{tabular}{ccc}
\toprule
\bf Method & \begin{tabular}[c]{@{}c@{}}w\\ {[}$^{\circ}$/s{]}\end{tabular} & \begin{tabular}[c]{@{}c@{}} v \\ {[}m/s{]}\end{tabular} \\ \midrule
\bf SoS & 3.0091 & 0.0208 \\ 
\bf GA & 11.5023 & 0.0379 \\
\bottomrule
\end{tabular}
\end{minipage}
\begin{minipage}{0.22\textwidth}
\caption{RMS errors for \protect\\ the different textures}
\label{tab:error_texture}
\centering
{
\renewcommand\arraystretch{1.2}
\begin{tabular}{ccc}
\toprule
\bf Scene & \begin{tabular}[c]{@{}c@{}}w\\ {[}$^{\circ}$/s{]}\end{tabular} &  \begin{tabular}[c]{@{}c@{}} v \\ {[}m/s{]}\end{tabular} \\ \midrule
\bf Carpet & 4.730 & 0.034 \\ 
\bf Poster & 3.122 & 0.030 \\
\bottomrule
\end{tabular}
}
\end{minipage}
\end{table}

\begin{figure*}[t]
\centering
\subfigure
{
\includegraphics[width=0.19\textwidth]{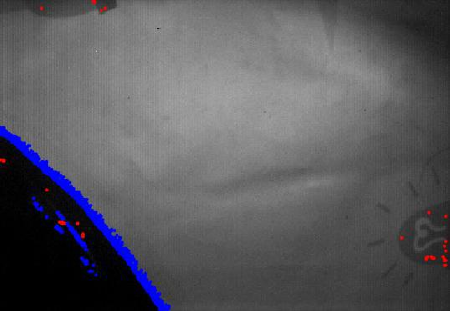}
\includegraphics[width=0.19\textwidth]{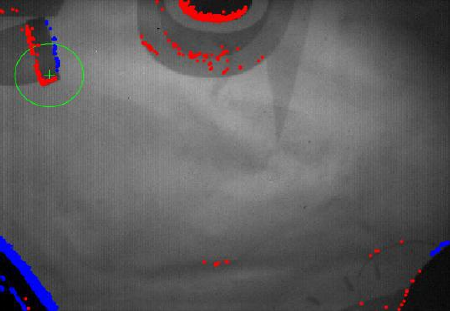}
\includegraphics[width=0.19\textwidth]{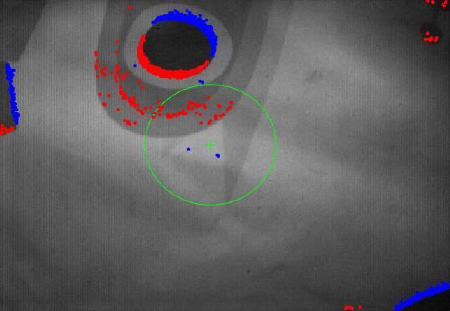}
\includegraphics[width=0.19\textwidth]{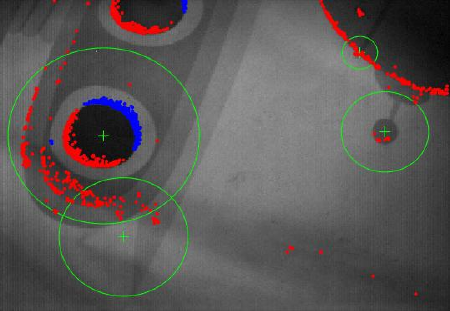}
\includegraphics[width=0.19\textwidth]{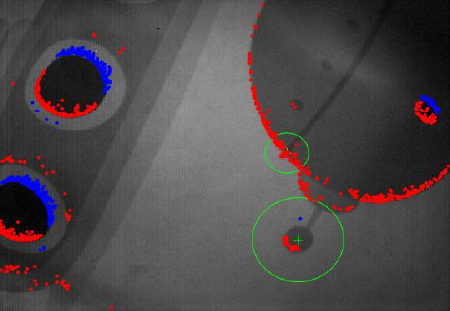}
}
\subfigure
{
\includegraphics[width=0.19\textwidth]{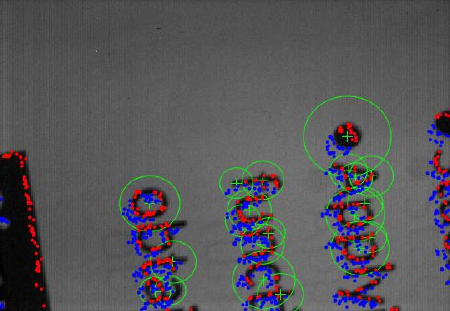}
\includegraphics[width=0.19\textwidth]{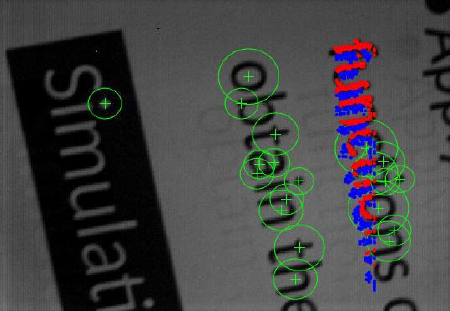}
\includegraphics[width=0.19\textwidth]{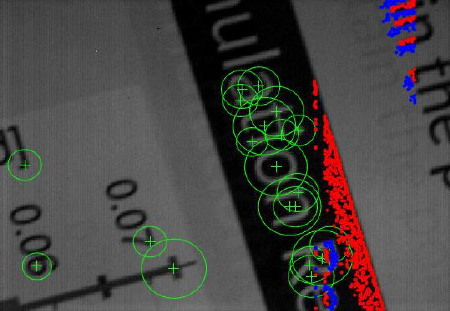}
\includegraphics[width=0.19\textwidth]{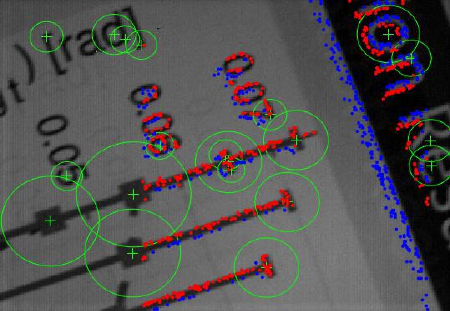}
\includegraphics[width=0.19\textwidth]{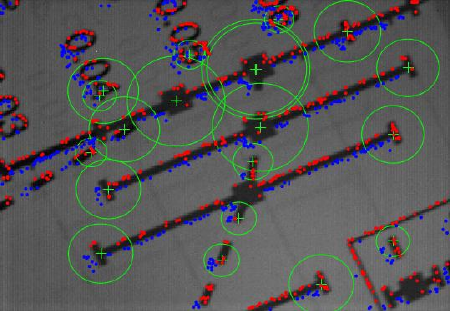}
}
\caption{Frames from dataset \textit{Carpet} (first row) and \textit{Poster} (second row).}
\label{fig:carpet_and_poster}
\end{figure*}

\textbf{Various textures}:  To further analyse the robustness, we test our algorithm on datasets collected with various textures. Fig.~\ref{fig:carpet_and_poster} presents frames from two further datasets named \textit{Carpet} and \textit{Poster}. The \textit{Carpet} sequences are collected on a carpet with non-repetitive almost featureless texture, while the \textit{Poster} sequences are collected on a poster with characters and figures for which it is easy to extract features. The estimated errors are summarised in Table~\ref{tab:error_texture}. As can be observed, our algorithm consistently shows a similar level of accuracy for the various textures in the datasets.

\subsection{Rotational motion estimation}
\label{sec:Rotational}

Our final application of our contrast maximisation framework considers event-based pure rotation estimation \cite{gallego2017accurate,gallego2018unifying,liu2020globally}. The latter is a recurrent topic as general motion estimation often depends on a non-zero baseline assumption, and thus may break in situations of negligibly small camera translation. The technique is furthermore commonly applied to video stabilisation \cite{delbruck2014integration} and panoramic image creation \cite{Kim2014Simultaneous}. Our technique is applicable as the flow of the events in a purely rotational displacement situation can also be characterized by a homographic warping (i.e. the homography at infinity).

We assume a constant angular velocity $\boldsymbol{\omega} \in \mathbb{R}^3$ to parameterize the rotational motion over a sufficiently small time interval. The rotation is given by
\begin{equation}
    \mathbf{R}(t;\boldsymbol{\omega}) =  \exp{({\boldsymbol{\omega}^{\wedge}}t)},
\end{equation}
where $\boldsymbol{\omega}^{\wedge}$ is the $3\times3$ screw symmetric matrix form of $\boldsymbol{\omega}$, and $\exp$ is the exponential map of the rotation group $SO(3)$ \cite{Holmgren2004An}. The warping function is finally given by  
\begin{equation}
    \mathbf{f}_{k}^{\prime} = \exp{({\boldsymbol{\omega}}^{\wedge} t_k)} \mathbf{f}_{k},
\end{equation}
where $\mathbf{f}_{k} = normalize(\textbf{K}^\mathsf{-1} \left[ \begin{matrix} \mathbf{x}_k^\mathsf{T} & 1 \end{matrix}\right]^\mathsf{T})$ is the bearing vector of the event $\mathbf{x}_{k}$ at time $t_k$, and $\mathbf{f}_{k}^{\prime}$ is the rotated bearing vector expressed in the reference frame at time $t_\text{ref}$. Hence we project $\mathbf{f}_{k}^{\prime}$ to the image plane to obtain the location of the warped event $\mathbf{x}_{k}^{\prime}$.

\begin{figure}
\centering
\includegraphics[width=0.35\textwidth]{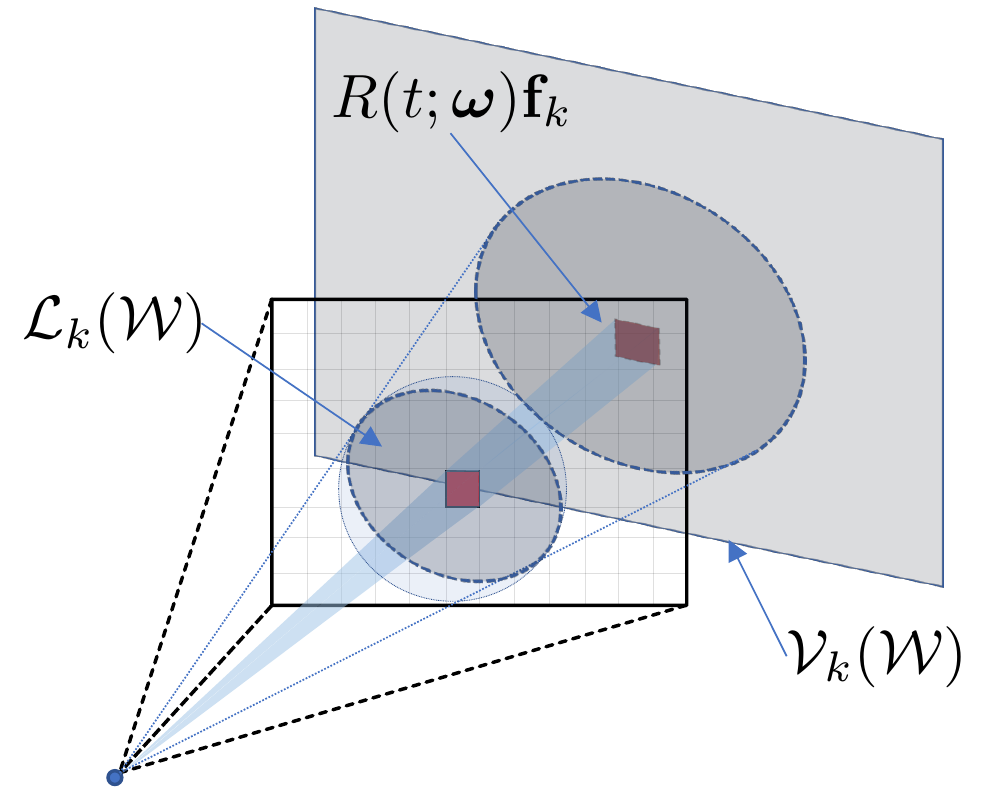}
\caption{The cone $\mathcal{V}_k(\boldsymbol{\Omega})$ that $\mathbf{R}(t;\boldsymbol{\omega})\mathbf{f}_k$ might locate in with search space $\boldsymbol{\Omega}$. We also illustrate the elliptical region $\mathcal{L}_k(\boldsymbol{\Omega})$ on the pixel plane.}
\label{fig:rotation_bounding_box}
\end{figure}

As mentioned in Algorithm~\ref{alg:GOCMF} and Algorithm~\ref{alg:RB}, it is again essential to derive a bounding box for each warped event in the IWE. We adopt the bounding box derived in \cite{liu2020globally}. Given a search space $\boldsymbol{\omega} \in \boldsymbol{\Omega}$ with center $\boldsymbol{\omega}_0$ and bounds $\boldsymbol{\omega}_\text{min}$ and $\boldsymbol{\omega}_\text{max}$, define 
\begin{equation}
    \alpha_k(\boldsymbol{\Omega}) := 0.5 \parallel \boldsymbol{\omega}_{\text{min}} t_k - \boldsymbol{\omega}_{\text{max}} t_k \parallel_2 .
\end{equation}
According to \cite{hartley2009global} and \cite{liu2020globally}, all the possible rotated bearing vectors $\mathbf{f}_{k}^{\prime}$ will then lie in a cone 
\begin{equation}
     \mathcal{V}_k(\boldsymbol{\Omega}) := \{ \mathbf{f} \in \mathbb{R}^3 |\angle( \exp{(\boldsymbol{\omega}^{\wedge}_0 t_k)} \mathbf{f}_{k},  \mathbf{f} ) \leq \alpha_k (\boldsymbol{\Omega}) \}.
\end{equation}
As illustrated in Fig.~\ref{fig:rotation_bounding_box}, the projection of bearing vectors in cone $\mathcal{V}_k(\boldsymbol{\Omega})$ yields an elliptical region $\mathcal{L}_k(\boldsymbol{\Omega})$ on the pixel plane. The derivation of the center and semi-major axes of $\mathcal{L}_k(\boldsymbol{\Omega})$ can be found in the work by Liu et al. \cite{liu2020globally}.
%
\subsubsection{Results and Discussion}

We proceed to our comparison between the proposed algorithm GOCMF and the alternative globally-optimal algorithm CMBNB presented by Liu et al. \cite{liu2020globally}. We furthermore compare our results against a local optimiser, denoted CMGD (\texttt{fmincon} function from Matlab). We use the publicly available sequences \textit{poster}, \textit{boxes} and \textit{dynamic}
from \cite{gallego2017accurate,2016arXiv161008336M}, which were recorded using a Davis240C \cite{6889103} under rotational motion over a static indoor scene. The ground truth motion was captured using a motion capture system. We utilized the last 15s of each sequence and split them into 10ms subsequences to run the algorithms. For each subsequence from \textit{poster} and \textit{boxes}, the events size is $N \approx 50k$, while for \textit{dynamic}, the size is $N \approx 25k$. To speed up the algorithm, we downsampled the event stream by a factor 2, which means we dropped half the number of events in each subsequence. Note that the objective function we used in the experiments is $L_{\mathrm{SoS}}$ on which CMBNB~\cite{liu2020globally} is based. We ran algorithms on a standard desktop with a Intel Core i7-7700 and CPU @ 3.60GHz $\times$ 8.  Moreover, GOCMF and CMBNB were both run with 8 threads.

We evaluated the algorithms using two error metrics. One is
\begin{equation}
  \epsilon = \|\omega_{\text{gt}} - \omega_{\text{est}} \|_2,
\end{equation}
where $\omega_{\text{gt}}$ and $\omega_{\text{est}}$ are the ground truth and estimated angular velocities, respectively. The other error metric is
\begin{equation}
  \phi = | \|\omega_{\text{gt}}\|_2 - \|\omega_{\text{est}} \|_2 |.
\end{equation}
\begin{table*}[t!]
\centering
\caption{Rotational Motion Estimation Errors}
\label{tab:rotation_estimated_errors}
\renewcommand\arraystretch{2}
\begin{tabular}{ccllllcllllcllll}
\toprule
\multirow{2}{*}{\textbf{Method}} & \multicolumn{5}{c}{\textit{dynamic}} & \multicolumn{5}{c}{\textit{boxes}} & \multicolumn{5}{c}{\textit{poster}} \\ \cline{2-16} &  $\mu(\phi)$     & $\delta(\phi)$   & $\mu(\epsilon)$      &  $\delta(\epsilon)$     &  time   & $\mu(\phi)$      & $\delta(\phi)$  & $\mu(\epsilon)$      &  $\delta(\epsilon)$    & time    & $\mu(\phi)$    & $\delta(\phi)$  & $\mu(\epsilon)$      &  $\delta(\epsilon)$   & time  \\ \midrule
\textbf{CMGD}    &161.6      &127.9      & 168.7  & 124.1 &                      \textbf{6.96} &                        61.75  & 103.6  &71.82 & 101.0  & \textbf{9.08}   &  43.64 & 74.29  &59.29 &79.29  & \textbf{16.63}     \\ 
\textbf{CMBNB}   & 9.88     &  6.94   & \textbf{17.02} & 8.93 & 76.46    & \textbf{19.41}     & \textbf{13.34}  &\textbf{30.21} &\textbf{15.49}   & 92.48   &  22.05 & 23.18  &32.94 &23.87   & 221.0    \\ 
\textbf{GOCMF}  & \textbf{9.75}   & \textbf{6.66}    & 17.05 & \textbf{8.58}    & 22.88   & 20.19 & 16.19   & 31.27 &21.10  & 30.08    & \textbf{21.98}  & \textbf{23.02}  & \textbf{32.39} & \textbf{23.44}  & 88.48 \\
\bottomrule
\end{tabular}
\end{table*}
\begin{figure*}[t]
\centering
\includegraphics[width=0.24\textwidth ,height = 0.16\textwidth]{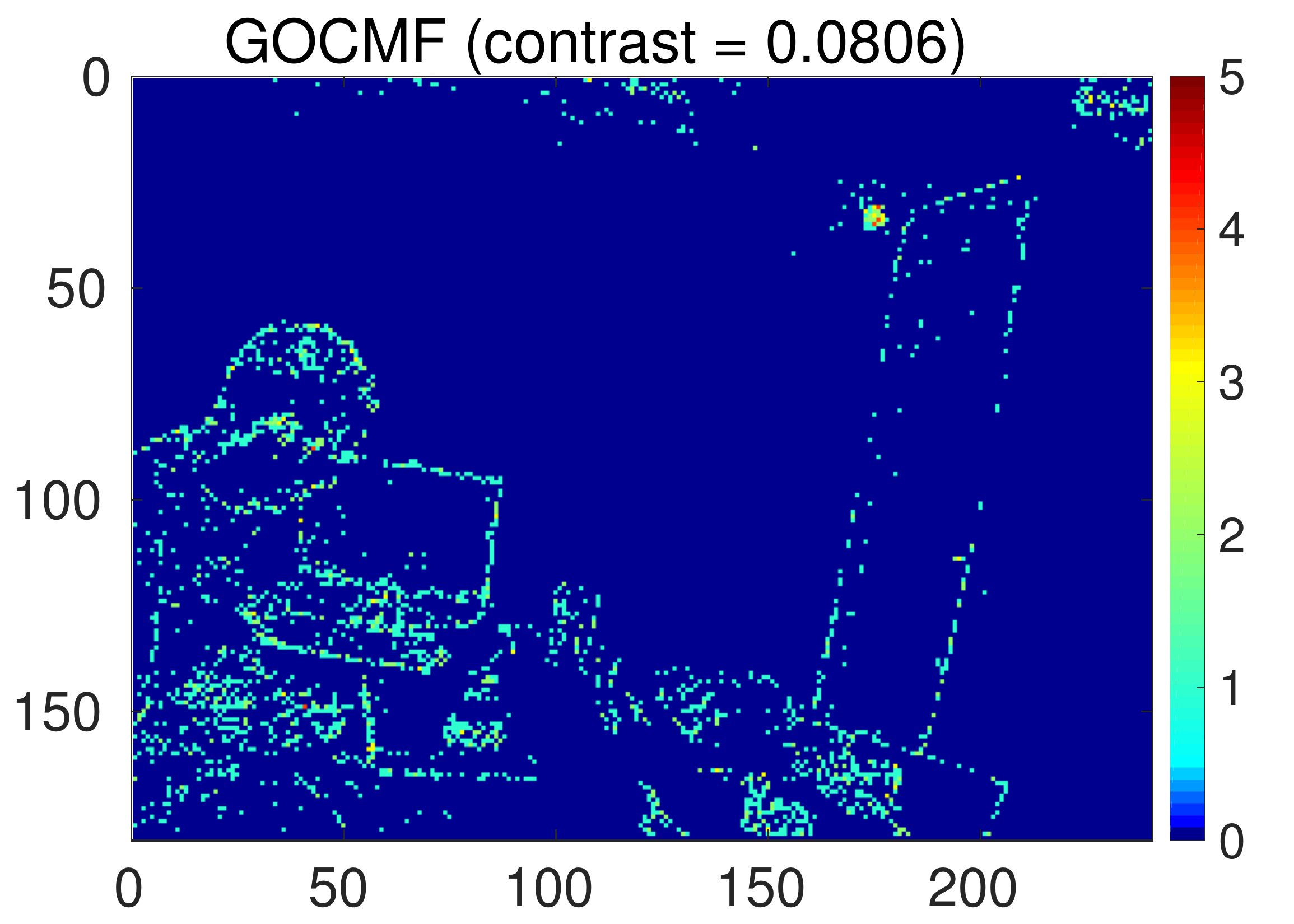}
\includegraphics[width=0.24\textwidth ,height = 0.16\textwidth]{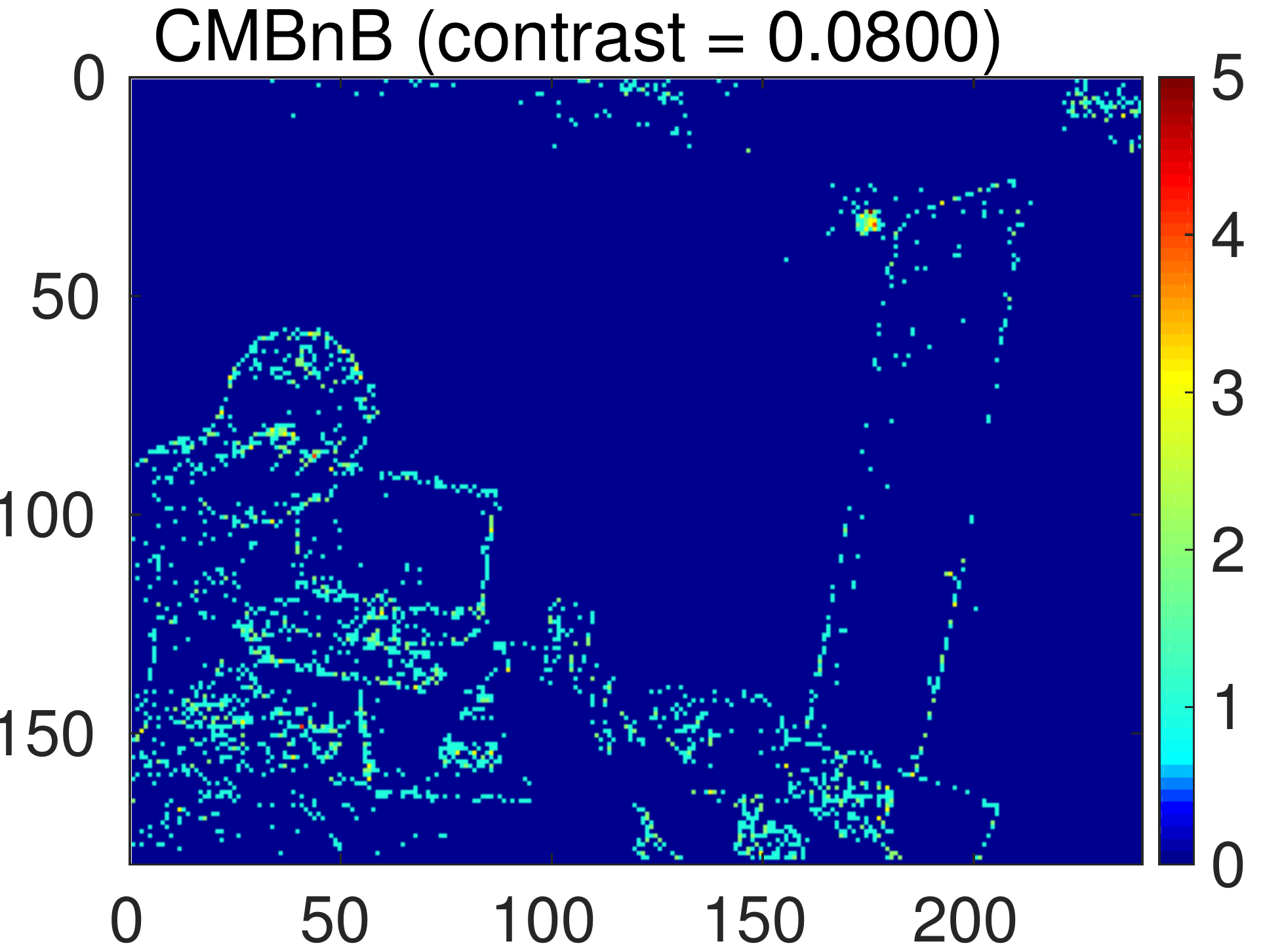}
\includegraphics[width=0.24\textwidth ,height = 0.16\textwidth]{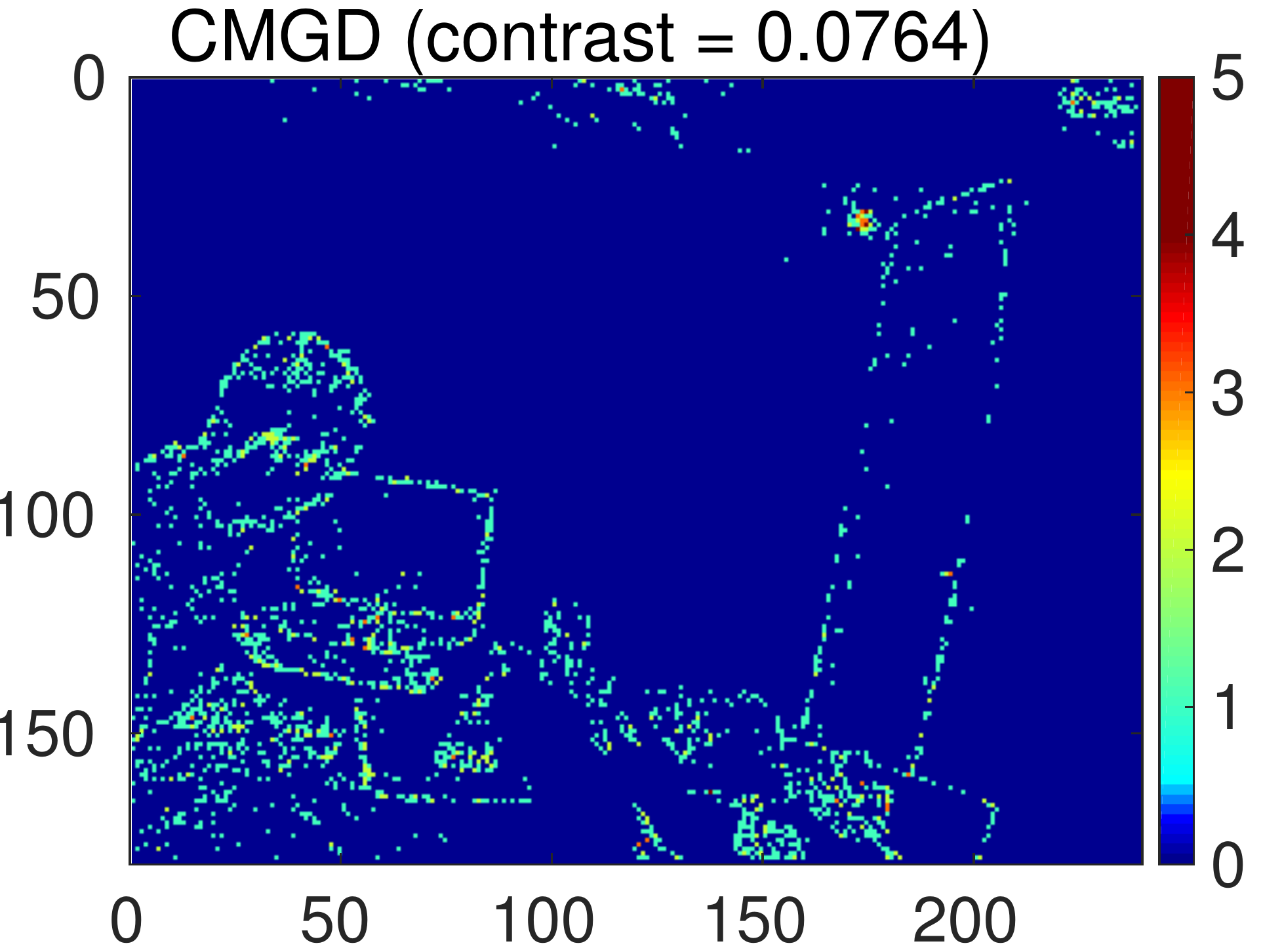}
\includegraphics[width=0.24\textwidth ,height = 0.16\textwidth]{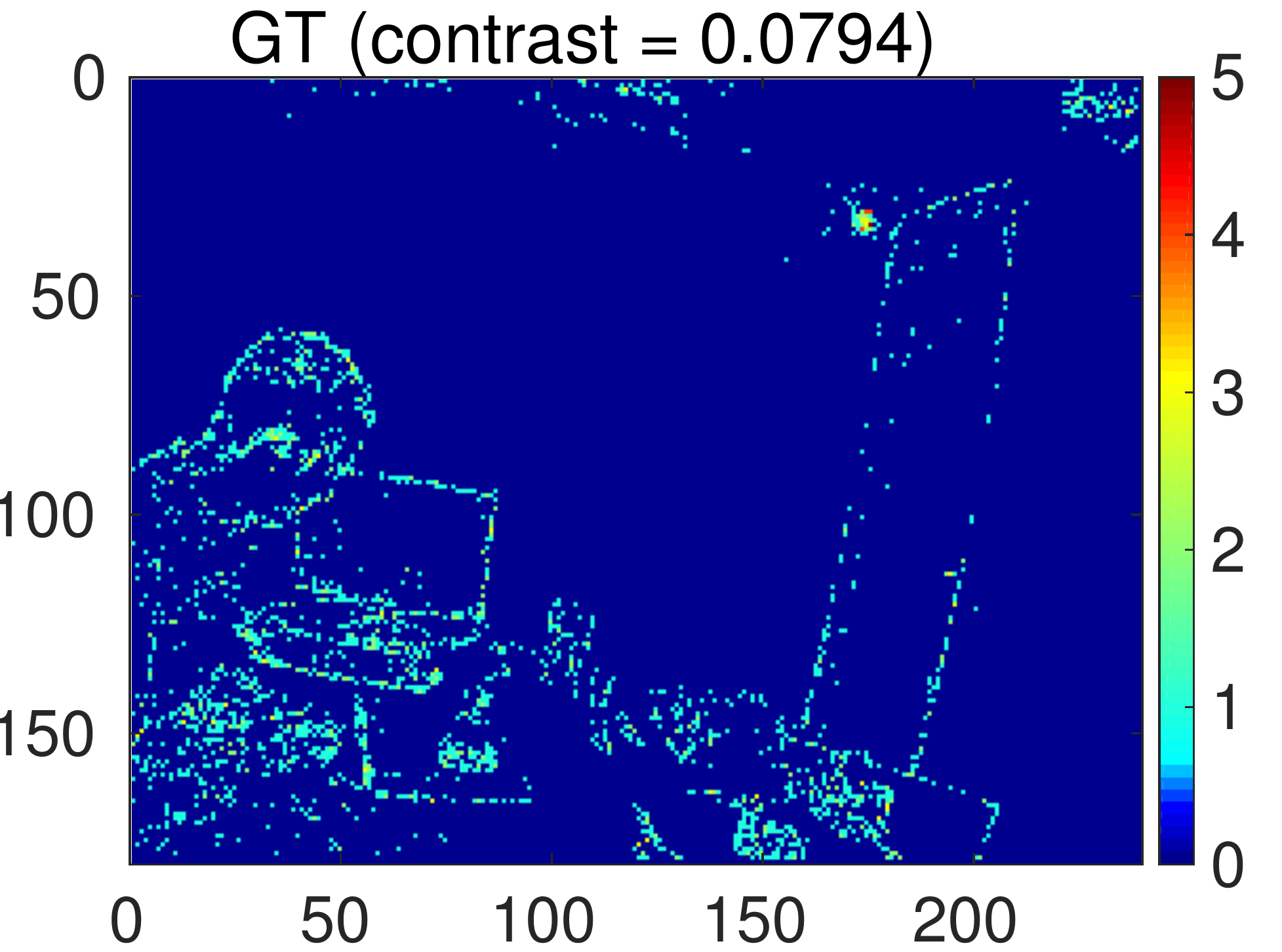}\\

\includegraphics[width=0.24\textwidth ,height = 0.16\textwidth]{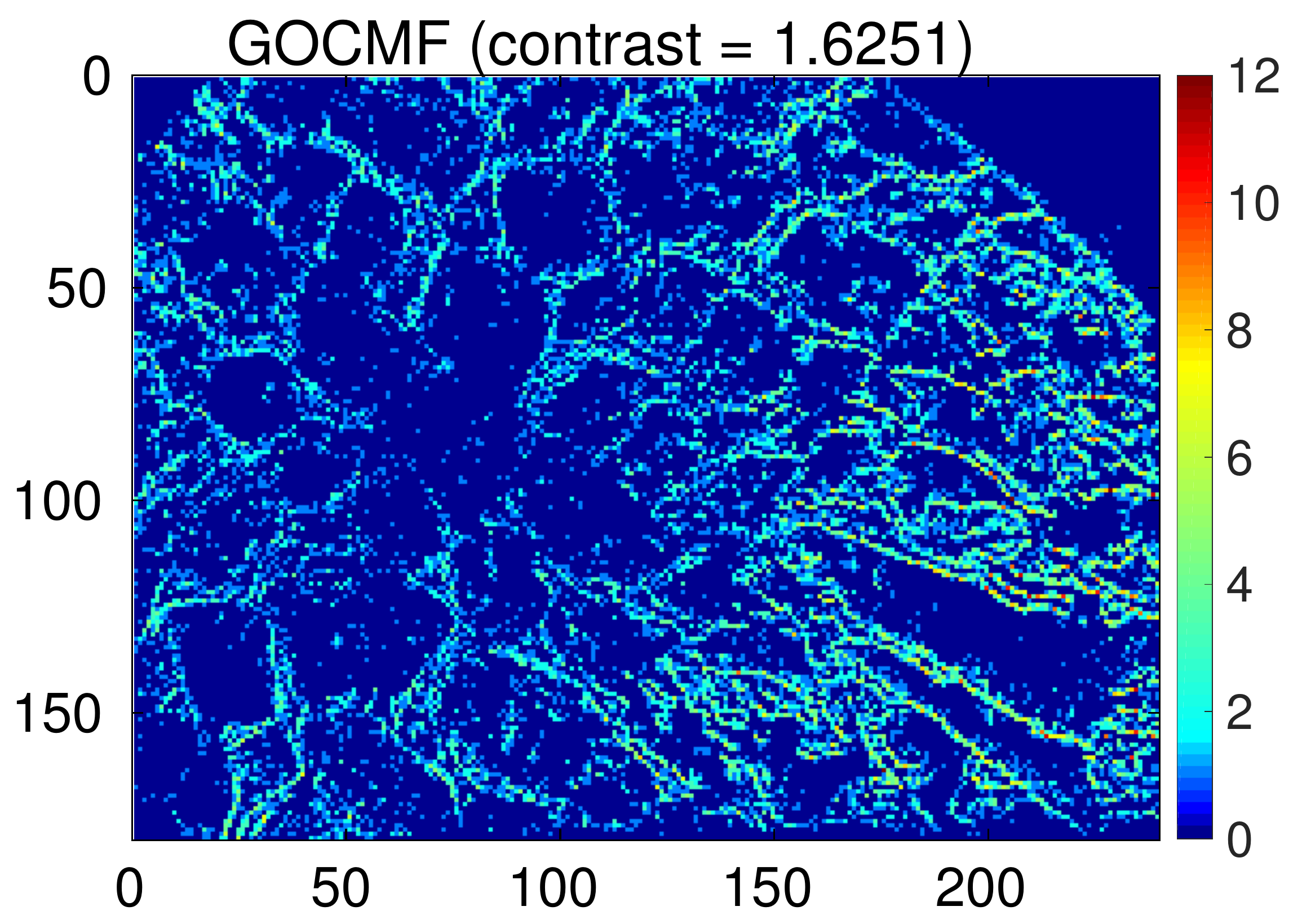}
\includegraphics[width=0.24\textwidth ,height = 0.16\textwidth]{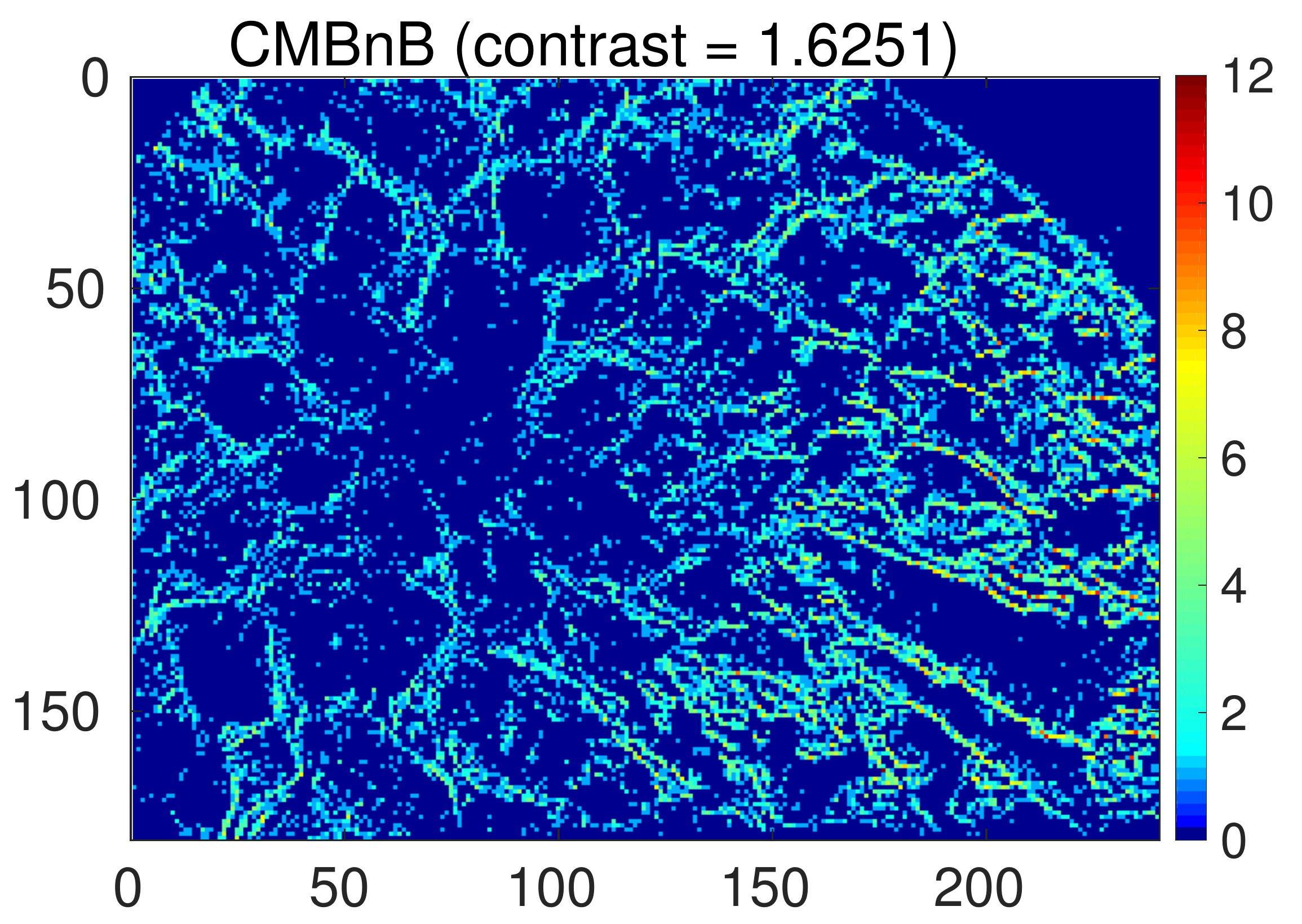}
\includegraphics[width=0.24\textwidth ,height = 0.16\textwidth]{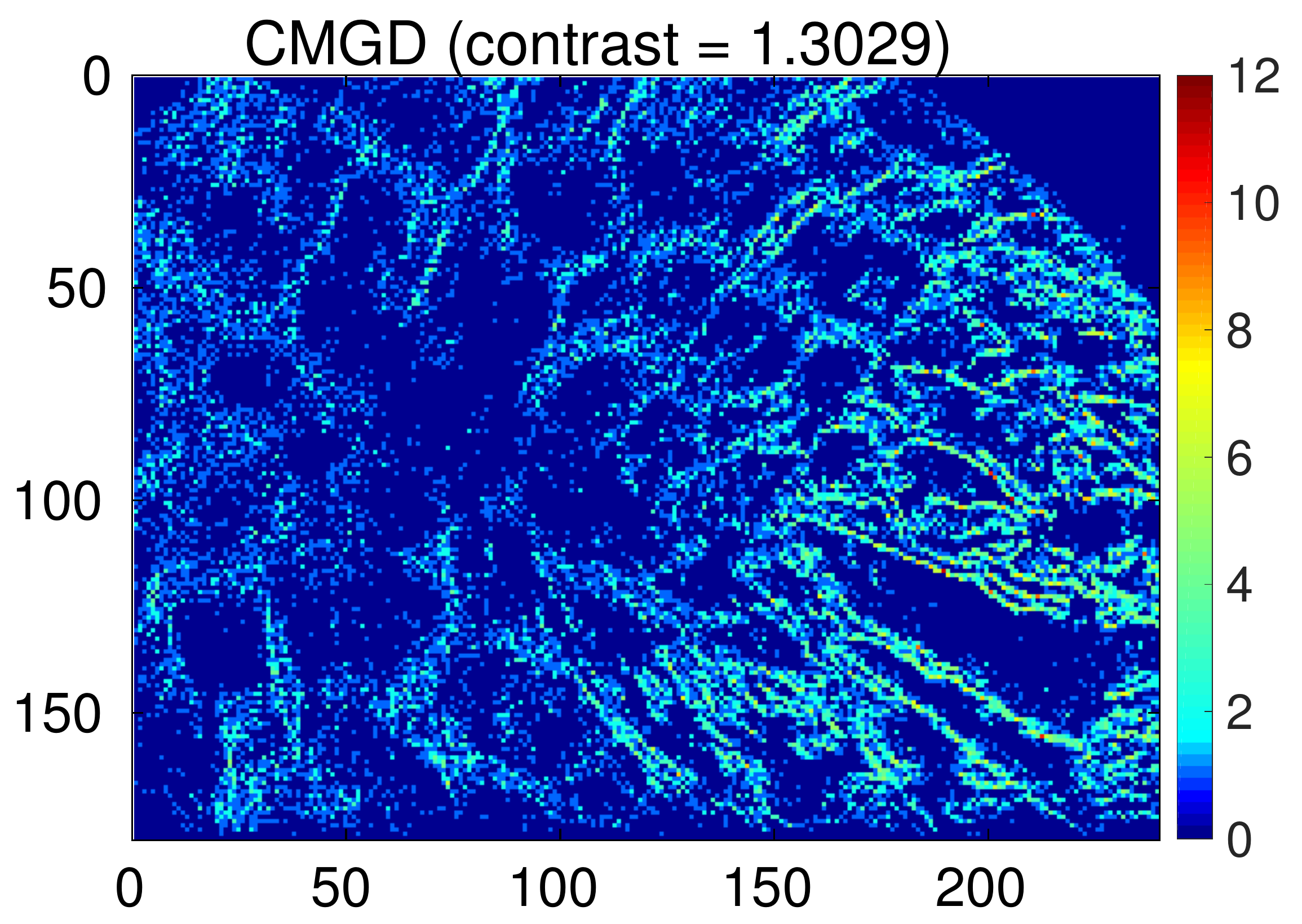}
\includegraphics[width=0.24\textwidth ,height = 0.16\textwidth]{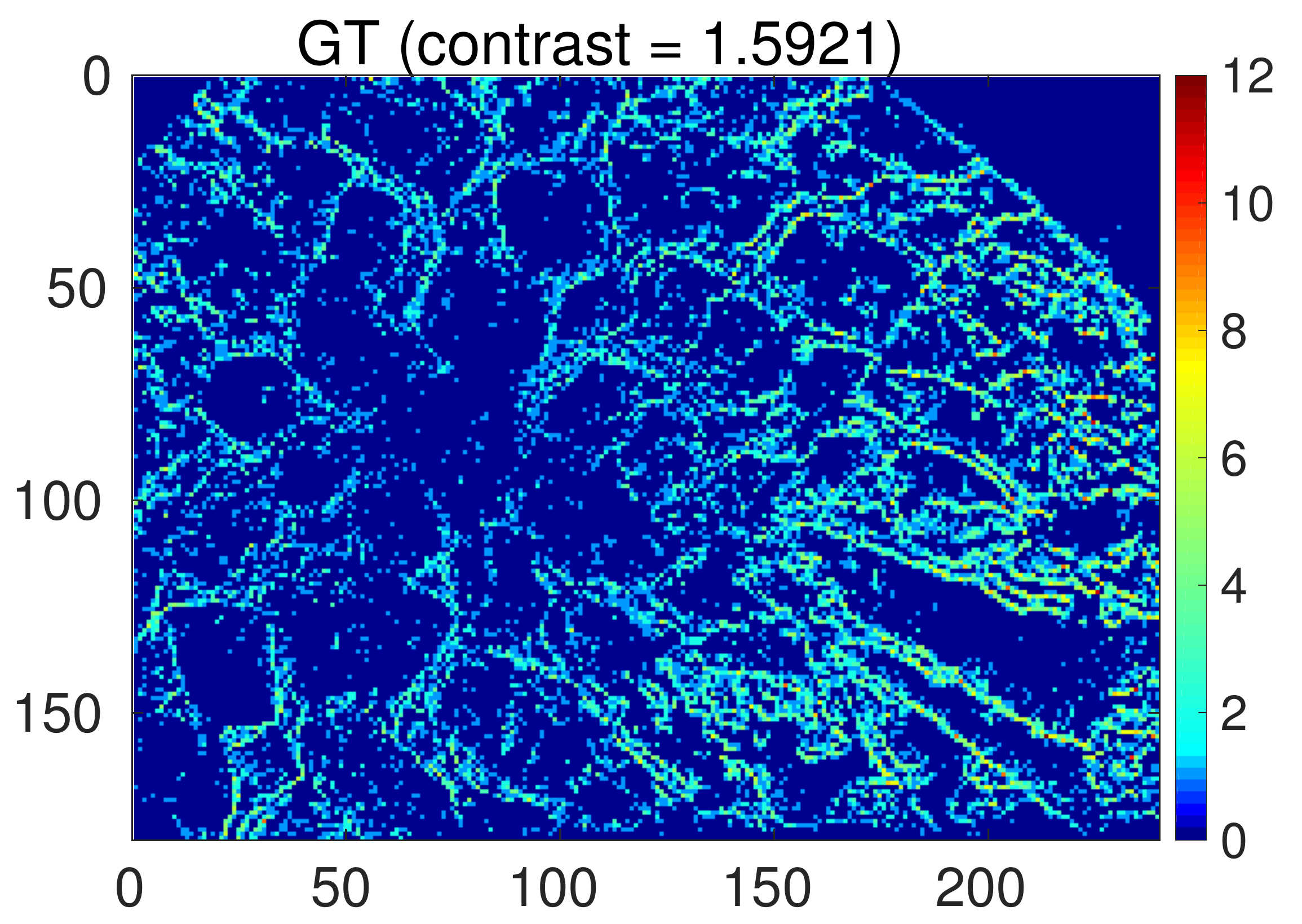}\\

\includegraphics[width=0.24\textwidth ,height = 0.16\textwidth]{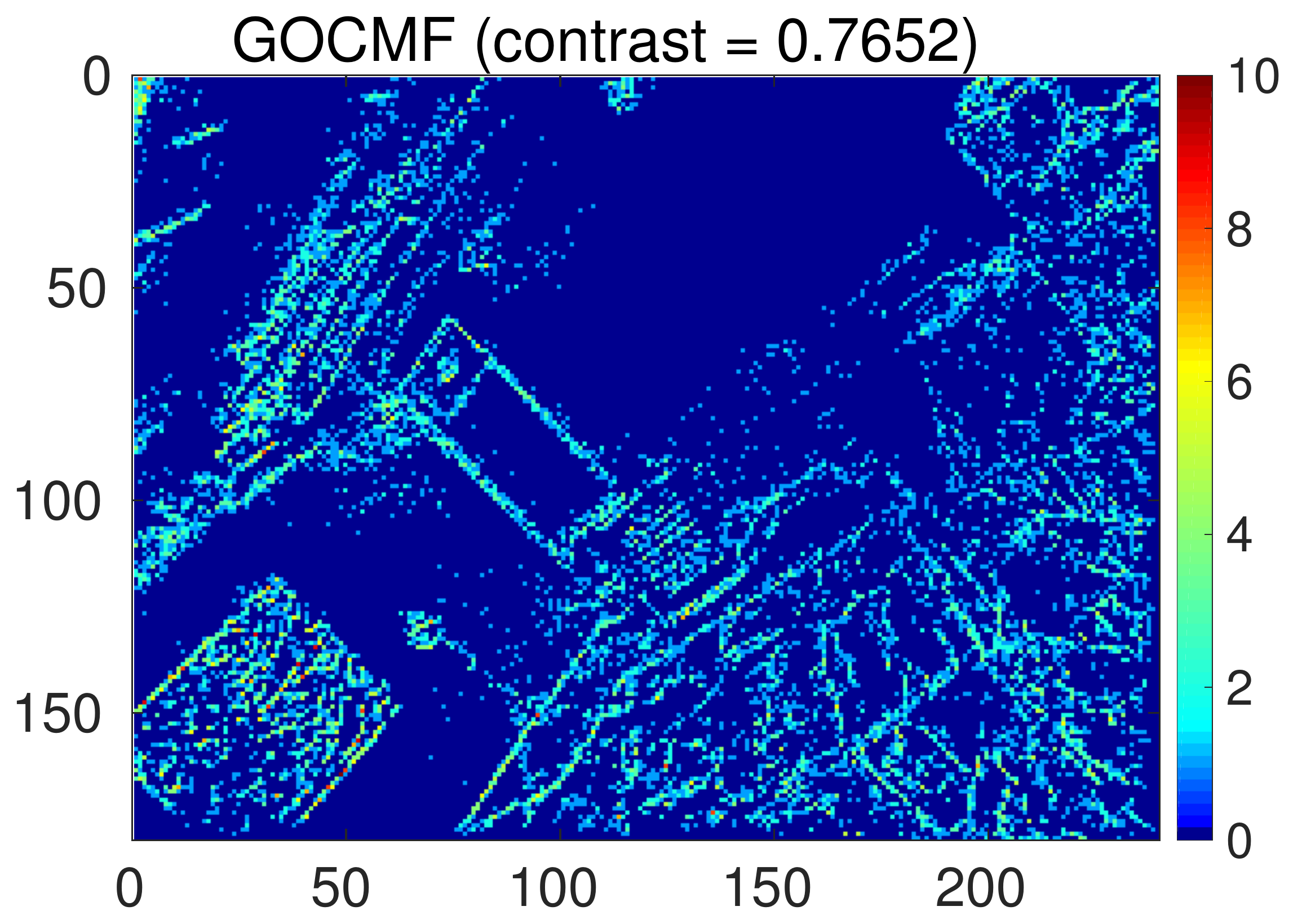}
\includegraphics[width=0.24\textwidth ,height = 0.16\textwidth]{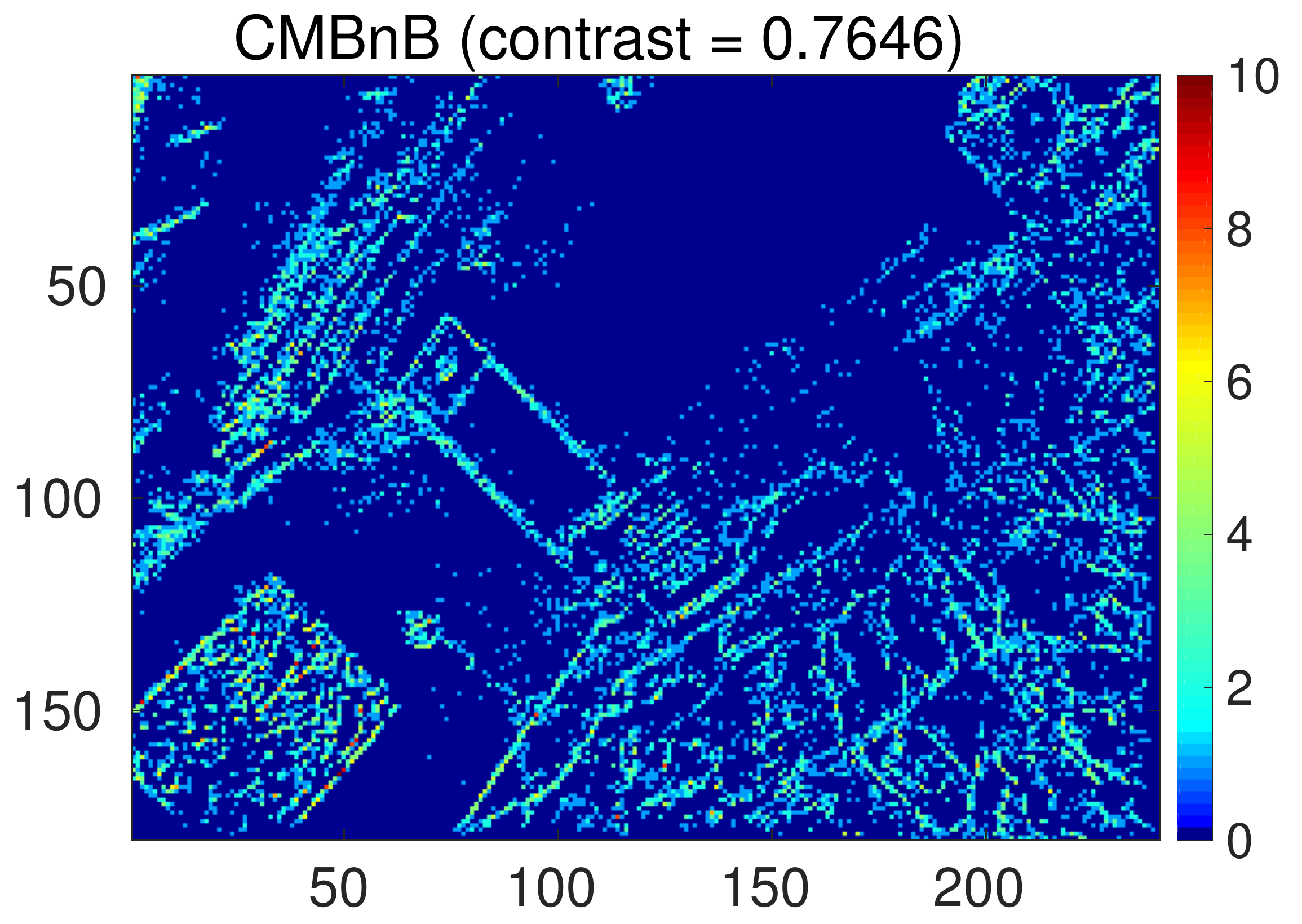}
\includegraphics[width=0.24\textwidth ,height = 0.16\textwidth]{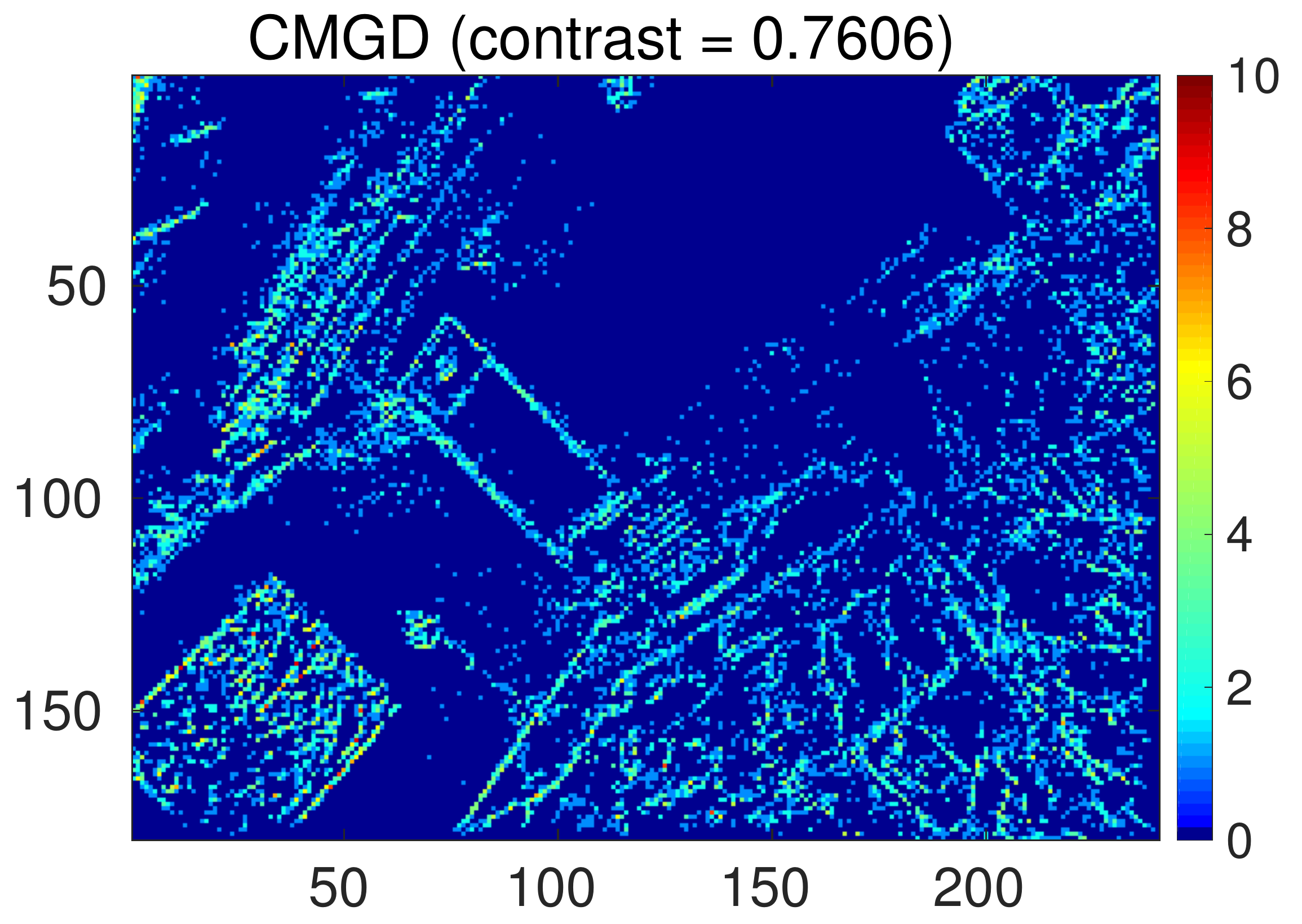}
\includegraphics[width=0.24\textwidth ,height = 0.16\textwidth]{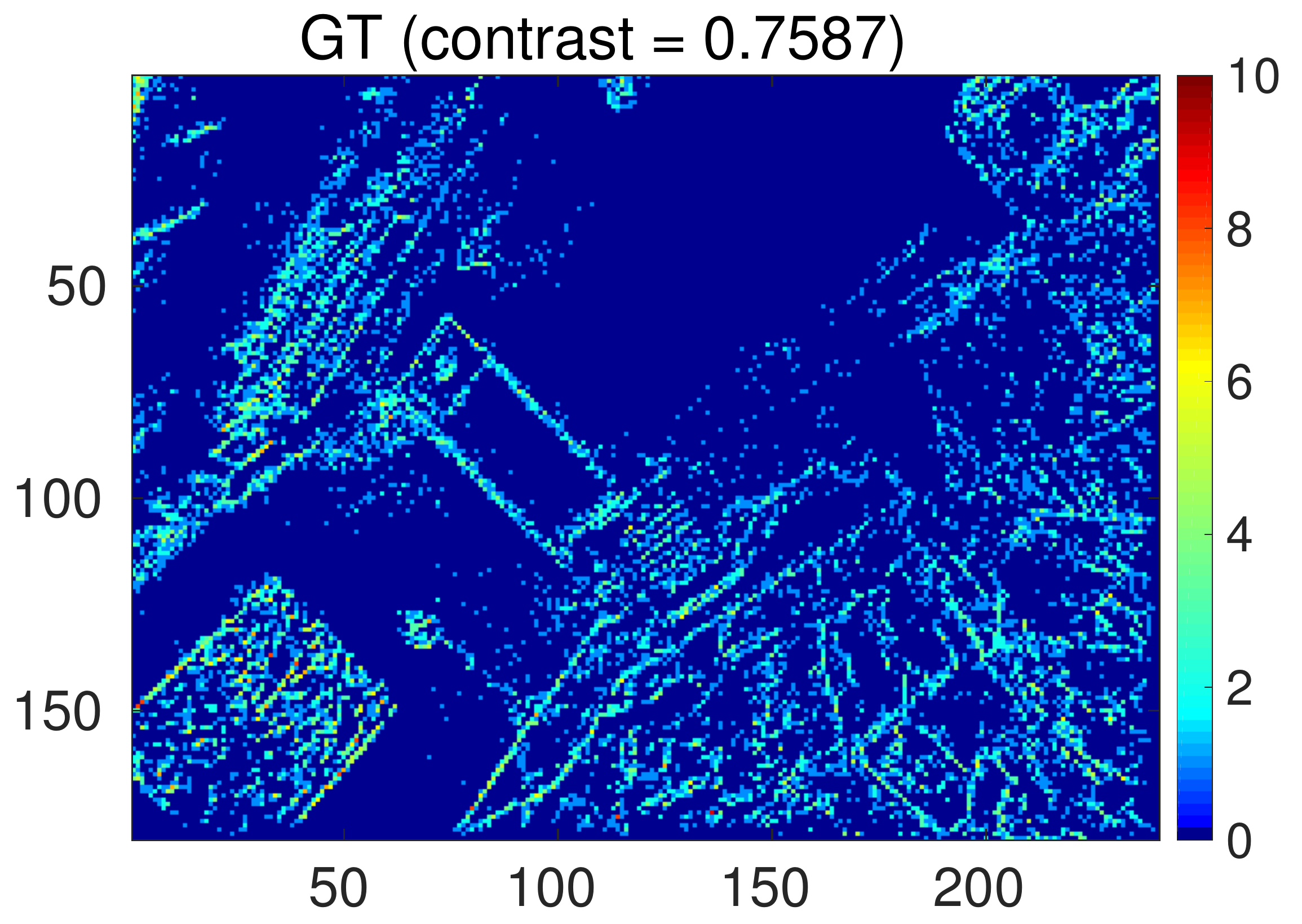}\\

\caption{Images of warped events with anguler velocities estimated by GOCMF, CMBNB, CMGD and groudtruth (GT). The three rows are images using events stream from \textit{dynamic}, \textit{poster} and \textit{boxes} respectively. It is clear that CMGD can indeed often converge to bad local solutions.}
\label{fig:IWE_rotation}
\end{figure*}
Table~\ref{tab:rotation_estimated_errors} presents the average $\mu$ and standard deviation $\sigma$ of $\epsilon$ and $\phi$ as well as the average runtime over all subsequences. Due to its exact nature, globally optimal approaches have lower errors than CMGD. Moreover, our algorithm performs slightly better than CMBNB over datasets \textit{dynamic} and \textit{poster}, whereas CMBNB performs slightly better over \textit{boxes}. Most interestingly, the average runtime of GOCMF is 3.34, 3.07 and 2.49 times faster than CMBNB respectively over \textit{dynamic}, \textit{poster} and \textit{boxes}. The reason is that our efficient, recursively evaluated upper bound proves to be tighter than the upper bound proposed in \cite{liu2020globally} which is further analysed in Section~\ref{sec:Convergence_Analysis}.

Fig.~\ref{fig:IWE_rotation} displays the IWEs with optimal parameters estimated by different approaches.


\section{Analysis}
\label{sec:analysis}

We analyse multiple further aspects of our GOCMF algorithm. We first analyse the precision and the robustness of our contrast maximisation framework. Second, we compare the bound convergence of GOCMF and CMBNB. Third, we speed up our algorithm by implementing a downsampling mechanism. To conclude, we furthermore compare the accuracy of GOCMF under all six possible objective functions.

\subsection{Precision and robustness}

We start by evaluating the precision of motion estimation with the contrast maximisation function $L_{\mathrm{SoS}}$ over synthetic data. As already implied in~\cite{gallego2019focus}, $L_{\mathrm{SoS}}$ can be considered as a solid starting point for the evaluation. Our synthetic data consists of randomly generated horizontal and vertical line segments on a plane at a depth of 2.0m. We consider Ackermann motion with an angular velocity $\omega = 28.6479^{\circ}$/s (0.5rad/s) and a linear velocity $v = 0.5$m/s. Events are generated by randomly choosing a 3D point on a line, and reprojecting it into a random camera pose sampled by a random timestamp within the interval $[0, 0.1\text{s}]$. The result of our method is finally evaluated by running BnB over the search space $\mathcal{W}=[0.4,0.6]$ and $\mathcal{V}=[0.4,0.6]$, and comparing the retrieved solution against the result of an exhaustive search with sampling points every $\delta \omega=0.001$rad/s and $\delta v=0.001$m/s. The experiment is repeated 1000 times.

\begin{figure}[t!]
\centering
\subfigure[]{
\includegraphics[width=0.22\textwidth]{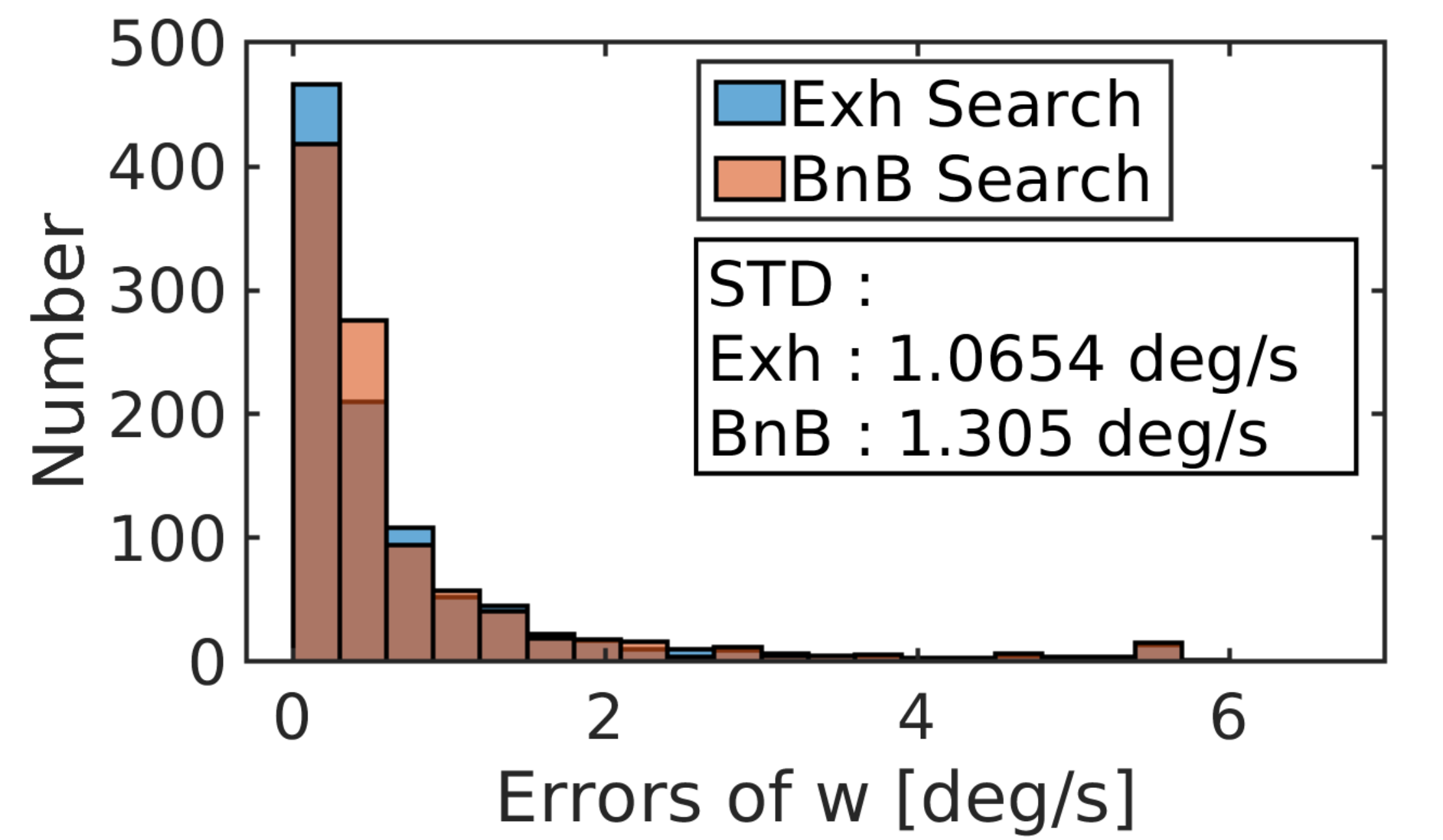}
\label{fig:CM_Error_w}
}
\subfigure[]{
\includegraphics[width=0.22\textwidth]{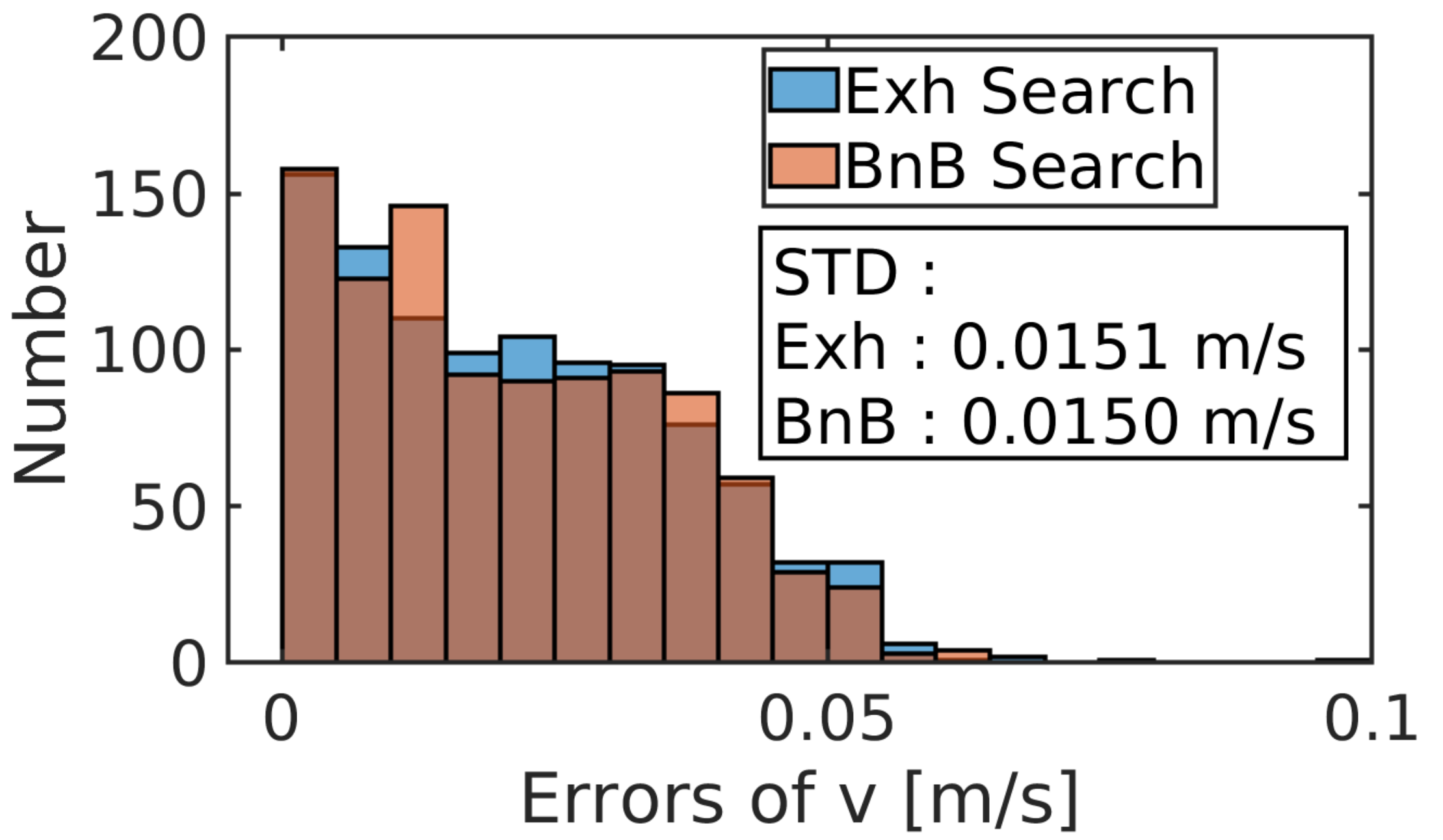} 
\label{fig:CM_Error_v}
}
\subfigure[]{
\includegraphics[width=0.22\textwidth]{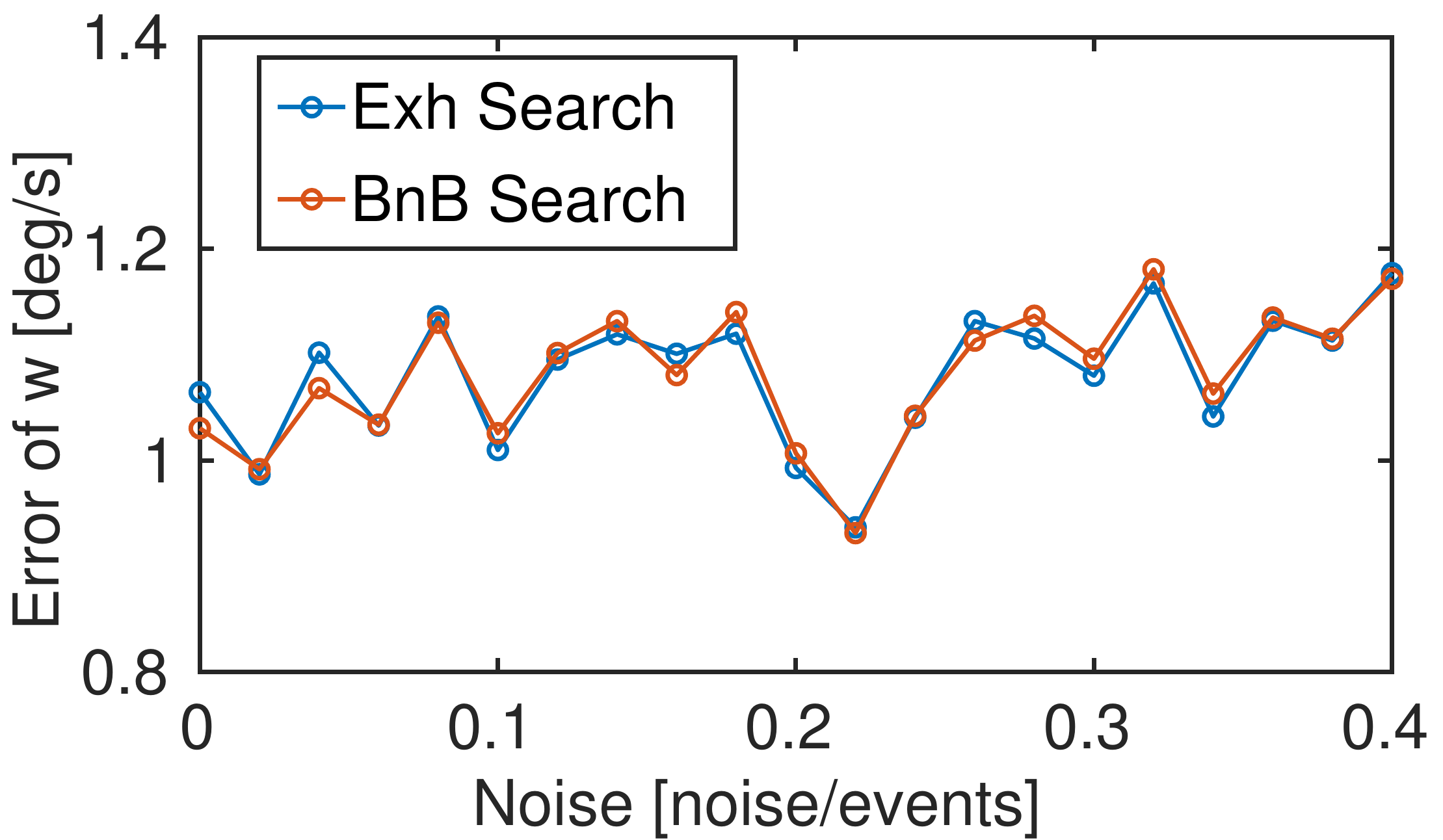} 
\label{fig:CM_Error_Noise_w}
}
\subfigure[]{
\includegraphics[width=0.22\textwidth]{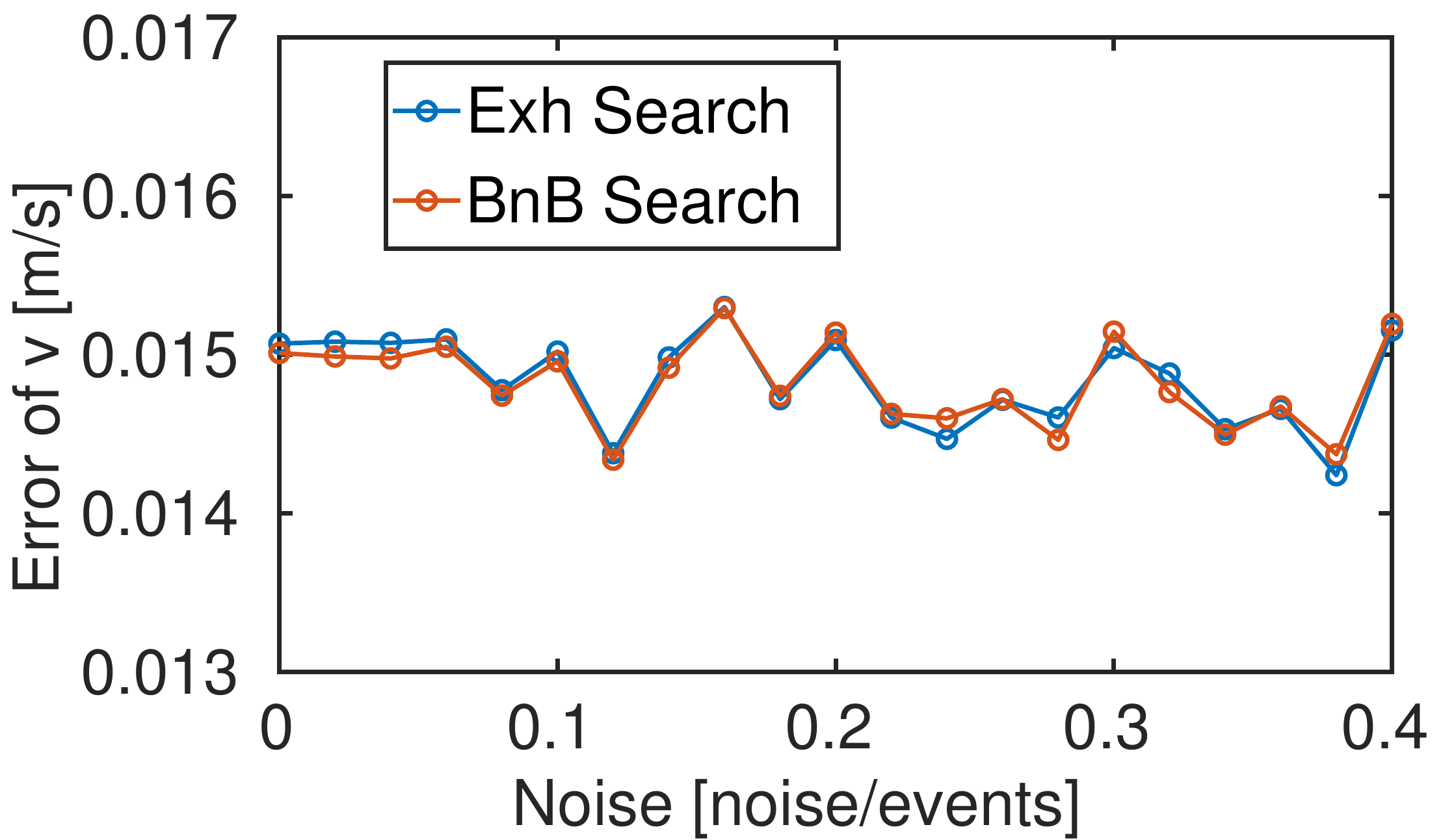}
\label{fig:CM_Error_Noise_v}
}
\caption{Simulation Results. (a) and (b) indicate the error distribution for $\omega$ and $v$ over all experiments for both our proposed method as well as an exhaustive search. (c) and (d) visualise the average error of the estimated parameters caused by additional salt and pepper noise on the event stream. Results are averaged over 1000 random experiments. Note that our proposed method has excellent robustness even for N/E ratios up to 40\%.}
\label{fig:simulation results}
\end{figure}

Figs.~\ref{fig:CM_Error_w} and \ref{fig:CM_Error_v} illustrate the distribution of the errors for both methods in the noise-free case. The standard deviation of the exhaustive search and BnB are $\sigma_{\omega}=1.0645^{\circ}$/s, $\sigma_{v}=0.0151$m/s and $\sigma_{\omega}=1.305^{\circ}$/s, $\sigma_{v}=0.0150$m/s, respectively. While this result suggests that BnB works well and sustainably returns a result very close to the optimum found by exhaustive search, we still note that the optimum identified by both methods has a bias with respect to ground truth, even in the noise-free case. Note however that this is related to the nature of the contrast maximisation function, and not our globally optimal solution strategy.

In order to analyse robustness, we randomly add salt and pepper noise to the event stream with noise-to-event (N/E) ratios between 0 and 0.4 (Example objective functions for different N/E ratios have already been illustrated in Figs.~\ref{fig:heat map}). Fig.~\ref{fig:CM_Error_Noise_w} and \ref{fig:CM_Error_Noise_v} show the error for each noise level again averaged over 1000 experiments. As can be observed, the errors are very similar and behave more or less independently of the amount of added noise. The latter result underlines the high robustness of our approach.

\subsection{Convergence}
\label{sec:Convergence_Analysis}
Next, we give a proof that the derived bounds converge as the branches decrease in size. Given a search space $\boldsymbol{\Theta}=[\boldsymbol{\theta}_0-\delta\boldsymbol{\theta},\boldsymbol{\theta}_0+\delta\boldsymbol{\theta}]$, mathematically correct bounds should converge when $\delta\boldsymbol{\theta}\rightarrow 0$. The convergence of the bounds depends on the tightness of the bounding boxes $\mathcal{P}^{\boldsymbol{\Theta}}_{N}$, which varies depending on the considered scenario. However, when $\delta\boldsymbol{\theta}\rightarrow 0$, $\mathcal{P}^{\boldsymbol{\Theta}}_{N}$ will tend to be equal to the single pixel given by $W(\mathbf{x}_N,t_N;\boldsymbol{\theta}_0)$. Hence---according to equation (\ref{eq:upper_bound})---the upper bound becomes
\begin{eqnarray}
    \lim_{\delta\boldsymbol{\theta}\rightarrow 0} \overline{L_N} &=& \lim_{\delta\boldsymbol{\theta}\rightarrow 0} \overline{L_{N-1}}+1+2Q^{N-1} \nonumber \\
    & = & \lim_{\delta\boldsymbol{\theta}\rightarrow 0} \overline{L_{N-1}}+1+2\lim_{\delta\boldsymbol{\theta}\rightarrow 0} \max_{\substack{\mathbf{p}_{ij}\in\mathcal{P}_{N}^{\boldsymbol{\Theta}}}} \overline{I}^{N-1}(\mathbf{p}_{ij}) \nonumber \\
    & \approx & \lim_{\delta\boldsymbol{\theta}\rightarrow 0} \overline{L_{N-1}}+1+2 \overline{I}^{N-1}(\boldsymbol{\eta}_{N}^{\boldsymbol{\theta}_0};\boldsymbol{\theta}_{0})
\end{eqnarray}
By using mathematical induction, it is again relatively straightforward to prove that $\lim_{\delta\boldsymbol{\theta}\rightarrow 0} \overline{L_N} \rightarrow \underline{L_N}$ and $\overline{I}^{N} \rightarrow \underline{I}^{N}$.

\begin{figure}
    \centering
    \includegraphics[width = 0.24\textwidth]{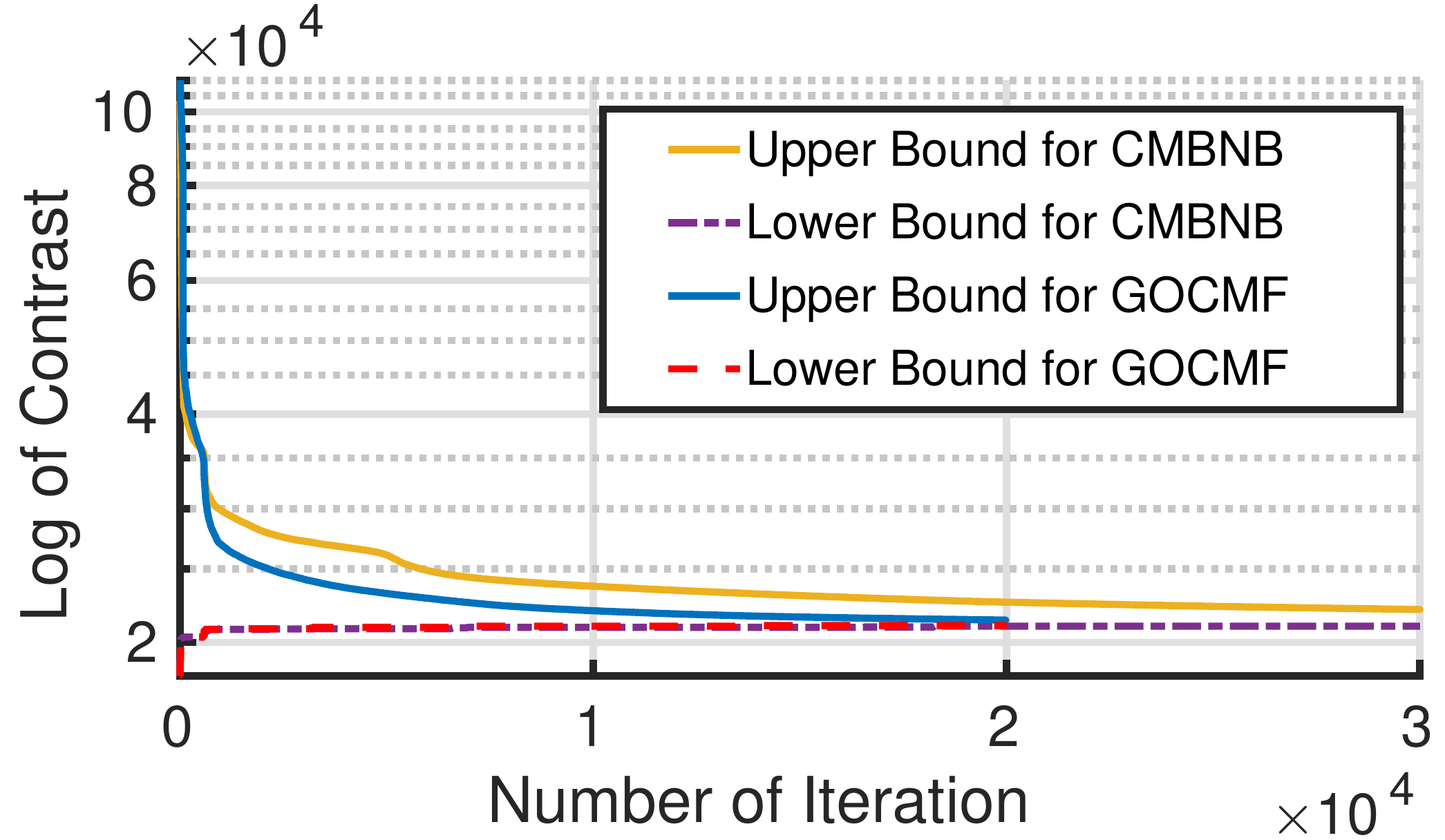}
    \includegraphics[width = 0.24\textwidth]{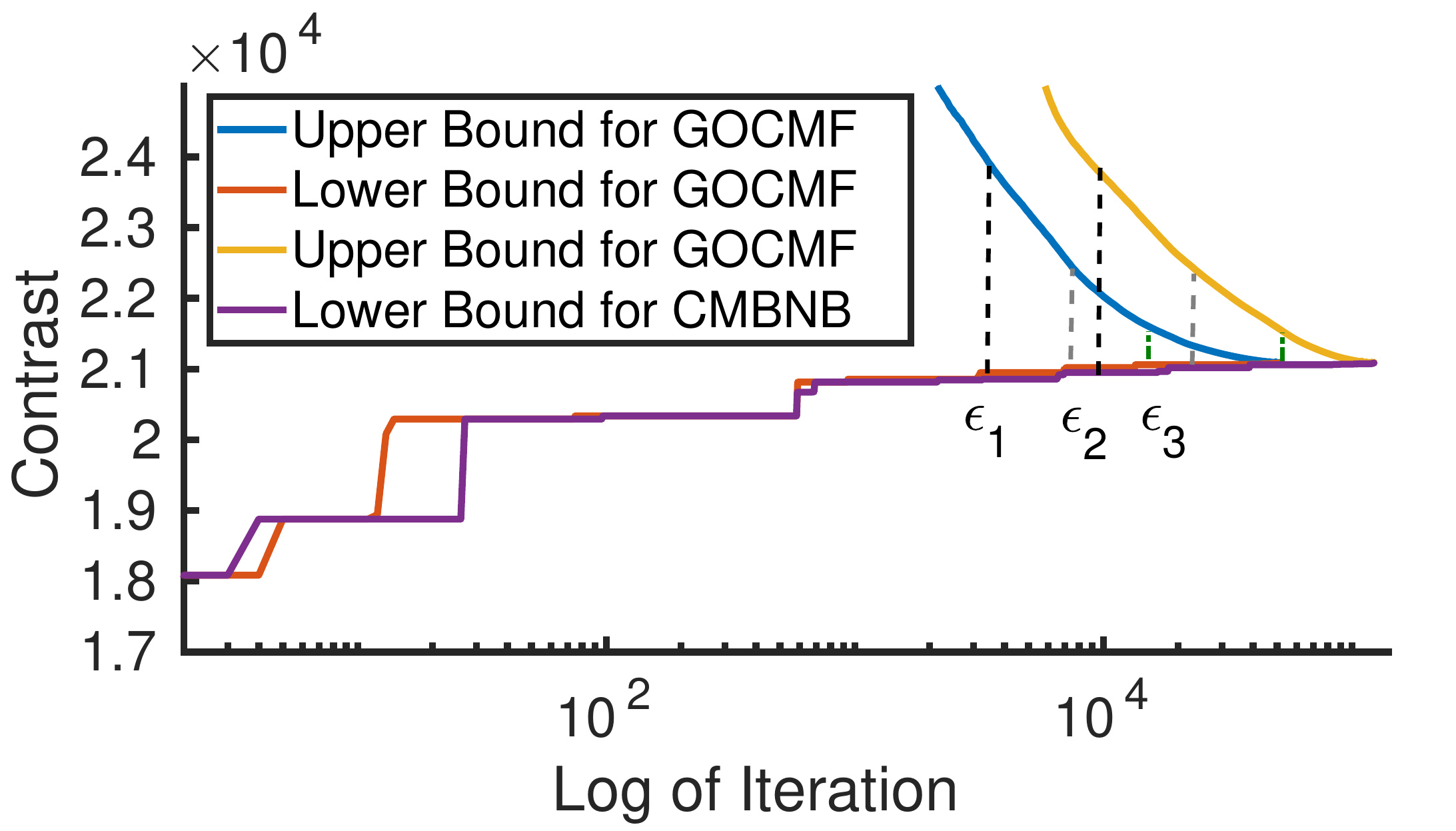}
    \caption{Upper and lower bound evolution in GOCMF and CMBNB. It is obvious that GOCMF converges faster than CMBNB which terminated with at least 30000 iterations, while GOCMF terminated at about 20000 iterations, illustrated in the left figure. The difference in performance is due to the much tighter bounding of the GOCMF. To indicate the iteration of the lower bounds, we exhibit the right figure with log of iteration. $\epsilon$ means the chosen convergence criterion - gap between the upper and lower bound. The smaller the $\epsilon$, the more iteration and more accurate.}
    \label{fig:convergence}
\end{figure}

The convergence speed of the upper and lower bounds indicates the efficiency of the BnB paradigm and the tightness of the bounds themselves. To visualize and empirically compare the convergence of the two globally optimal algorithms GOCMF and CMBNB, we plot the evolution of the upper and lower bounds within a 10 ms subsequence of the \textit{boxes} data. The result is shown in Fig.~\ref{fig:convergence}. From a theoretical point of view, the BnB framework terminates when the gap between the upper and lower bound reduces to zero. As we can see in Fig.~\ref{fig:convergence}, the gap between the upper and lower bounds decreases slowly. Hence, for practical applications we set the convergence criterion to a non-zero gap $\epsilon$ between the upper and lower bound. The parameter $\epsilon$ generates a trade-off between accuracy and computational efficiency. The specific value for epsilon is different in each experiment and chosen automatically as a function of the average number of events in a given time interval. For example, the gap is set to 3000 for the sequence \textit{boxes}.

As illustrated in Fig.~\ref{fig:convergence}, it is obvious that GOCMF converges significantly faster than CMBNB. The latter terminates after at least 30000 iterations, while GOCMF terminates already after about 20000 iterations. This is consistent with Table~\ref{tab:rotation_estimated_errors} in which the runtime indicated for GOCMF is about 3 times faster than the one for CMBNB. The difference in performance indicates tighter (recursive) bounds for GOCMF.
\subsection{Event Downsampling}

An event camera is a low-latency sensor that asynchronously outputs several million events per second which is challenging to process, especially for a CPU-based implementation. This adds to the anyway expensive nature of the globally-optimal BnB algorithm. We therefore implement a preprocessing step to speed up the algorithm. Contrast maximisation evaluates the sharpness of the IWE, a property that owns a certain robustness against downsampling of the event stream. We evaluate the accuracy of local optimisation (CMGD) and GOCMF over the last 15s of sequence \textit{boxes} with different downsampling factors. A downsampling factor of $m$ simply means that---within the temporally ordered sequence of events---only every $m$-th event is maintained. In this experiment, we test downsampling factors ranging from 1 to 10.

\begin{figure}
    \centering
    \subfigure[errors with decreasing sample rate]{
        \includegraphics[width = 0.23\textwidth]{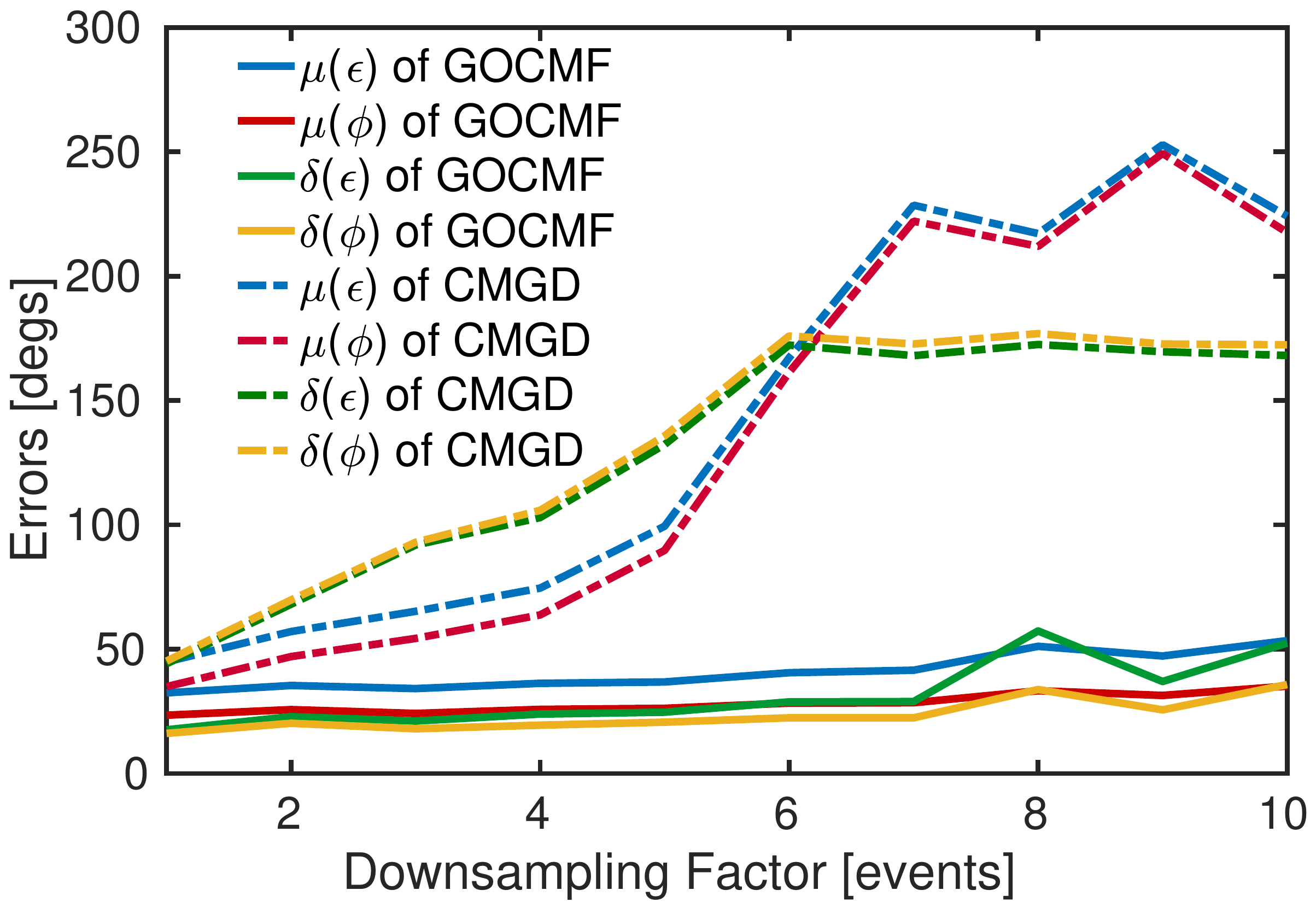}
        \label{fig:downsample_errors}
    }
    \subfigure[average runtime with decreasing sample rate]{
    \includegraphics[width = 0.23\textwidth]{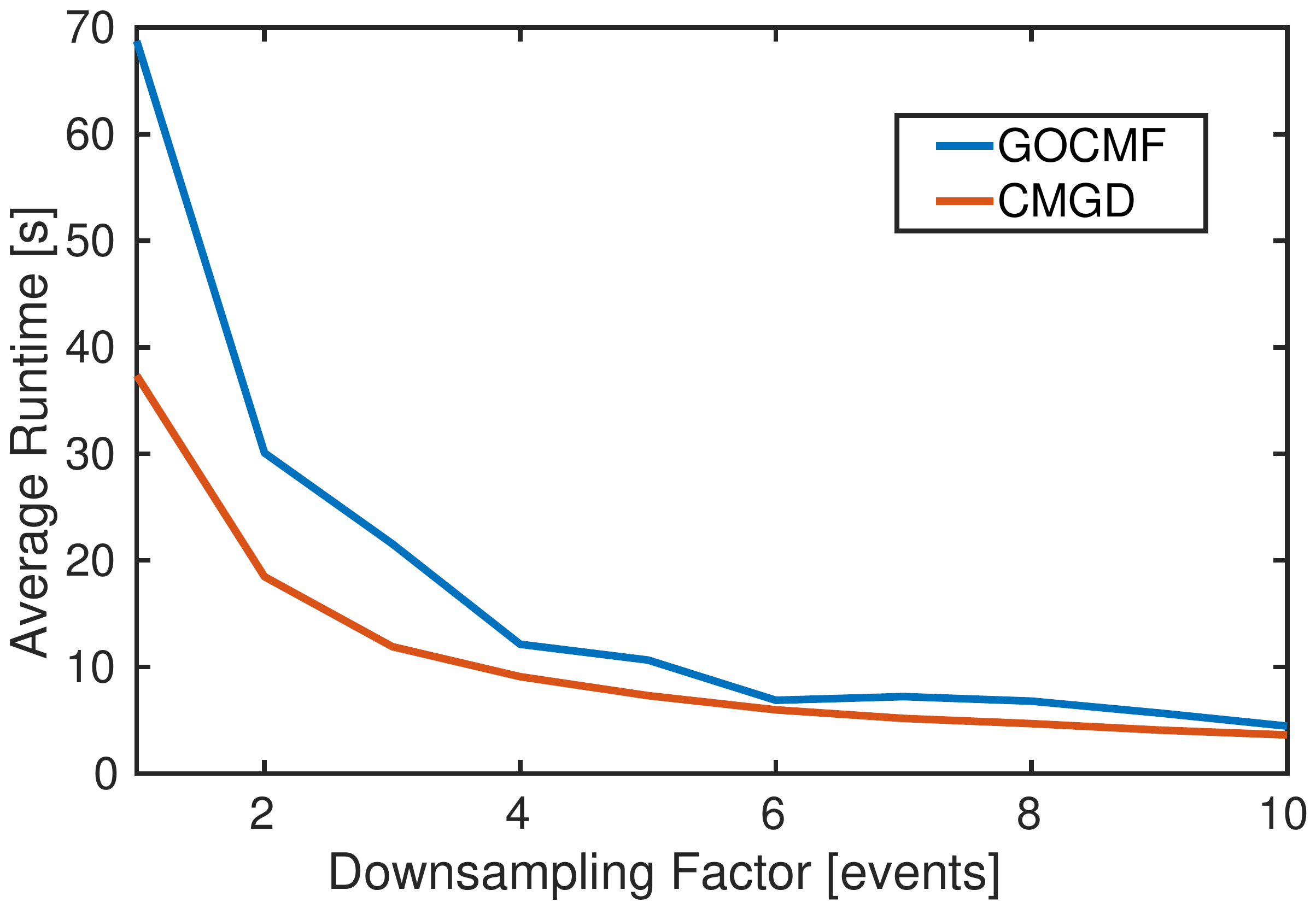}
    \label{fig:downsample_runtime}
    }
    \caption{Downsampling results. (a) shows the errors of GOCMF and local optimisation with increasing downsampling factors. GOCMF always proves more robust than CMGD. (b) illustrates the average runtime of GOCMF and CMGD. An interesting phenomenon is that the runtime of the two approaches tends to be consistent. }
    \label{fig:downsample}
\end{figure}

Fig.~\ref{fig:downsample_errors} illustrates the errors of GOCMF and local optimisation with increasing downsampling factors. It is clear that errors increase as the downsampling factor increases. However, GOCMF remains much more robust than CMGD when compared in terms of the two error metrics $\epsilon$ and $\phi$. The mean error $\phi$ of CMGD is in the range of $[30^\circ, 250^\circ]$, while GOCMF stays within the range $[25^\circ, 55^\circ]$. The error of GOCMF is stable for a downsampling factor between 1 and 6. Fig.~\ref{fig:downsample_runtime} furthermore indicates the average runtime of GOCMF and CMGD. The runtime decreases exponentially as the sampling rate decreases. An interesting phenomenon is that the runtime of the two approaches tends to be identical. As a tradeoff between accuracy and runtime, all aforementioned experiments in this paper use a down-sampling factor of 2.  

\subsection{Different objective functions}
The proposed recursive bounds for our globally-optimal contrast maximisation framework handle a total number of six different focus loss functions. We test our algorithm with all aforementioned six contrast functions over various types of motions, including a straight line, a circle, and an arbitrarily curved trajectory (datasets used in Section~\ref{sec:downward-facing}). Table~\ref{tab:errors} shows the RMS errors of the estimated dynamic parameters, and compares the accuracy of all six alternatives. As can be observed, $L_{\mathrm{SoS}}$ and $L_{\mathrm{Var}}$ perform well, but the best performance is given by $L_{\mathrm{SoSAaS}}$.
\begin{table}[t]
\renewcommand\arraystretch{1.5}
\caption{RMS errors for different datasets and methods}
\centering
\setlength{\tabcolsep}{1.4mm}
\begin{tabular}{ccccccc}
\toprule
\textbf{Method} 
& {\begin{tabular}[c]{@{}c@{}}\textit{Line}\\ w {[}$^{\circ}$/s{]}\end{tabular}} 
& {\begin{tabular}[c]{@{}c@{}}\textit{Line}\\ v {[}m/s{]}\end{tabular}} 
& {\begin{tabular}[c]{@{}c@{}}\textit{Circle}\\ w {[}$^{\circ}$/s{]}\end{tabular}} 
& {\begin{tabular}[c]{@{}c@{}}\textit{Circle}\\ v {[}m/s{]}\end{tabular}} 
& {\begin{tabular}[c]{@{}c@{}}\textit{Curve}\\ w {[}$^{\circ}$/s{]}\end{tabular}} 
& {\begin{tabular}[c]{@{}c@{}}\textit{Curve}\\ v {[}m/s{]}\end{tabular}}\\ \midrule

SoE     & 2.408      & 0.015     & 2.212     & 0.025     & 3.628 & 0.026  \\ 
SoEaS   & 2.405      & 0.015    & 2.017     & 0.024     & 3.628 & 0.026  \\ 
SoS     & \textbf{0.512}      & \textbf{0.008}   & 1.088  & 0.008    & 3.009     & 0.021  \\ 
SoSA    & 1.961      & 0.028           & 4.249   & 0.073    & 9.290    & 0.072  \\ 
SoSAaS  & \textbf{0.517}  & \textbf{0.008}   & \textbf{0.529}   & \textbf{0.004}  & \textbf{0.554} & \textbf{0.018}                                               \\ 
Var     & \textbf{0.512}      & \textbf{0.008}   & 1.088   & 0.008    & 3.009   & 0.020  \\  \bottomrule
\end{tabular}
\label{tab:errors}
\end{table}

\section{Conclusion}
\label{sec:conclusion}
We have introduced a novel globally optimal solution to contrast maximisation for homographically rectified event streams. We have successfully applied this to three different scenarios, including optical flow and pure rotation estimation. To the best of our knowledge, we are the first to apply the idea of homography estimation via contrast maximisation to the real-world case of non-holonomic motion estimation with a downward facing event camera mounted on an AGV. The challenging conditions in these scenarios advertise the use of event cameras. Our globally optimal solutions are crucial for successful contrast maximisation, and significantly outperform incremental local refinement. As shown by our results, the planar motion estimation scenario ultimately favors dynamic vision sensors over regular frame-based cameras.

\ifCLASSOPTIONcompsoc
  \section*{Acknowledgments}
\else
  \section*{Acknowledgment}
\fi

The authors would like to thank Prof. Kyros Kutulakos for his kind advice and Mr. Daqi Liu who kindly offered his code. The authors would also like to thank the fundings sponsored by Natural Science Foundation of Shanghai (grant number: 19ZR1434000) and Natural Science Foundation of China (grant number: 61950410612).

\begin{table*}[ht]
\caption{Bounding Box Cases}
\centering
\begin{tabular}{|c|c|c|c|c|c|c|c|c|}
\hline
\multicolumn{3}{|c|}{\multirow{2}{*}{Conditions}}  & \multicolumn{3}{c|}{ $\underline{x^{\prime}_{k}}$}                         & \multicolumn{3}{c|}{ $\underline{y^{\prime}_{k}}$} \\ \cline{4-9} 
\multicolumn{3}{|c|}{}    & $a_x$ & $b_x$   & $c_x$ & $a_y$ & $b_y$ & $c_y$ 
\\ \hline
\multirow{16}*{\ $\omega_{\text{min}} > 0$\ } & \multicolumn{1}{c|}{\multirow{8}{*}{$\ \ v_{\text{min}} \geq 0$\ \ }} & \begin{tabular}[c]{@{}l@{}}$\ x_k \geq u_0$,\\ $\ y_k \geq v_0 - l \cdot \frac{f}{d}$\ \ \end{tabular}   &\ $\omega_{\text{max}}\ $     &\ $\omega_{\text{max}}\ $    &\ \begin{tabular}[c]{@{}l@{}}$\omega_{\text{max}}$,\ \ \\ $v_{\text{min}}$\end{tabular}                     &\ $\omega_{\text{min}}$\ \ &\ \begin{tabular}[c]{@{}l@{}}$\omega_{\text{min}}$,\ \ \\ $v_{\text{max}}$\end{tabular}        & \ $\omega_{\text{max}}\ $       \\ \cline{3-9} 
                  & \multicolumn{1}{c|}{}                  & \begin{tabular}[c]{@{}l@{}}$x_k < u_0$,\\ $y_k \geq v_0 - l \cdot  \frac{f}{d}$\end{tabular}  & $\omega_{\text{max}}$     &                      $\omega_{\text{min}}$ & \begin{tabular}[c]{@{}l@{}} $\omega_{\text{max}}$,\\ $v_{\text{min}}$\end{tabular}                      & $\omega_{\text{max}}$         & \begin{tabular}[c]{@{}l@{}}$\omega_{\text{min}}$,\\ $v_{\text{max}}$\end{tabular}        &  $\omega_{\text{max}}$       \\ \cline{3-9} 
                  & \multicolumn{1}{c|}{}                  & \begin{tabular}[c]{@{}l@{}}$x_k < u_0$,\\ $y_k < v_0 - l \cdot \frac{f}{d}$\end{tabular}  &  $\omega_{\text{min}}$    &                      $\omega_{\text{min}}$ &   \begin{tabular}[c]{@{}l@{}}$\omega_{\text{max}}$,\\ $v_{\text{min}}$\end{tabular}                     &  $\omega_{\text{max}}$        &  \begin{tabular}[c]{@{}l@{}}$\omega_{\text{min}}$,\\ $v_{\text{max}}$\end{tabular}                           &  $\omega_{\text{min}}$         \\ \cline{3-9} 
                  & \multicolumn{1}{c|}{}                  & \begin{tabular}[c]{@{}l@{}}$x_k \geq u_0$,\\ $y_k < v_0 - l \cdot \frac{f}{d}$\end{tabular}  & $\omega_{\text{min}}$     &  $\omega_{\text{max}}$           &   \begin{tabular}[c]{@{}l@{}}$\omega_{\text{max}}$,\\ $v_{\text{min}}$\end{tabular}                    &$\omega_{\text{min}}$          &  \begin{tabular}[c]{@{}l@{}}$\omega_{\text{min}}$,\\ $v_{\text{max}}$\end{tabular}        & $\omega_{\text{min}}$        \\ \cline{2-9} 
 & { \multirow{8}{*}{$v_{\text{min}} < 0$}} & \begin{tabular}[c]{@{}l@{}}$x_k \geq u_0$,\\ $y_k \geq v_0 - l \cdot \frac{f}{d}$\end{tabular}   & $\omega_{\text{max}}$     & $\omega_{\text{max}}$    &  \begin{tabular}[c]{@{}l@{}}$\omega_{\text{min}}$,\\ $v_{\text{min}}$\end{tabular}                     & $\omega_{\text{min}}$      & \begin{tabular}[c]{@{}l@{}}$\omega_{\text{min}}$,\\ $v_{\text{max}}$\end{tabular}        &  $\omega_{\text{max}}$       \\ \cline{3-9} 
                  & \multicolumn{1}{c|}{}                  & \begin{tabular}[c]{@{}l@{}}$x_k < u_0$,\\ $y_k \geq v_0 - l \cdot  \frac{f}{d}$\end{tabular}  & $\omega_{\text{max}}$     &                      $\omega_{\text{min}}$ & \begin{tabular}[c]{@{}l@{}} $\omega_{\text{min}}$,\\ $v_{\text{min}}$\end{tabular}                      & $\omega_{\text{max}}$         & \begin{tabular}[c]{@{}l@{}}$\omega_{\text{max}}$,\\ $v_{\text{max}}$\end{tabular}        &  $\omega_{\text{max}}$       \\ \cline{3-9} 
                  & \multicolumn{1}{c|}{}                  & \begin{tabular}[c]{@{}l@{}}$x_k < u_0$,\\ $y_k < v_0 - l \cdot \frac{f}{d}$\end{tabular}  &  $\omega_{\text{min}}$    &                      $\omega_{\text{min}}$ &   \begin{tabular}[c]{@{}l@{}}$\omega_{\text{min}}$,\\ $v_{\text{min}}$\end{tabular}                     &  $\omega_{\text{max}}$        &  \begin{tabular}[c]{@{}l@{}}$\omega_{\text{max}}$,\\ $v_{\text{max}}$\end{tabular}                           &  $\omega_{\text{min}}$         \\ \cline{3-9} 
                  & \multicolumn{1}{c|}{}                  & \begin{tabular}[c]{@{}l@{}}$x_k \geq u_0$,\\ $y_k < v_0 - l \cdot  \frac{f}{d}$\end{tabular}  & $\omega_{\text{min}}$     &  $\omega_{\text{max}}$           &   \begin{tabular}[c]{@{}l@{}}$\omega_{\text{min}}$,\\ $v_{\text{min}}$\end{tabular}                    &$\omega_{\text{min}}$          &  \begin{tabular}[c]{@{}l@{}}$\omega_{\text{max}}$,\\ $v_{\text{max}}$\end{tabular}        & $\omega_{\text{min}}$        \\ \hline
\end{tabular}
\label{tab:bounding box cases}
\end{table*}

\appendix[Application to visual odometry with a downward-facing event camera]
\section{Bounding Box Definition}
Let $t=t_k-t_{\text{ref}}$, we recall that
\begin{eqnarray}
  \mathbf{x}^{\prime}_k & = & W(\mathbf{x}_k,t_k;[\omega\text{ }v]^\mathsf{T}) = \left[ \begin{matrix} x^{\prime}_k & y^{\prime}_k \end{matrix} \right]^\mathsf{T} \\
  x^{\prime}_k & = & 
      - [y_k - v_0 + l \cdot \frac{f}{d}] \sin(\omega t) \nonumber \\
      & & + [x_k - u_0 - \frac{f}{d} \cdot \frac{v}{w}] \cos(\omega t)   
      +  \frac{f}{d} \cdot \frac{v}{w} + u_0 , \nonumber \\
      &=& a_x + b_x + c_x + u_0 \nonumber\\
  y^{\prime}_k & = & [x_k - u_0 - \frac{f}{d} \cdot  \frac{v}{w}] \sin(\omega t) \nonumber \\
      & & + [y_k - v_0 + l \cdot \frac{f}{d}] \cos(\omega t) - l \cdot \frac{f}{d} + v_0 . \nonumber \\
      &=&  a_y + b_y + c_y - l \cdot \frac{f}{d} + v_0
\end{eqnarray}
where
\begin{eqnarray}
    a_x &=& - [y_k - v_0 + l \cdot \frac{f}{d}] \sin(\omega t), \nonumber \\
    b_x  &=& [x_k - u_0 ] \cos(\omega t) , \nonumber \\
    c_x  &=& \frac{f}{d} \cdot \frac{v}{\omega} [1- \cos(\omega t)] , \nonumber \\
    a_y &=& [x_k - u_0 ] \sin(\omega t), \nonumber \\
    b_y &=&  - \frac{f}{d} \cdot \frac{v}{\omega} \sin(\omega t), \nonumber \\
    c_y &=&  [y_k - v_0 + l \cdot \frac{f}{d}] \cos(\omega t).
\end{eqnarray}
The bounding box $\mathcal{P}_k^{\boldsymbol{\Theta}}$ over the intervals $\omega\in\mathcal{W}=\left[\omega_{\text{min}},\omega_{\text{max}} \right]$ and $v\in\mathcal{V}=\left[v_{\text{min}} , v_{\text{max}} \right]$. Here we only consider the case $|\omega t| < \pi/2$. The bounding box is easily achieved if simply considering the monotonicity and different cases. There are 17 cases in total. One special case is when $\omega = 0$. Given the Ackermann motion model, we then obtain 
\small
\begin{equation}
  \underline{x^{\prime}_k} = x_k,\ \overline{x^{\prime}_k}  = -x_k, 
\end{equation}
\begin{equation}
    \underline{y^{\prime}_k} = y_k + \frac{f}{d} \cdot v_{\text{min}} t, \  \overline{y^{\prime}_k} =  y_k + \frac{f}{d} \cdot v_{\text{max}} t.
\end{equation}
\normalsize
The other 16 cases are based on the monotonicity of functions. For example, if $\omega_{\text{min}} \geq 0$, $v_{\text{min}} \geq 0$ and $x_k \geq u_0,\ y_k \geq v_0 - l \cdot \frac{f}{d}$, the lower bound of $x^{\prime}_k$ is
\small
\begin{eqnarray}
    \underline{x^{\prime}_k} &=& \min_{\omega} a_x + \min_{\omega} b_x + \min_{\omega,v} c_x + u_0 \ \text{, with} \\
    \min_{\omega} a_x & \geq & - [y_k - v_0 + l \cdot \frac{f}{d}] \sin(\omega_{\text{max}} t), \nonumber \\
    \min_{\omega} b_x & \geq & [x_k - u_0 ] \cos(\omega_{\text{max}} t) , \nonumber \\
    \min_{\omega,v} c_x & \geq & \frac{f}{d} \cdot \frac{v_{\text{min}}}{\omega_{\text{max}}}[1- \cos(\omega_{\text{max}} t)].
\end{eqnarray}
\normalsize
Table~\ref{tab:bounding box cases} lists $\underline{x^{\prime}_k}$ and $\underline{y^{\prime}_k}$ with $\omega$ and $v$ arguments when the search space is $\omega_{\text{min}} > 0$. Meanwhile $\overline{x^{\prime}_k}$ and $\overline{y^{\prime}_k}$ are obtained by $\omega_{\text{min}}$ against $\omega_{\text{max}}$, and $v_{\text{min}}$ against $v_{\text{max}}$. The other 8 cases with $\omega_{\text{max}} < 0$ are derived by a similar strategy.

\ifCLASSOPTIONcaptionsoff
  \newpage
\fi

\vfill



%


\end{document}